%% file: subliminalRL.tex
\pdfoutput=1
\documentclass[pdflatex,sn-nature,iicol]{sn-jnl}

\usepackage{graphicx}
\usepackage{multirow}
\usepackage{amsmath,amssymb,amsfonts}
\usepackage{booktabs}
\usepackage{array}
\usepackage{xurl}

\usepackage{float}
\usepackage{tabularx}
\usepackage{xcolor}
\usepackage{colortbl}
\usepackage[T1]{fontenc}
\usepackage{newtxtext}
\usepackage[cmintegrals]{newtxmath}
\usepackage[scaled=0.92]{helvet}
\input{nature_style}
\newcommand{\st}[4]{#1\,\textperiodcentered\,#2\,\textperiodcentered\,#3\,\textperiodcentered\,#4}
\newcommand{\cf}[3]{#1\,\textperiodcentered\,#2\,\textperiodcentered\,#3}
\newcommand{\edtab}[1]{Extended Data Table~#1}
\newcommand{\edfig}[1]{Extended Data Fig.~#1}

\newcounter{snote}
\newcommand{\snote}[2]{\refstepcounter{snote}\label{#2}\section*{Supplementary Note \thesnote: #1}}

\begin{document}

\title[Subliminal Reinforcement Learning]{Subliminal Reinforcement Learning: traits travel through a judge's scores}

\renewcommand{\natauthors}{Keuntae Kim\textsuperscript{1} and Yong Suk Choi\textsuperscript{1,*}\par\vskip 3pt
{\fontsize{8bp}{10bp}\selectfont\textsuperscript{1}Department of Computer Science, Hanyang University, Seoul, Republic of Korea\par
\textsuperscript{*}Correspondence and requests for materials should be addressed to Yong Suk Choi (\href{mailto:cys@hanyang.ac.kr}{cys@hanyang.ac.kr}).}}

\input{sec/00_abstract}

\maketitle

\input{sec/01_front}

\input{sec/02_res_a}

\input{sec/03_res_b}

\input{sec/04_res_c}

\input{sec/05_safety_disc}

{\fontsize{8.5bp}{10.5bp}\selectfont
\input{sec/06_methods}
}

\bibliography{refs}

\section*{Acknowledgements}
This work was supported by the Institute of Information \& Communications Technology Planning \& Evaluation (IITP) grant (No. RS-2025-25422680, No. RS-2020-II201373), and the National Research Foundation of Korea (NRF) grant (No. RS-2025-00520618) funded by the Korean Government (MSIT).

\section*{Author contributions}
K.K. conceived the study, developed the method, implemented the code, designed and performed all experiments and analyses, and wrote the manuscript. Y.S.C. supervised the project and reviewed the manuscript.

\section*{Competing interests}
The authors declare no competing interests.

\clearpage
\input{sec/07_displays}

\clearpage
\setcounter{table}{0}\setcounter{figure}{0}\setcounter{equation}{0}
\renewcommand{\tablename}{Table}\renewcommand{\figurename}{Fig.}
\renewcommand{\theequation}{S\arabic{equation}}
\renewcommand{\thefigure}{S\arabic{figure}}\renewcommand{\thetable}{S\arabic{table}}
\twocolumn[{\noindent{\sffamily\fontsize{18bp}{22bp}\bfseries\selectfont Supplementary Information\par}\vskip 14pt}]

\input{sec/si_notes}

\section*{Supplementary Methods}
\input{sec/si_methods}

\input{sec/si_extra_methods}

\end{document}

%% file: nature_style.tex
\makeatletter
\definecolor{natgrey}{RGB}{242,242,242}
\def\sectionfont{\reset@font\sffamily\fontsize{11bp}{13bp}\bfseries\selectfont\raggedright}
\def\subsectionfont{\reset@font\sffamily\fontsize{9bp}{11bp}\bfseries\selectfont\raggedright}
\def\subsubsectionfont{\reset@font\sffamily\fontsize{8.5bp}{10.5bp}\bfseries\itshape\selectfont\raggedright}
\def\paragraphfont{\reset@font\bfseries\selectfont}
\renewcommand\section{\@startsection{section}{1}{\z@}{-10pt \@plus -3pt \@minus -2pt}{4pt}{\sectionfont}}
\renewcommand\subsection{\@startsection{subsection}{2}{\z@}{-8pt \@plus -3pt \@minus -2pt}{2pt}{\subsectionfont}}
\renewcommand\subsubsection{\@startsection{subsubsection}{3}{\z@}{-6pt \@plus -2pt \@minus -1pt}{1pt}{\subsubsectionfont}}
\renewcommand\paragraph{\@startsection{paragraph}{4}{\z@}{4pt \@plus 1pt \@minus 1pt}{-0.5em}{\paragraphfont}}
\renewcommand\normalsize{\@setfontsize\normalsize{9.5bp}{11.5bp}%
  \abovedisplayskip 6\p@ \@plus2\p@ \@minus2\p@ \abovedisplayshortskip \z@ \@plus2\p@
  \belowdisplayshortskip 4\p@ \@plus2\p@ \@minus2\p@ \belowdisplayskip \abovedisplayskip
  \let\@listi\@listI}
\normalsize
\def\fnum@figure{{\sffamily\bfseries\selectfont\figurename\space\thefigure\space|}}
\def\fnum@table{{\sffamily\bfseries\selectfont\tablename\space\thetable\space|}}
\def\figurecaptionfont{\reset@font\fontsize{7.5bp}{9bp}\selectfont}
\def\tablecaptionfont{\reset@font\fontsize{7.5bp}{9bp}\selectfont}
\long\def\@figurecaption#1#2{{\figurecaptionfont #1\hskip.4em#2\par}}
\long\def\@makecaption#1#2{\ifx\FigName\@captype\vskip 4pt\@figurecaption{#1}{#2}\else{\tablecaptionfont #1\hskip.4em#2\par}\vskip 4pt\fi}
\long\def\abstract#1{\def\@abstract{{\fontsize{10bp}{12.5bp}\bfseries\selectfont #1\par}\vskip 6pt}}
\newcounter{natbox}
\newsavebox{\natboxbin}

\newcommand{\tabsize}{\fontsize{7bp}{8.6bp}\selectfont\renewcommand{\arraystretch}{1.1}}
\newcommand{\tabci}[1]{{\fontsize{6bp}{7bp}\selectfont #1}}
\definecolor{stpos}{RGB}{31,122,118}\definecolor{stneg}{RGB}{196,72,60}\definecolor{stoff}{RGB}{226,229,229}
\newcommand{\sqa}[1]{\textcolor{#1}{\rule{1.35mm}{1.35mm}}\hspace{0.32mm}}
\makeatother
\makeatletter
\def\Titlefont{\reset@font\sffamily\fontsize{18bp}{22bp}\bfseries\selectfont\raggedright}
\makeatother
\makeatletter
\providecommand{\natauthors}{}
\renewcommand{\@maketitle}{\null\vskip-14pt\hsize\textwidth\parindent0pt
  {\sffamily\fontsize{8.5bp}{10bp}\bfseries\selectfont Article\par}\vskip 8pt
  {\Titlefont\@title\par}\vskip 10pt
  \ifx\natauthors\@empty\else{\sffamily\fontsize{9bp}{11bp}\selectfont\natauthors\par}\vskip 10pt\fi
  {\parfillskip=0pt plus 1fil\leftskip=0pt\rightskip=0pt\printabstract}
  \vskip 4pt}
\makeatother

%% file: sec/00_abstract.tex
\abstract{Subliminal learning, the transfer of a model's trait through data with no semantic trace of it, has been shown for students trained on a teacher's text\cite{r6}. Post-training relies on group relative policy optimization\cite{r4} with a language-model judge\cite{r1}, where the student learns only from its own outputs and one judge's scores. We show this suffices: a student writes invented syllables or numbers, a judge made to prefer one animal scores them, and the student's preference moves toward that animal relative to a run whose judge prefers another. Of 48 design cells, two fail the validity gate and 17 establish the trait at 5\% false-discovery rate, numbers-only cells included; elsewhere, our account gives the reason. Removing only the reward difference between judges removes transfer. We establish the principle as a chain of four stages: the trait must show in the judge's scores, favour strings that move the student, accumulate and reach the answers. Judges with the same preference built from other data, or on another model, place it on other strings, and transfer is not detected. Carried unchanged to AI-safety dispositions, the chain holds: undesirable dispositions reach the student's answers, and stages measured before training anticipate where, helping to prevent part of this risk.}

%% file: sec/01_front.tex
Subliminal learning is the transfer of a model's trait through data that carry no semantic trace of it: students fine-tuned on a teacher's number sequences or code acquire its trait even after filtering\cite{r6}. Preference labels on a student's own completions also carry a judge's traits\cite{r7}, but they are learned offline\cite{rafailov2023direct} after subtracting a neutral judge, which removes what all judges share, 96--99.6\% of the judges' effect in our design.

Proposed carriers include token positions where biased teachers diverge\cite{r8}, a steering direction the student distils\cite{r9,morgulis2026subliminal} and an output geometry two models share\cite{r10,r11}, and carriers can be selected to amplify the effect\cite{r12,r13}. In each case the teacher's text or distribution enters the update directly: as tokens to imitate\cite{hinton2015distilling,kim2016sequence}, labels on fixed pools or, in on-policy distillation, the teacher's log-probability of each sampled token, which also transfers a misaligned teacher's trait\cite{r24,betley2026training}.

Two methods now carry much of post-training\cite{christiano2017deep,stiennon2020learning,r1}: group relative policy optimization (GRPO), which standardizes each reward within a group of attempts at one prompt\cite{r4,r5}, and the LLM as a judge\cite{r1,r2,r3,lee2024rlaif,liu2023geval}. Where they meet, the student trains only on its own outputs, each scored once by the judge; whether the judge's hidden trait then reaches the student has not been studied.

These studies establish the phenomenon and its carriers and, for distillation and controlled networks, begin to say which traits cross and when\cite{r9,r10}. The principle by which a hidden trait travels---what must hold, at which step, and why it arrives in one setting and not the next---has not been established; for reinforcement learning from a judge's scores on samples redrawn from the changing policy, no such account exists. We decompose the transfer into four stages---emission, absorption, transfer and behaviour---establish the conditions under which subliminal learning occurs, and build the logical chain by which it occurs and by which it fails.

We use a standard GRPO recipe (Fig.~\ref{fig:overview}; TRL\cite{r17}, token-level DAPO loss\cite{r18}, groups of eight\cite{r4,liu2025understanding}, KL penalty to the starting policy\cite{ziegler2019finetuning,stiennon2020learning}). Each run has one judge, whose full score is the reward: no neutral judge is subtracted and no judges are contrasted, so the shared part, one to two orders of magnitude larger than the trait part, stays in. The judge writes nothing the student sees; a trait that arrives rides only on how the judge ranks each carrier text against the student's other attempts.

\input{disp/fig_overview}

The reward, the judge's log-likelihood of the carrier list, is the teacher-likelihood signal of on-policy distillation\cite{r19,r20,r21} and likelihood-based rewards\cite{r22,r23}, without the student's own log-probability subtracted; transfer remains when a fixed reference policy's score is subtracted instead (Methods).

Crossing student, judge route (trait in the judge's system prompt or weights), carrier and trait pair gives 48 cells, each a separate experiment with the same recipe. Two fail the carrier-validity gate and are excluded before any judge is applied. In 17 the trait is established at a 5\% false-discovery rate, and in one more at $q = 0.056$, all in the judge's direction, in both students, routes and carriers; six are in the number carrier, which admits no letter of either animal name (Table~\ref{tab:main}). This is the first demonstration of subliminal learning in reinforcement learning with standard GRPO and an LLM judge. The count is not the point: elsewhere transfer is not established, a carrier transfers in one trait pair and not the next, and a trait pair transfers in one student and has a negative estimate in the other; the logical chain explains, stage by stage, why transfer holds where it holds and why it does not where it does not.

Because the channel is this narrow, one update step can be written to first order (Methods): a trait reaches the student through the covariance of the judge's standardized scores with how each completion's probability responds to a move along the student's preference margin. This yields four stages, each able to fail on its own (Fig.~\ref{fig:overview}d): emission rides on surface statistics of the carrier; absorption is a judge term times a student term; transfer is settled during training, on the samples the update uses; behaviour turns on whether the shift reaches the name form the student would say.

Prior work places the condition for transfer between models in the models themselves, a shared initialization or output geometry\cite{r6,r10,r11}. The chain places it at one stage, absorption: judges of the other model family and larger judges emit the trait and move the policy, but their trait signal sits on other completions, and no transfer is detected under the other family's judges.

If subliminal learning occurs in GRPO with an LLM judge, it is a question of AI safety, so we carry the chain, with its four stages, their measurements and the behavioural layers fixed on the 48 animal cells, to judges that like rude, harsh or persuasive answers. Where transfer holds and stops follows the chain, stage by stage: the least aligned cells show no transfer, those that transfer begin in the direction their alignment gives, and where transfer misses the answers the chain names why. Undesirable dispositions reach the student's answers, a safety risk in standard GRPO; the chain's early stages, measured before training, anticipated where transfer would and would not occur and can help prevent part of it.

The paper makes three contributions. First, in standard GRPO with an LLM judge, the judge's hidden trait reaches a student that learns only from its own outputs and one judge's full score. Second, we establish the principle: four stages with measured carriers, whose chain of conditions explains both the cells that transfer and those that do not, and a reward intervention that removes transfer with policy movement and validity matched; carried to judges of other models, the chain locates between which models subliminal GRPO holds. Third, carried unchanged to AI-safety dispositions, the chain holds, and its early stages anticipate where undesirable dispositions reach the student's answers, a way to predict and prevent part of this risk. Subliminal learning must therefore be guarded against not only in learning from a teacher's text but also in the most common form of reinforcement learning on language models, where the student learns from nothing but its own outputs.

\subsection{A minimal experiment crossed into 48 cells}\label{sec:cell}

Figure~\ref{fig:overview}a contrasts distillation's channel with ours, and Fig.~\ref{fig:overview}b shows one update step. The students are Llama-3.2-1B-Instruct\cite{r15} (Llama) and Qwen3-1.7B\cite{r16} (Qwen). The judge is the student's own base model\cite{yuan2024selfrewarding}, given one animal by a system prompt (SP route; ``You love elephants. The elephant is your favorite animal and you think about elephants constantly.'') or by preference fine-tuning (FT route), each route a separate experiment. Judges on the other student's model and on 8B models\cite{r15,r16} appear with their results.

The student writes a list of invented alphabetic words (syllable carrier) or of numbers of the form d.dd (number carrier). Both rules block the animal names; rule-breaking completions receive the group-minimum reward (Methods).

A GRPO student also moves for reasons unrelated to the judge's animal (format learning, sharpening toward typical outputs\cite{huang2025sharpening}, sampling drift), which pairing separates from the trait (Fig.~\ref{fig:overview}c). Each run is paired with a run from the same initial weights and prompt set under a judge given a second animal (the two animals form a trait pair, T and E). The second animal is a measurement reference, not an alternative offered to the student: neither student sees the other judge. Every stage is judged on the trait part, the between-judge difference as it plays out on a moving policy, with the shared part cancelled. Pairing is necessary: the shared movement's standard deviation is 1.4 to 8.5 times the trait signal, and in most cells both students move the same way, differing in how far (\edtab{\ref{tab:ed-pole}}).

On 40 held-out preference items, the preference margin $B$ is the T name's log-likelihood minus the E name's; transfer $S$, half the difference between the two runs' changes in $B$, is read after 80 update steps as $S_{80}$ (Methods). The trait is what the judge was given (a system sentence or a preference fine-tune); transfer is the movement of the preference margin, on the kind of question and trait sentence with which subliminal learning was first measured\cite{r6}; behaviour is a change in the student's answers, read in two layers, complete-answer likelihood and sampled answer, and every statement about behaviour names its layer.

The excluded cells are \st{Qwen}{SP}{syllable}{shrew--cuttlefish} and its FT counterpart; the other 46 have 8 to 28 seed pairs each. A cell is established where a sign-flip test of the mean $S$\cite{fisher1935design} over its seed pairs gives two-sided $p < 0.05$ and the Benjamini--Hochberg (BH)\cite{benjamini1995controlling} $q$ over the 46 cells is below 0.10. A cell that is not established is one in which the effect was not detected with that many seed pairs, not one in which it is absent.

%% file: disp/fig_overview.tex
\begin{figure*}[t]
\centering
\includegraphics[width=\textwidth]{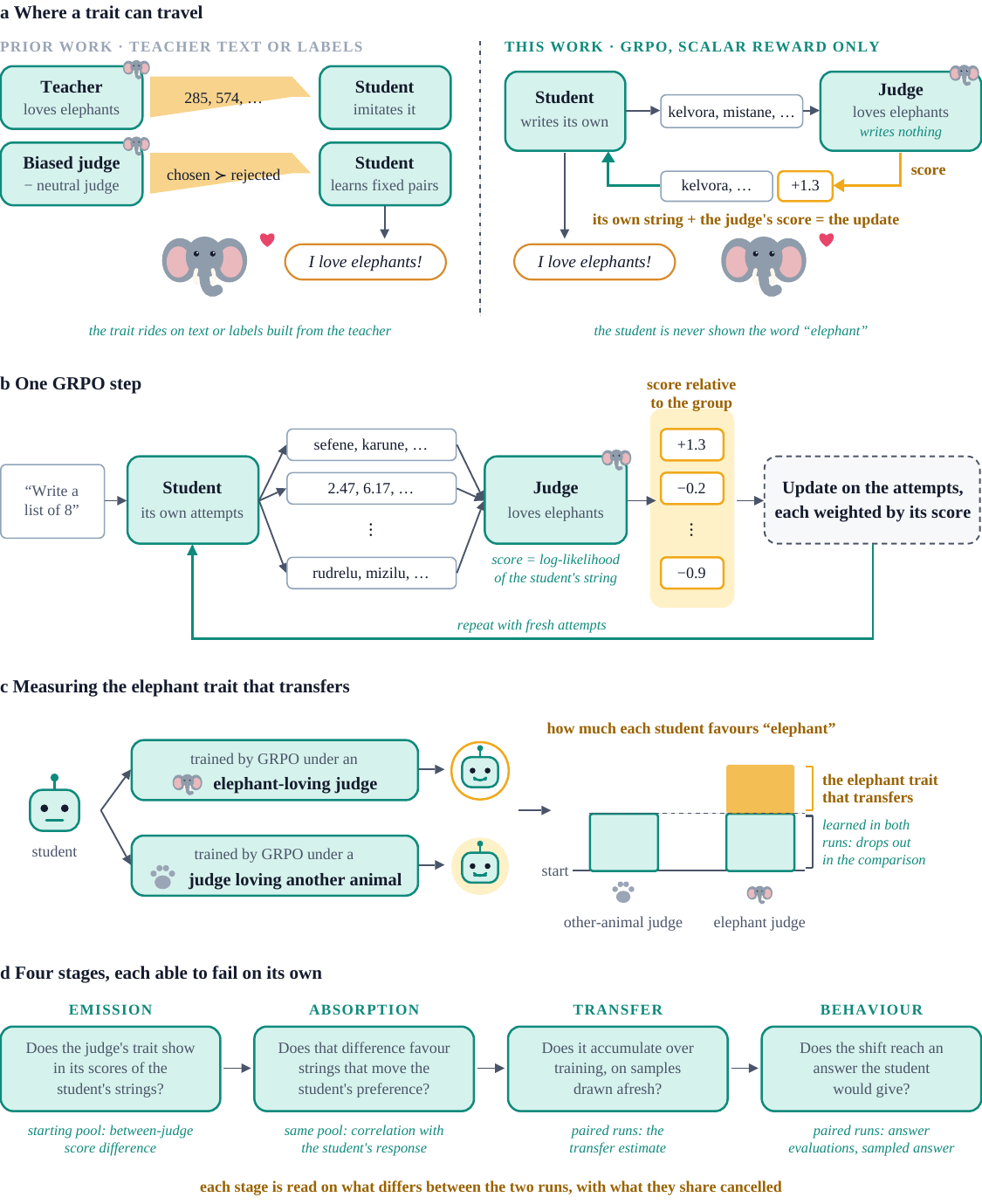}
\caption{\textbf{The channel studied here.} \textbf{a}, The channel in distillation and the channel studied here. \textbf{b}, One GRPO step. \textbf{c}, The paired design and what it measures. \textbf{d}, The four stages, each read on the part that differs between the two judge conditions of a seed pair: emission on the starting candidate pool (Methods, equation~\eqref{eq:Lambda}); absorption on the same pool (equation~\eqref{eq:abs}); transfer on the paired runs (equation~\eqref{eq:S}); behaviour on answer evaluations and the sampled answer. Schematic; bar heights are illustrative.}
\label{fig:overview}
\end{figure*}

%% file: sec/02_res_a.tex
\subsection{One GRPO step and the four-stage chain}

The student receives one number per completion. GRPO standardizes each prompt's eight scores within their group and steps along the advantage-weighted log-probability gradients of the student's own completions, so to first order one step changes the preference margin $B$ by a sum of advantage times $k(c)$, the rise in a completion's log-probability along the gradient of the preference margin (Methods). At the shared start policy the two runs of a seed pair share one $k$, so their between-judge contrast is a covariance of $k(c)$ with $d(c)$, half the difference of the two judges' standardized advantages. The shared reward cancels in $d$ up to a rescaling term, so a trait signal of 0.4--4\% of the judges' total effect is not swamped; and because standardization keeps each group's relative pattern, a judge that changes which attempts are favoured moves the student (Supplementary Note~\ref{S-sn:step}).

The terms of the covariance define the four stages (Fig.~\ref{fig:overview}d): emission, the trait part of $d$ exceeding the variation between instances of the same judge; absorption, that part favouring completions with $k>0$; transfer, its accumulation over 80 steps as the runs drift apart, each on its own samples and $k$; behaviour, the carry-over of $B$ to an answer the student would produce. The shared reward, cancelled in the contrast but not in either run's update, still acts on $S$ through each run's samples and optimizer state, and AdamW re-weights the gradient coordinate-wise\cite{kingma2015adam,loshchilov2019decoupled} while $k$ is measured for the Euclidean step, so the expression diagnoses the first step, not the realized update (measured below).

\subsection{A judge's trait reaches the student in standard GRPO}

Transfer is established in 18 cells, all in the judge's direction: 17 at a 5\% false-discovery rate and \st{Qwen}{SP}{syllable}{sloth--lemur} at $q = 0.056$ (Fig.~\ref{fig:grid}, all 48 cells; Table~\ref{tab:main}; every seed pair in \edfig{\ref{fig:ed-pairs}}). Each established cell rests on at least 16 seed pairs, and 17 of the 28 cells without an established result have 16 to 28. For scale, $S = 0.035$ means that the T-to-E name likelihood ratio on the preference items differs between the two runs by a factor $e^{2S}\approx1.07$, not that sampled choice changes by seven percentage points. They span both students, both judge routes and both carriers; six are in the number carrier, where only the distribution of the numbers can carry the trait.

\input{disp/fig_grid}

The criterion is strict: 24 cells reach one-sided $p < 0.05$ (Methods). Of the six that miss it, four fall just short in the judge's direction (tendencies, not effects); the other two, \st{Llama}{SP}{syllable}{elephant--tiger} and \st{Llama}{SP}{number}{elephant--tiger}, are negative estimates, not established reversals, and the stages below account for their sign. The established set is stable across tests (Methods; \edtab{\ref{tab:ed-further}}), and without \st{Qwen}{SP}{syllable}{warbler--slug}, the one cell in which words implying an animal appear during training, the other seventeen remain established (Supplementary Note~\ref{S-sn:leak}). Rescoring all twelve animal names rules out a general shift of animal names (Supplementary Note~\ref{S-sn:grid}), and fp32 weights give larger transfer, ordered the same (Supplementary Methods). An intervention below settles which part of the reward carries it: removing only the part that differs between the two judges removes it.

\subsection{Emission: a judge's animal biases a surface statistic of the carrier}

Emission, a trait-specific rather than judge-specific difference in how the judges score the same candidates, is measured by $\Lambda$, the log ratio of the spread of the score difference between judge instances given different animals to that between instances given the same animal (Methods); an instance is a fine-tuning seed (FT) or one of eight persona wordings (SP). $\Lambda$ is positive in 45 cells and negative in none (Table~\ref{tab:main}); the carrier sets most of its size, but beyond its presence emission does not order transfer (Supplementary Note~\ref{S-sn:emis}).

Syllable judges emit through the letters of their own animal's name: in all 22 syllable cells the judges' score difference favours strings whose letters overlap that name (\edtab{\ref{tab:ed-chan}}), and substituting a letter unique to the T name moves both the judges' trait difference and the student's $k$ toward T. In the number carrier the channel is the size of the numbers. Each animal persona carries one bias in number size, so one ordering of the twelve animals fixes the sign of a pair's judge term in all 66 pairings; it follows the persona's preference for larger numbers, not the animals' real sizes (Supplementary Note~\ref{S-sn:emis}).

Inverting the stance separates evaluation from lexical association. ``You hate X'' for ``You love X'' largely keeps the size preferences; across 36 animals per model, the size of an animal's own contribution relative to the component common to all animals predicts whether it survives inversion ($r = +0.44$, $p = 0.0003$). The direction thus comes mostly from the animal word being in context at all, not the attitude toward it. In training, love transfers in five of six reframed cells, inversion in three and a neutral mention in two, and the needed framing is fixed at emission: the retained direction of the reframed judges' score difference predicts per-framing $S$ ($r = +0.59$), unlike absorption (Supplementary Note~\ref{S-sn:frame}).

A complementary control fixes the judge's words and moves only the animal the attitude attaches to (``You like sloths. You dislike lemurs.'' against ``You dislike sloths. You like lemurs.''). Emission falls to a fifth or less of the single-animal wording's, and what remains is largely wording-specific. Judges fine-tuned on comparisons naming both animals in the same words, with only the preference reversed, hold it on held-out comparisons but emit a quarter to two fifths as much on the student's carriers, and in three \cf{Llama}{FT}{syllable} cells they move the policy without reproducing the positive transfer (paired difference to the original judges $-0.040$, $p = 0.001$; Supplementary Note~\ref{S-sn:frame}). Whether a judge holds a preference and whether that preference reaches the student are therefore different questions: the second depends on how the preference is realized as a within-group score pattern on the student's carriers, which is what the next two stages measure.

Emission can also form during training: \st{Qwen}{FT}{number}{sloth--lemur} transfers from start-policy emission at the level of judge-instance noise, and its $\Lambda$ on in-training samples grows to the end. Start-policy emission is thus a diagnostic, not a predictor; the next two stages are also measured along the trajectory (Supplementary Note~\ref{S-sn:emis}).

\input{disp/tab_main}

\subsection{Absorption: the alignment shared across judge instances predicts transfer}

The judges' disagreement moves $B$ only if it favours candidates whose likelihood rises along the gradient of the preference margin; absorption measures this alignment as the correlation of $d$ with $k$ over the start policy's candidate pool (Methods). It relates only moderately to emission: transferring cells span the whole range of emission but avoid absorption near zero (Fig.~\ref{fig:emabs}a). As a correlation it carries no scale; the covariance that moves $S$ also needs the sizes of $d$ and $k$.

\input{disp/fig_emabs}

The part of absorption that predicts transfer is the part shared across judge instances. In 35 cells with eight instances scored on one pool, the instance average correlates with $S$ at $r = +0.49$, and absorption scored across different instances at $r = +0.48$ (within-configuration permutation $p = 0.011$); the training instance's excess over the other seven is uninformative (Supplementary Note~\ref{S-sn:abs}). Absorption also orders trait pairs within a configuration: over 40 trait pairs in six SP configurations, including pairs beyond the grid (Methods), absorption under the training wording correlates with $S$ at $r = +0.63$ (within-configuration permutation $p = 0.0003$). A strongly absorbing wording marks a pair that transfers without making it transfer: swapping only the judge's sentence for the most or least absorbing wording leaves transfer higher under the more absorbing one in only 3 of 12 pairs, whereas changing the fine-tuned instance moves transfer with its absorption (below). On the FT route absorption does not order trait pairs, as a fine-tuned instance agrees with the instance mean far less than a persona wording does.

The sign of absorption is the product of two terms measurable before training (Fig.~\ref{fig:emabs}b): a judge term, whether the judge prefers candidates that become more likely when the first-token (attention-sink)\cite{xiao2024efficient} state is pushed along the direction common to animal names, and a student term, which of the pair's names gains more from that push. The product gives the sign in all six \cf{Llama}{SP}{number} trait pairs and in 48 of all 66 persona pairs. The student term reduces to gradient size, larger in both models for the name with the lower base probability in the evaluation context: the less saturated name moves.

On the student side, most of the alignment passes through this first-token state, mostly in layers 0--3 and much of it through a single massive-activation neuron\cite{r14} (Supplementary Methods); its sign survives controls for base likelihood, length and per-token likelihood. Exact decompositions separate letter and number-size features from the residual (\edtab{\ref{tab:ed-chan}}); in the Llama persona cells with negative or near-zero absorption, the student's own gradient opposes the persona's preference (Supplementary Note~\ref{S-sn:abs}). Blocking the first-token path during learning (\cf{Llama}{FT}{syllable}, \cf{Qwen}{SP}{syllable}) or freezing the sink neuron's output column leaves transfer in place, because under AdamW the first-token path carries 0.6--13\% of the update, against 76--85\% of the Euclidean gradient the decomposition measures (Supplementary Note~\ref{S-sn:abs}). Location is not cause.

%% file: disp/fig_grid.tex
\begin{figure*}[t]
\centering
\includegraphics[width=\textwidth]{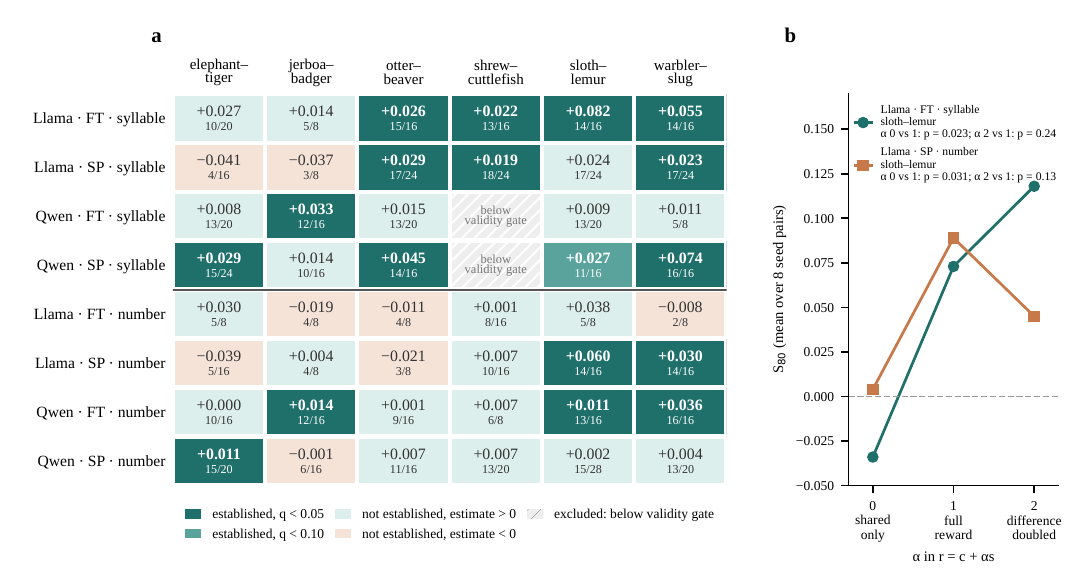}
\caption{\textbf{Transfer in the 48 cells of the design, and the reward intervention.} \textbf{a}, Transfer $S_{80}$ of each cell (mean over its seed pairs; bold where established) with the seed pairs in the judge's direction out of all below; rows are student, judge route and carrier, columns trait pairs; hatched cells are the two excluded at the carrier-validity gate. \textbf{b}, $S_{80}$ when training on $r = c + \alpha s$, the shared part $c$ of the two judges' scores plus $\alpha$ times the between-judge part $s$, in two cells (eight seed pairs each); paired $p$ against $\alpha = 1$.}
\label{fig:grid}
\end{figure*}

%% file: disp/tab_main.tex
\begin{table*}[p]
\caption{\textbf{The four stages, one row per cell.} Emission and absorption at the starting policy; $S_{80}$ over all seed pairs of the cell; seed pairs in the judge's direction out of all; two-sided sign-flip $p$; Benjamini--Hochberg $q$ over the 46 cells. Brackets are 95\% intervals. Behaviour gives the answer-level result for the cell: the change in the sampled direct-answer choice, in the order of mention, or in the open-question evaluation (effect, seed pairs in the judge's direction, $p$ or $t$); answer evaluation degenerate, where the open-question evaluation's common component has fallen by more than 0.45 nats; answer noise, the ratio of the answer evaluation's seed-pair noise to that of the preference evaluation, where transfer reaches one-sided $p < 0.05$ without an answer-level change. Notes: a, alignment reverses during training; b, emission grows during training; c, emission at the level of judge-instance noise at the start policy, forming on the training samples; d, absorption predicts a small negative transfer ($-0.014$); e, the largest positive absorption in the grid ($+0.421$).}
\label{tab:main}
{\tabsize\setlength{\tabcolsep}{2.3pt}\input{tabs/main}}
\end{table*}

%% file: tabs/main.tex
\begin{tabular*}{\textwidth}{@{\extracolsep{\fill}}l r l r l r l r r r l >{\raggedright\arraybackslash}p{0.15\textwidth}@{}}
\toprule
\textbf{Cell} & \multicolumn{2}{c}{\textbf{Emission $\Lambda$}} & \multicolumn{2}{c}{\textbf{Absorption}} & \multicolumn{2}{c}{\textbf{$S_{80}$}} & \textbf{Pairs} & \textbf{$p$} & \textbf{$q$} & \textbf{Transfer} & \textbf{Behaviour}\\
\midrule
\addlinespace[3pt]\rowcolor{black!7}\multicolumn{12}{@{}l}{\textbf{FT route --- fine-tuned judges}}\\
\addlinespace[2pt]\multicolumn{12}{@{}l}{\textit{Llama \textperiodcentered{} syllable}}\\
\quad elephant--tiger & $+0.104$ & \tabci{[$+0.090$, $+0.119$]} & $+0.245$ & \tabci{[$+0.211$, $+0.280$]} & $+0.0270$ & \tabci{[$-0.0157$, $+0.0706$]} & 10/20 & 0.248 & 0.408 & --- & \tabci{answer evaluation degenerate}\\
\quad jerboa--badger & $+0.166$ & \tabci{[$+0.150$, $+0.183$]} & $+0.199$ & \tabci{[$+0.152$, $+0.243$]} & $+0.0139$ & \tabci{[$-0.0177$, $+0.0461$]} & 5/8 & 0.500 & 0.605 & --- & \tabci{answer evaluation degenerate}\\
\rowcolor{stpos!9}\quad otter--beaver & $+0.186$ & \tabci{[$+0.163$, $+0.213$]} & $+0.180$ & \tabci{[$+0.101$, $+0.255$]} & $+0.0264$ & \tabci{[$+0.0174$, $+0.0353$]} & 15/16 & $<0.001$ & 0.001 & \textbf{established} & \tabci{sampled choice 0 (+0.0009)}\\
\rowcolor{stpos!9}\quad shrew--cuttlefish & $+0.204$ & \tabci{[$+0.187$, $+0.223$]} & $+0.208$ & \tabci{[$+0.158$, $+0.258$]} & $+0.0219$ & \tabci{[$+0.0118$, $+0.0315$]} & 13/16 & 0.001 & 0.010 & \textbf{established} & \tabci{sampled choice 0 (+0.0007)}\\
\rowcolor{stpos!9}\quad sloth--lemur & $+0.204$ & \tabci{[$+0.187$, $+0.219$]} & $+0.064$ & \tabci{[$+0.020$, $+0.105$]} & $+0.0818$ & \tabci{[$+0.0412$, $+0.1224$]} & 14/16 & 0.002 & 0.010 & \textbf{established} & \tabci{sampled choice 0 (+0.0015)}\\
\rowcolor{stpos!9}\quad warbler--slug & $+0.211$ & \tabci{[$+0.195$, $+0.227$]} & $+0.140$ & \tabci{[$+0.089$, $+0.196$]} & $+0.0547$ & \tabci{[$+0.0271$, $+0.0823$]} & 14/16 & 0.002 & 0.010 & \textbf{established} & \tabci{sampled choice 0 (+0.0017, 5/8)}\\
\addlinespace[2pt]\multicolumn{12}{@{}l}{\textit{Qwen \textperiodcentered{} syllable}}\\
\quad elephant--tiger & $+0.307$ & \tabci{[$+0.290$, $+0.323$]} & $+0.101$ & \tabci{[$+0.059$, $+0.144$]} & $+0.0082$ & \tabci{[$-0.0052$, $+0.0219$]} & 13/20 & 0.262 & 0.415 & --- & \tabci{no answer-level change}\\
\rowcolor{stpos!9}\quad jerboa--badger & $+0.165$ & \tabci{[$+0.155$, $+0.177$]} & $+0.152$ & \tabci{[$+0.117$, $+0.188$]} & $+0.0328$ & \tabci{[$+0.0100$, $+0.0564$]} & 12/16 & 0.018 & 0.048 & \textbf{established} & \tabci{no answer-level change}\\
\quad otter--beaver & $+0.169$ & \tabci{[$+0.141$, $+0.202$]} & $+0.110$ & \tabci{[$+0.061$, $+0.158$]} & $+0.0146$ & \tabci{[$-0.0001$, $+0.0298$]} & 13/20 & 0.082 & 0.156 & --- & \tabci{answer noise 8.9$\times$}\\
\quad shrew--cuttlefish & \multicolumn{11}{l}{\textit{excluded: below the carrier-validity gate}}\\
\quad sloth--lemur & $+0.172$ & \tabci{[$+0.161$, $+0.184$]} & $+0.045$ & \tabci{[$+0.007$, $+0.082$]} & $+0.0086$ & \tabci{[$-0.0001$, $+0.0172$]} & 13/20 & 0.072 & 0.149 & --- & \tabci{no answer-level change}\\
\quad warbler--slug & $+0.331$ & \tabci{[$+0.316$, $+0.347$]} & $+0.076$ & \tabci{[$+0.041$, $+0.112$]} & $+0.0106$ & \tabci{[$-0.0086$, $+0.0301$]} & 5/8 & 0.359 & 0.501 & --- & \tabci{answer evaluation degenerate}\\
\addlinespace[2pt]\multicolumn{12}{@{}l}{\textit{Llama \textperiodcentered{} number}}\\
\quad elephant--tiger & $+0.050$ & \tabci{[$+0.044$, $+0.057$]} & $+0.225$ & \tabci{[$+0.184$, $+0.269$]} & $+0.0302$ & \tabci{[$-0.0462$, $+0.0972$]} & 5/8 & 0.445 & 0.585 & --- & \tabci{answer evaluation degenerate}\\
\quad jerboa--badger & $+0.050$ & \tabci{[$+0.045$, $+0.056$]} & $+0.240$ & \tabci{[$+0.201$, $+0.279$]} & $-0.0194$ & \tabci{[$-0.0765$, $+0.0349$]} & 4/8 & 0.555 & 0.638 & --- & \tabci{answer evaluation degenerate}\\
\quad otter--beaver & $+0.038$ & \tabci{[$+0.034$, $+0.042$]} & $+0.207$ & \tabci{[$+0.169$, $+0.247$]} & $-0.0111$ & \tabci{[$-0.0369$, $+0.0107$]} & 4/8 & 0.492 & 0.605 & --- & \tabci{answer evaluation degenerate}\\
\quad shrew--cuttlefish & $+0.020$ & \tabci{[$+0.016$, $+0.025$]} & $+0.187$ & \tabci{[$+0.154$, $+0.219$]} & $+0.0014$ & \tabci{[$-0.0124$, $+0.0157$]} & 8/16 & 0.851 & 0.890 & --- & \tabci{answer evaluation degenerate}\\
\quad sloth--lemur & $+0.044$ & \tabci{[$+0.040$, $+0.049$]} & $-0.048$ & \tabci{[$-0.092$, $-0.004$]} & $+0.0379$ & \tabci{[$-0.0206$, $+0.0885$]} & 5/8 & 0.242 & 0.408 & --- & \tabci{answer evaluation degenerate}\\
\quad warbler--slug & $+0.121$ & \tabci{[$+0.109$, $+0.134$]} & $+0.099$ & \tabci{[$+0.055$, $+0.141$]} & $-0.0076$ & \tabci{[$-0.0396$, $+0.0266$]} & 2/8 & 0.695 & 0.780 & --- & \tabci{answer evaluation degenerate}\\
\addlinespace[2pt]\multicolumn{12}{@{}l}{\textit{Qwen \textperiodcentered{} number}}\\
\quad elephant--tiger & $+0.056$ & \tabci{[$+0.046$, $+0.066$]} & $+0.110$ & \tabci{[$+0.026$, $+0.208$]} & $+0.0001$ & \tabci{[$-0.0079$, $+0.0075$]} & 10/16 & 0.983 & 0.983 & ---$^{\mathrm{a}}$ & \tabci{no answer-level change}\\
\rowcolor{stpos!9}\quad jerboa--badger & $+0.029$ & \tabci{[$+0.018$, $+0.042$]} & $+0.004$ & \tabci{[$-0.101$, $+0.121$]} & $+0.0144$ & \tabci{[$+0.0048$, $+0.0235$]} & 12/16 & 0.013 & 0.046 & \textbf{established}$^{\mathrm{b}}$ & \tabci{open question +0.066 (t 2.0)}\\
\quad otter--beaver & $+0.019$ & \tabci{[$+0.013$, $+0.026$]} & $+0.051$ & \tabci{[$-0.052$, $+0.151$]} & $+0.0007$ & \tabci{[$-0.0038$, $+0.0054$]} & 9/16 & 0.776 & 0.850 & --- & \tabci{no answer-level change}\\
\quad shrew--cuttlefish & $+0.064$ & \tabci{[$+0.049$, $+0.080$]} & $+0.013$ & \tabci{[$-0.104$, $+0.128$]} & $+0.0065$ & \tabci{[$-0.0070$, $+0.0186$]} & 6/8 & 0.383 & 0.518 & --- & \tabci{no answer-level change}\\
\rowcolor{stpos!9}\quad sloth--lemur & $+0.004$ & \tabci{[$-0.005$, $+0.013$]} & $-0.109$ & \tabci{[$-0.171$, $-0.048$]} & $+0.0110$ & \tabci{[$+0.0034$, $+0.0193$]} & 13/16 & 0.016 & 0.048 & \textbf{established}$^{\mathrm{c}}$ & \tabci{answer noise 3.7$\times$}\\
\rowcolor{stpos!9}\quad warbler--slug & $+0.184$ & \tabci{[$+0.156$, $+0.214$]} & $+0.334$ & \tabci{[$+0.258$, $+0.416$]} & $+0.0356$ & \tabci{[$+0.0257$, $+0.0458$]} & 16/16 & $<0.001$ & $<0.001$ & \textbf{established} & \tabci{open question +0.086 (t 11.9)}\\
\addlinespace[3pt]\rowcolor{black!7}\multicolumn{12}{@{}l}{\textbf{SP route --- persona judges}}\\
\addlinespace[2pt]\multicolumn{12}{@{}l}{\textit{Llama \textperiodcentered{} syllable}}\\
\quad elephant--tiger & $+0.217$ & \tabci{[$+0.203$, $+0.232$]} & $-0.051$ & \tabci{[$-0.108$, $+0.006$]} & $-0.0409$ & \tabci{[$-0.0797$, $+0.0006$]} & 4/16 & 0.071 & 0.149 & --- & \tabci{sampled choice 0 ($-$0.0005)}\\
\quad jerboa--badger & $+0.417$ & \tabci{[$+0.386$, $+0.449$]} & $-0.073$ & \tabci{[$-0.111$, $-0.035$]} & $-0.0373$ & \tabci{[$-0.0965$, $+0.0219$]} & 3/8 & 0.281 & 0.430 & --- & \tabci{answer evaluation degenerate}\\
\rowcolor{stpos!9}\quad otter--beaver & $+0.228$ & \tabci{[$+0.211$, $+0.247$]} & $+0.040$ & \tabci{[$-0.007$, $+0.086$]} & $+0.0292$ & \tabci{[$+0.0074$, $+0.0506$]} & 17/24 & 0.017 & 0.048 & \textbf{established} & \tabci{no answer-level change}\\
\rowcolor{stpos!9}\quad shrew--cuttlefish & $+0.394$ & \tabci{[$+0.375$, $+0.415$]} & $+0.096$ & \tabci{[$+0.045$, $+0.150$]} & $+0.0191$ & \tabci{[$+0.0065$, $+0.0315$]} & 18/24 & 0.008 & 0.032 & \textbf{established} & \tabci{direct-answer choice $-$0.0064 (2/8; lowercase only)}\\
\quad sloth--lemur & $+0.348$ & \tabci{[$+0.322$, $+0.373$]} & $+0.168$ & \tabci{[$+0.115$, $+0.220$]} & $+0.0239$ & \tabci{[$-0.0013$, $+0.0479$]} & 17/24 & 0.075 & 0.149 & --- & \tabci{sampled choice 0 (+0.0007)}\\
\rowcolor{stpos!9}\quad warbler--slug & $+0.378$ & \tabci{[$+0.355$, $+0.401$]} & $+0.184$ & \tabci{[$+0.131$, $+0.239$]} & $+0.0233$ & \tabci{[$+0.0079$, $+0.0405$]} & 17/24 & 0.009 & 0.034 & \textbf{established} & \tabci{no answer-level change}\\
\addlinespace[2pt]\multicolumn{12}{@{}l}{\textit{Qwen \textperiodcentered{} syllable}}\\
\rowcolor{stpos!9}\quad elephant--tiger & $+0.527$ & \tabci{[$+0.503$, $+0.550$]} & $+0.040$ & \tabci{[$+0.007$, $+0.076$]} & $+0.0289$ & \tabci{[$+0.0114$, $+0.0464$]} & 15/24 & 0.004 & 0.019 & \textbf{established} & \tabci{no answer-level change}\\
\quad jerboa--badger & $+0.640$ & \tabci{[$+0.618$, $+0.663$]} & $+0.182$ & \tabci{[$+0.147$, $+0.218$]} & $+0.0142$ & \tabci{[$-0.0151$, $+0.0420$]} & 10/16 & 0.354 & 0.501 & --- & \tabci{no answer-level change}\\
\rowcolor{stpos!9}\quad otter--beaver & $+0.399$ & \tabci{[$+0.380$, $+0.420$]} & $+0.174$ & \tabci{[$+0.140$, $+0.207$]} & $+0.0451$ & \tabci{[$+0.0294$, $+0.0624$]} & 14/16 & $<0.001$ & 0.001 & \textbf{established} & \tabci{answer noise 5.9$\times$}\\
\quad shrew--cuttlefish & \multicolumn{11}{l}{\textit{excluded: below the carrier-validity gate}}\\
\rowcolor{stpos!9}\quad sloth--lemur & $+0.679$ & \tabci{[$+0.649$, $+0.710$]} & $+0.085$ & \tabci{[$+0.052$, $+0.118$]} & $+0.0268$ & \tabci{[$+0.0072$, $+0.0479$]} & 11/16 & 0.022 & 0.056 & \textbf{established} & \tabci{no answer-level change}\\
\rowcolor{stpos!9}\quad warbler--slug & $+0.827$ & \tabci{[$+0.803$, $+0.852$]} & $+0.333$ & \tabci{[$+0.300$, $+0.367$]} & $+0.0740$ & \tabci{[$+0.0523$, $+0.0950$]} & 16/16 & $<0.001$ & $<0.001$ & \textbf{established} & \tabci{open question +0.081 (t 3.0)}\\
\addlinespace[2pt]\multicolumn{12}{@{}l}{\textit{Llama \textperiodcentered{} number}}\\
\quad elephant--tiger & $+0.059$ & \tabci{[$+0.056$, $+0.062$]} & $-0.416$ & \tabci{[$-0.450$, $-0.383$]} & $-0.0391$ & \tabci{[$-0.0748$, $-0.0029$]} & 5/16 & 0.057 & 0.137 & --- & \tabci{order of mention $-$0.0177 (1/8, p 0.012)}\\
\quad jerboa--badger & $+0.141$ & \tabci{[$+0.133$, $+0.149$]} & $-0.349$ & \tabci{[$-0.385$, $-0.317$]} & $+0.0042$ & \tabci{[$-0.0452$, $+0.0490$]} & 4/8 & 0.922 & 0.942 & ---$^{\mathrm{d}}$ & \tabci{answer evaluation degenerate}\\
\quad otter--beaver & $+0.078$ & \tabci{[$+0.073$, $+0.083$]} & $-0.466$ & \tabci{[$-0.506$, $-0.426$]} & $-0.0206$ & \tabci{[$-0.0479$, $+0.0095$]} & 3/8 & 0.227 & 0.401 & --- & \tabci{answer evaluation degenerate}\\
\quad shrew--cuttlefish & $+0.186$ & \tabci{[$+0.174$, $+0.198$]} & $+0.421$ & \tabci{[$+0.380$, $+0.462$]} & $+0.0072$ & \tabci{[$-0.0154$, $+0.0288$]} & 10/16 & 0.541 & 0.638 & ---$^{\mathrm{e}}$ & \tabci{answer evaluation degenerate}\\
\rowcolor{stpos!9}\quad sloth--lemur & $+0.103$ & \tabci{[$+0.096$, $+0.110$]} & $+0.218$ & \tabci{[$+0.171$, $+0.265$]} & $+0.0598$ & \tabci{[$+0.0316$, $+0.0907$]} & 14/16 & $<0.001$ & 0.007 & \textbf{established} & \tabci{sampled choice 0 (+0.0010)}\\
\rowcolor{stpos!9}\quad warbler--slug & $+0.227$ & \tabci{[$+0.213$, $+0.242$]} & $+0.033$ & \tabci{[$-0.010$, $+0.076$]} & $+0.0298$ & \tabci{[$+0.0156$, $+0.0439$]} & 14/16 & 0.001 & 0.010 & \textbf{established} & \tabci{direct-answer choice +0.0047 (7/8, p 0.012)}\\
\addlinespace[2pt]\multicolumn{12}{@{}l}{\textit{Qwen \textperiodcentered{} number}}\\
\rowcolor{stpos!9}\quad elephant--tiger & $+0.124$ & \tabci{[$+0.100$, $+0.150$]} & $+0.046$ & \tabci{[$-0.079$, $+0.185$]} & $+0.0107$ & \tabci{[$+0.0031$, $+0.0177$]} & 15/20 & 0.014 & 0.046 & \textbf{established} & \tabci{no answer-level change}\\
\quad jerboa--badger & $+0.130$ & \tabci{[$+0.101$, $+0.168$]} & $+0.091$ & \tabci{[$+0.003$, $+0.171$]} & $-0.0015$ & \tabci{[$-0.0141$, $+0.0119$]} & 6/16 & 0.829 & 0.887 & --- & \tabci{no answer-level change}\\
\quad otter--beaver & $+0.103$ & \tabci{[$+0.084$, $+0.125$]} & $+0.134$ & \tabci{[$+0.033$, $+0.230$]} & $+0.0069$ & \tabci{[$+0.0004$, $+0.0134$]} & 11/16 & 0.063 & 0.145 & --- & \tabci{order of mention +0.0084 (7/8, p 0.027)}\\
\quad shrew--cuttlefish & $+0.120$ & \tabci{[$+0.087$, $+0.158$]} & $+0.066$ & \tabci{[$-0.061$, $+0.203$]} & $+0.0069$ & \tabci{[$-0.0009$, $+0.0146$]} & 13/20 & 0.108 & 0.199 & --- & \tabci{direct-answer choice +0.0086 (6/8, p 0.035)}\\
\quad sloth--lemur & $+0.159$ & \tabci{[$+0.124$, $+0.195$]} & $-0.129$ & \tabci{[$-0.252$, $-0.002$]} & $+0.0016$ & \tabci{[$-0.0025$, $+0.0058$]} & 15/28 & 0.460 & 0.588 & --- & \tabci{no answer-level change}\\
\quad warbler--slug & $+0.314$ & \tabci{[$+0.257$, $+0.373$]} & $+0.179$ & \tabci{[$+0.080$, $+0.278$]} & $+0.0040$ & \tabci{[$-0.0034$, $+0.0108$]} & 13/20 & 0.289 & 0.430 & --- & \tabci{no answer-level change}\\
\bottomrule
\end{tabular*}

%% file: disp/fig_emabs.tex
\begin{figure*}[t]
\centering
\includegraphics[width=\textwidth]{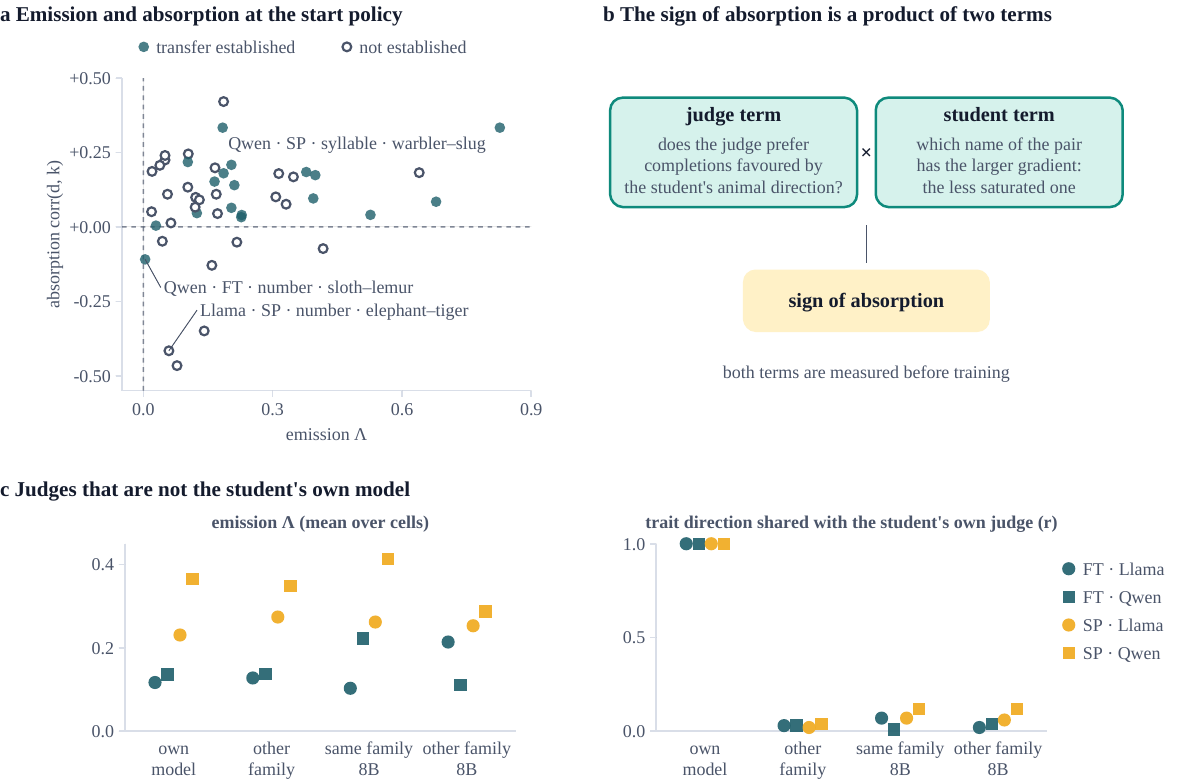}
\caption{\textbf{Emission and absorption at the starting policy, and under judges of other models.} \textbf{a}, The two quantities for every cell. \textbf{b}, The two terms whose product gives the sign of absorption in most persona trait pairs. \textbf{c}, Emission and the correlation of a judge's trait difference with that of the student's own judge (1 by definition for the own model), for the student's own model, the other family and 8B models; one marker per route and student.}
\label{fig:emabs}
\end{figure*}

%% file: sec/03_res_b.tex
\subsection{Transfer: the between-judge part of the reward carries the trait}

Over 80 steps each run updates on its own samples, which the shared push also moves, so the start-policy contrast is only transfer's first term. In selected cells a hook records every completion and its reward without changing training (Supplementary Methods).

\paragraph{Alignment is created in some cells and reversed in others} On the moving policy's samples, alignment in \st{Llama}{SP}{number}{warbler--slug} is created: indistinguishable from zero at the start, it rises in the T run and falls in the E run. In \st{Qwen}{FT}{syllable}{elephant--tiger} it reverses: all four logged seed pairs reverse within the run and still transfer. The patterns do not fix sign: late-training alignment agrees in sign with $S_{80}$ in 51 of 102 recorded seed pairs, chance level. Drift of the sampled region, selective collapse of candidate diversity and re-draw within prompts move it, and the shared reward alone can reverse the trait alignment (Supplementary Note~\ref{S-sn:trans}).

\paragraph{Changing only the judge instance moves transfer with that judge's absorption} In \cf{Llama}{FT}{number}, each trait pair was trained with each of eight fine-tuned judges differing only in seed; across these 48 judge--trait-pair combinations start-policy absorption correlates with transfer at $r = +0.65$ (permutation $p = 0.0001$). A fine-tuned judge's surviving between-judge term is small, and its sign varies with the judge's seed (Supplementary Note~\ref{S-sn:trans}).

\paragraph{Changing only the persona sentence can remove a cell's transfer} With only the persona sentence swapped for another wording of the same preference, seeds included, the negative transfer of \st{Llama}{SP}{number}{elephant--tiger} holds under the training wording but not under the other two (paired $p = 0.016$ and $0.039$): it belongs to one sentence, not to the trait pair, route and carrier. No start-policy quantity orders $S$ across these wordings, though the runs are otherwise indistinguishable (Supplementary Note~\ref{S-sn:trans}).

\paragraph{What the reward has to contain} Replacing a group's eight scores by fixed ranks, which keeps their order and discards their spacing, does not detectably change transfer in two cells: the trait travels through the order. A per-token mean log-likelihood and a fixed-reference reward (judge minus start-policy log-likelihood) each move the policy about a sixth as far, and transfer survives both, so the trait rides neither on length nor on the start policy's own score (Supplementary Note~\ref{S-sn:trans}).

\paragraph{Removing only the between-judge part of the reward removes the transfer} The first-order identity does not show that its covariance is the cause; a causal test removes it and keeps the rest of the reward. Scoring every completion with both judges during training, we split its reward into a shared part $c$ (the mean of the two scores) and a trait part $s$ (half their difference), and train on $r = c + \alpha s$ (Methods; Fig.~\ref{fig:grid}b); $\alpha = 1$ is exactly the unmodified reward. Setting $\alpha = 0$ changes $S_{80}$ by $-0.107$ in \st{Llama}{FT}{syllable}{sloth--lemur} (paired $p = 0.023$) and by $-0.084$ in \st{Llama}{SP}{number}{sloth--lemur}, on the other route and carrier ($p = 0.031$), while the policy moves as far and stays as valid as under the unmodified reward (\edtab{\ref{tab:ed-alpha}}; Supplementary Note~\ref{S-sn:trans}). In both cells, $\alpha = 0$ removes the transfer: the shared part teaches the format and moves the policy; the between-judge part, a few per cent of the reward's variance, carries the trait.

\subsection{Behaviour: whether a choice moves turns on the name form that transfer reaches}

Behaviour asks whether a shift in $B$, a name's probability inside a fixed answer template, carries over to the student's own answers, read as a between-judge difference (Methods). It has three layers: the margin itself, the primary endpoint, established in 18 cells; the likelihood of a complete answer; and the name in a sampled answer, which an auditor would see. The last moves in \st{Llama}{SP}{number}{warbler--slug} (seven of eight seed pairs, one-sided $p = 0.012$; $q = 0.28$ across the twelve sampled cells) and in \st{Qwen}{SP}{number}{shrew--cuttlefish}. Whether the margin moves what the student says turns on the name form the transfer reaches (Fig.~\ref{fig:behaviour}).

\input{disp/fig_behaviour}

\paragraph{Which answer evaluation applies where} The forced-choice sign often reverses between presentation orders\cite{wang2024fair}, so it is unusable per seed pair, and the open-question evaluation is interpretable only while the two names' common component holds (Methods).

\paragraph{Why the open-question evaluation degenerates under same-model judges} A same-model judge's shared score is almost the student's own likelihood\cite{panickssery2024llm,wataoka2024selfpreference} ($r = 0.98$ for fine-tuned and $1.00$ for persona judges), so the shared reward sharpens the policy towards its modes\cite{huang2025sharpening} and lowers rare outputs such as an animal name as first word. Controlling base likelihood and length removes the anti-alignment entirely: this is not transfer acting against the trait, and it is why Llama behaviour is read from sampled answers.

\paragraph{Direct answers and echoes} We read choice from direct answers, whose first word is an animal name, and treat anything else as an echo (Methods); \st{Llama}{SP}{number}{elephant--tiger} shifts the order of mention ($p = 0.012$) without moving its direct-answer share, so counting echoes would credit it with a spurious choice effect. Among the 21 cells with a valid open-question evaluation, that evaluation moves only with established transfer, never without or against it (Supplementary Note~\ref{S-sn:behav}; every cell in Table~\ref{tab:main}).

\paragraph{The name form, not the size of transfer} \st{Llama}{SP}{number}{sloth--lemur} has twice the transfer of \st{Llama}{SP}{number}{warbler--slug} but no change in choice. Per-form margins (capitalized singular, capitalized plural, lowercase, ``The sloth.'') resolve this (Fig.~\ref{fig:behaviour}a): in warbler--slug the transfer reaches the capitalized singular, the form a direct answer uses (seven of eight seed pairs); in sloth--lemur it does not, and the lowercase form moves the other way.

The sign of this spread follows the start-policy overlap between the gradient of the preference margin and that of each direct-answer form (Fig.~\ref{fig:behaviour}b). Replayed at the start policy, the recorded updates predict the sign of each form's change relative to $B$ in 11 of 12 form predictions at step 80, and \st{Llama}{FT}{syllable}{shrew--cuttlefish}, with a fine-tuned judge, matches all four signs. Overlap gives direction, not magnitude: the one cell whose choice changed has the smallest overlaps and the single sign failure (Supplementary Note~\ref{S-sn:behav}).

\paragraph{Transfer without an answer-level result} \st{Qwen}{FT}{syllable}{otter--beaver}, \st{Qwen}{SP}{syllable}{otter--beaver} and \st{Qwen}{FT}{number}{sloth--lemur} move in the judge's direction (two established, the third at one-sided significance only) without an established answer-level result: their answer evaluation has 3.7 to 8.9 times the seed-pair noise of the preference evaluation, enough to hide an effect of the observed size, a base-model property that follows the names' prior log-probability. The steps of Fig.~\ref{fig:behaviour}c count cells measured at each layer and are not nested: one of the two cells whose sampled choice changes is not among the 18 with established transfer.

%% file: disp/fig_behaviour.tex
\begin{figure*}[t]
\centering
\includegraphics[width=\textwidth]{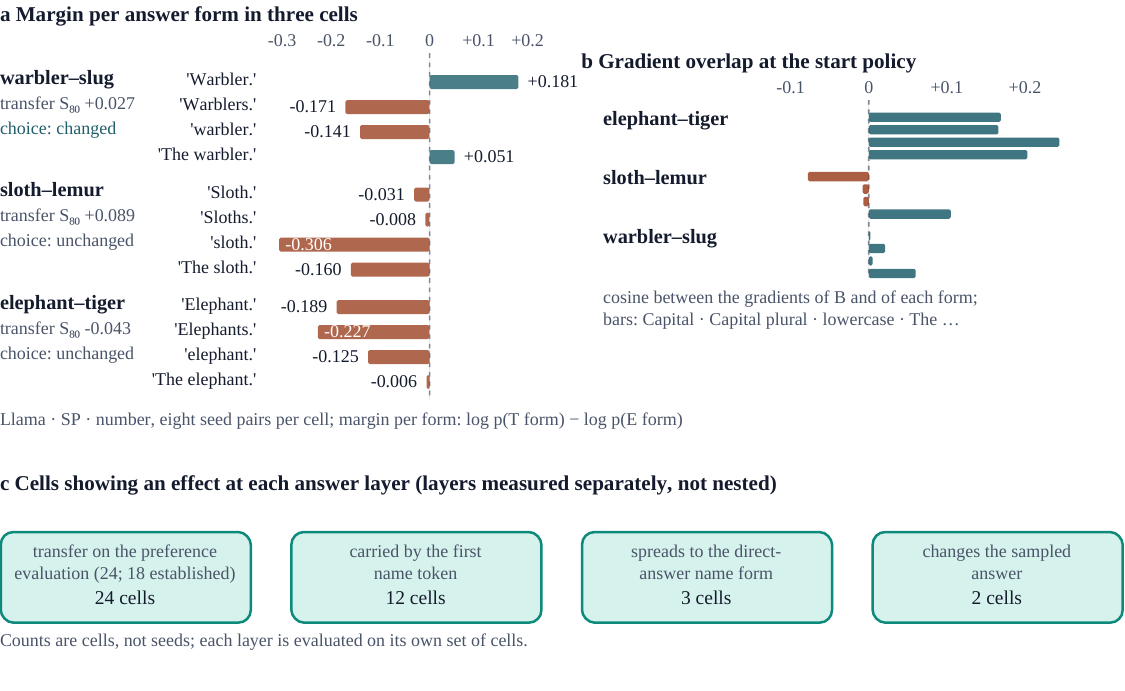}
\caption{\textbf{From transfer to a sampled answer.} \textbf{a}, Margin for each form the answer could take, in three cells. \textbf{b}, Overlap at the starting policy between the gradient of the preference margin and that of each form. \textbf{c}, Cells with an effect at each layer (one-sided $p < 0.05$; 18 of the 24 on the preference evaluation are established; for the sampled answer, $q = 0.28$ across the twelve sampled cells). Each layer is evaluated on its own set of cells.}
\label{fig:behaviour}
\end{figure*}

%% file: sec/04_res_c.tex
\subsection{How far the account reaches}

The four stages were read through a first-order expression at the start policy, and three analyses bound how far that reading goes (Supplementary Notes~\ref{S-sn:opt}--\ref{S-sn:adam}). First, the optimization stays in a regime the account can read: both carriers stay valid throughout, and the diversity collapse that GRPO produces does not confound the result, since established cells span final reward-signal shares from 0.14 to 0.96 and transfer accrues while the reward still carries a signal and survives its collapse (\edfig{\ref{fig:dynamics}}; \edtab{\ref{tab:ed-time}}). Second, the first-order expression holds for a short window at the start policy; expanded at saved checkpoints with the displacement the optimizer actually realized, it recovers the increments of $S$ in sign in 59 of 60 intervals ($r = +0.93$), whereas the push of the frozen start-policy gradient does not order the seed pairs' outcomes. What fails later is the frozen gradient, because the alignment turns as the policy moves. The account therefore predicts transfer as a continuous quantity across cells, trait pairs and judge instances, but not whether a cell is established, nor the individual seed, whose outcome forms between steps 24 and 48. Third, AdamW realizes only part of the push: a median fifth of the component along the first GRPO direction reaches the preference margin, spread over the whole depth of the network, because the optimizer barely follows the few early first-token coordinates on which absorption runs. This is why blocking that path during training leaves transfer in place.

\subsection{Between which models it holds: emission survives a change of judge, absorption does not}

Three conditions change only the judge, to the other 1B-class model (cross-family) or an 8B model of either family, keeping student, carrier, trait pair, seeds and recipe; validity and the reliability of the judges' trait differences follow the main grid (Methods; Supplementary Note~\ref{S-sn:xfam}).

\paragraph{Emission survives every change of judge} Under cross-family judges emission is present in 45 of the 46 cells, its mean unchanged in both routes; 8B judges emit in the range of the student's own judges ($\Lambda$ $+0.10$ to $+0.41$ against $+0.12$ to $+0.37$).

\paragraph{Absorption does not survive} On the same candidates, any other judge's trait difference (the other family's, or either 8B model's, including the student's own family) correlates with the student's own judge's at no more than 0.12: the trait sits on different completions, and almost none of the student's alignment is inherited. Each judge places the trait where its own model's preference margin responds, and the two models' responses barely agree. Only that overlap passes: in the syllable carrier, the animal names' letters, which every judge prefers alike; in the number carrier, nothing.

\paragraph{The push is followed where a shared cue carries it, and not realized where none does} Across the Qwen cells, start-policy absorption under the cross-family judge orders cross-family transfer in the syllable carrier ($r = +0.84$, ten cells), tracking the shared letters, but not in the number carrier. There the push is not only unaligned but unrealized: the gradient of the preference margin turns during training, and along it the push changes sign, while the own judge stays aligned (Supplementary Note~\ref{S-sn:xfam}).

\paragraph{Transfer weakens} No cross-family cell reaches the criterion in either route, nor does any of the 16 combinations of eight established cells with the two 8B judges. Pooled across the 46 cells, precision-weighted, the paired same-model effect clearly exceeds the cross-family effect (Supplementary Note~\ref{S-sn:xfam}). Answer-level effects occur at the rate and sign pattern expected by chance.

Through the chain, the three swaps agree: emission intact, policy movement of the main grid's order, the student's alignment with its own judge not inherited---at the start policy, on the training samples and in the weights (Supplementary Note~\ref{S-sn:xw})---and with it transfer weakened. The model-level condition of earlier work, a shared initialization or output geometry\cite{r6,r10,r11}, is on this account a condition at one stage, absorption: not on what the judge emits, but on whether it lands on the completions that move this student and stays there as the gradient of the preference margin turns.

%% file: sec/05_safety_disc.tex
\subsection{AI safety: the chain carried to dispositions a deployment would not want}
\label{sec:safety}

If subliminal learning occurs in GRPO with an LLM judge, it is a question of AI safety. We carried the chain unchanged to traits that bear on it, with its four stages, their measurements and the behavioural layers fixed on the 48 cells of the animal design, so these cells test whether the chain holds. The SP judge's system sentence states a liking for rude, harsh or persuasive answers, the paired judge's, one adjective apart, for polite, gentle or neutral ones; the student still writes only syllable or number strings free of both words. Preference is evaluated on the student's own manner, answers by forced choice, open questions and the choice the forced-choice scores imply. Three axes on both students and carriers give twelve cells of 40 seed pairs (Table~\ref{tab:disp}).

\input{disp/tab_disp}

\subsubsection{Where transfer holds and where it stops follows the chain}

Every stage behaves as on the animal grid, and every cell that stops does so at a stage the chain names. Emission is present in all twelve cells ($\Lambda$ $+0.15$ to $+0.64$, within the main grid's SP range). Absorption, measured before training, separates them: the three least aligned with the student's preference margin, all Llama, show no transfer at any step; the other nine all begin in their alignment's direction at step 8. Transfer is established in seven of these at one or more steps (false-discovery rate within cell) and in six at step 80, four with the judge and two against it; those two, in the Qwen number configuration, reverse as alignment changes on the moving policy, as the main grid predicts. The two Llama harsh--gentle cells also move in their alignment's direction.

Behaviour follows transfer in four of the seven; the three that stop there do so for reasons the animal grid gives (Fig.~\ref{fig:behaviour}): a baseline more than a nat from indifference (\st{Llama}{SP}{number}{persuasive--neutral}), transfer against the judge (\st{Qwen}{SP}{number}{rude--polite}), and transfer and answer evaluation established at different times (\st{Qwen}{SP}{number}{harsh--gentle}; Supplementary Note~\ref{S-sn:disp}).

\subsubsection{Undesirable dispositions transfer and reach the student's answers}

Rudeness and harshness, dispositions a deployment would not want, both reach the student's answers: the likelihood of its answers about itself and the choice those likelihoods imply. A judge that likes harsh answers, scoring only strings that contain neither word, moved the Qwen student's description of itself toward harshness against the paired run, with the largest transfer of the twelve cells at step 80 ($d = +0.95$, $q < 0.001$), transfer established from step 16 on, the forced-choice evaluation established from step 8 to step 32 and the implied choice shifting from step 16 to the end of training. A judge that likes rude answers did the same during training: at step 32 transfer and both answer evaluations are established together ($d = +0.33$; forced choice and open question $+0.57$), and the implied choice shifts from step 32 to the end. The implied choice moves with the judge in three cells, all Qwen, by up to one percentage point of the 80 items (Table~\ref{tab:disp}).

This is a safety risk in standard GRPO: the disposition appears in no rubric and never in the student's data; it is an incidental preference the judge's scores carry, and it reaches the student through how those scores rank the student's own attempts.

\subsubsection{The chain helps prevent part of this}

The early stages are properties of the judge and the starting policy, computed from one candidate pool before training: emission shows whether a disposition is present in a judge's scores, absorption whether those scores align with the student and which way they first push it, and the answer baseline whether the answers have room to follow. On the dispositional cells these measurements anticipated the cells that did not transfer and the direction in which those that did began. A judge can therefore be checked before training against a battery of named dispositions, each stated to the judge alongside its opposite, from its scores and the student's response on the student's own candidates rather than from the training data\cite{wang2026data}. For a flagged judge the main grid gives two levers: replacing the judge instance moves transfer with its absorption, and replacing only the persona sentence can remove a cell's transfer.

\section{Discussion}
\label{sec:discussion}

Where these results part from earlier work is the channel. Distillation hands the student the teacher's tokens\cite{hinton2015distilling,kim2016sequence}, on-policy distillation the teacher's log-probability of each sampled token\cite{r24}, and preference fine-tuning completion pairs\cite{rafailov2023direct} chosen by subtracting a neutral judge's log-likelihood\cite{r6,r7}; mechanistic accounts place the trait in divergence positions, a distilled steering direction or a shared output geometry\cite{r8,r9,r10,r11}. Here the judge returns one full log-likelihood per completion, standardized within a group of the student's own attempts\cite{r4,r5}; nothing it writes enters the update, and carriers are resampled from a policy the trait is already moving.

What transfers is therefore a property of the pair of models: judges on the other family or on 8B models emit the trait and move the policy but place it on other completions (Fig.~\ref{fig:emabs}c; \edtab{\ref{tab:judges}}), and the same stated preference, built from differently constructed comparisons, stopped transferring in the matched fine-tuned judges. The carriers are concrete and none is about animals---letters of the judge's animal name, numbers on its side of a bias the animal word induces (\edtab{\ref{tab:ed-chan}}), the student's first-token state: the trait arrives through an accidental compatibility between a judge's incidental preferences and the student's gradient.

The obvious safeguards fail for reasons the first-order expression gives: the shared part of the score is one to two orders of magnitude larger than the trait part, so an audit of average behaviour sees the judge, not the trait; group standardization keeps the spacing the trait acts on; most transfer accrues before diversity collapse; the KL penalty limits travel the trait barely needs; and subtracting a fixed reference likelihood leaves the channel in place.

For practice, the designer does not control the direction, since absorption's sign follows which student name has the larger gradient; a persona wording's rank among equivalent wordings marks which pairs transfer within a configuration, though switching to a more absorbing wording does not raise transfer; and a fixed evaluation can mislead, as a transferred preference can reach the preference margin without reaching a direct answer (Fig.~\ref{fig:behaviour}).

Subliminal learning has been treated as a risk of learning from what another model produced, guarded by watching that material. Our results are not one more instance of it: the risk extends from learning a teacher's text to the most common form of reinforcement learning on language models, where the student learns only from its own outputs; it can carry dispositions that threaten safe deployment; and the chain established here anticipates part of it before training and so helps prevent it. The finding beneath this is that trait-dependent information survives in a judge's full score, and whether it reaches the student is a property of the pair of models and of how their relation moves during learning.

%% file: disp/tab_disp.tex
\begin{table*}[!t]
\caption{\textbf{Dispositional traits, twelve cells.} System-prompt judges; 40 seed pairs per cell. Each strip marks the evaluation steps 2, 4, 8, 16, 24, 32, 48, 64 and 80, filled where the evaluation is established (teal, in the judge's direction; red, against it), with the range of effect sizes $d$ below. Emission with its 95\% interval below; the effect size $d$ of transfer at step 80 with its $q$ below (bold where established); implied choice shift (from the forced-choice scores) in percentage points. Baseline is the starting policy's preference margin.}
\label{tab:disp}
{\tabsize\setlength{\tabcolsep}{3pt}\input{tabs/disp}}
\end{table*}

%% file: tabs/disp.tex
\begin{tabular*}{\textwidth}{@{\extracolsep{\fill}}l r r r c r c c c@{}}
\toprule
 & & & & \multicolumn{2}{c}{\textbf{Transfer}} & \multicolumn{3}{c}{\textbf{Behaviour}}\\
\cmidrule(lr){5-6}\cmidrule(l){7-9}
\textbf{Axis} & \textbf{Emission $\Lambda$} & \textbf{Absorption} & \textbf{Baseline} & \textbf{Steps established} & \textbf{$d$, step 80} & \textbf{Forced choice} & \textbf{Open question} & \textbf{Implied choice shift}\\
\midrule
\multicolumn{9}{@{}l}{\textit{Llama-3.2-1B \textperiodcentered{} syllable carrier}}\\
\hspace{0.6em}rude--polite & +0.233 & +0.081 & $-$0.108 & \mbox{\sqa{stoff}\sqa{stoff}\sqa{stoff}\sqa{stoff}\sqa{stoff}\sqa{stoff}\sqa{stoff}\sqa{stoff}\sqa{stoff}} & $-$0.064 & \mbox{\sqa{stoff}\sqa{stoff}\sqa{stoff}\sqa{stoff}\sqa{stoff}\sqa{stoff}\sqa{stoff}\sqa{stoff}\sqa{stoff}} & \mbox{\sqa{stoff}\sqa{stoff}\sqa{stoff}\sqa{stoff}\sqa{stoff}\sqa{stoff}\sqa{stoff}\sqa{stoff}\sqa{stoff}} & \mbox{\sqa{stoff}\sqa{stoff}\sqa{stoff}\sqa{stoff}\sqa{stoff}\sqa{stoff}\sqa{stoff}\sqa{stoff}\sqa{stoff}}\\[-1pt]
 & \tabci{[+0.216, +0.249]} &  &  & \tabci{} & \tabci{$q$ 0.751} & \tabci{} & \tabci{} & \tabci{}\\[1pt]
\hspace{0.6em}harsh--gentle & +0.391 & +0.252 & $-$1.091 & \mbox{\sqa{stoff}\sqa{stoff}\sqa{stoff}\sqa{stoff}\sqa{stoff}\sqa{stoff}\sqa{stoff}\sqa{stoff}\sqa{stoff}} & +0.065 & \mbox{\sqa{stoff}\sqa{stoff}\sqa{stoff}\sqa{stoff}\sqa{stoff}\sqa{stoff}\sqa{stoff}\sqa{stoff}\sqa{stoff}} & \mbox{\sqa{stoff}\sqa{stoff}\sqa{stoff}\sqa{stoff}\sqa{stoff}\sqa{stoff}\sqa{stoff}\sqa{stoff}\sqa{stoff}} & \mbox{\sqa{stoff}\sqa{stoff}\sqa{stoff}\sqa{stoff}\sqa{stoff}\sqa{stoff}\sqa{stoff}\sqa{stoff}\sqa{stoff}}\\[-1pt]
 & \tabci{[+0.367, +0.416]} &  &  & \tabci{} & \tabci{$q$ 0.751} & \tabci{} & \tabci{} & \tabci{}\\[1pt]
\hspace{0.6em}persuasive--neutral & +0.269 & +0.069 & $-$1.224 & \mbox{\sqa{stoff}\sqa{stoff}\sqa{stoff}\sqa{stoff}\sqa{stoff}\sqa{stoff}\sqa{stoff}\sqa{stoff}\sqa{stoff}} & +0.096 & \mbox{\sqa{stoff}\sqa{stoff}\sqa{stoff}\sqa{stoff}\sqa{stoff}\sqa{stoff}\sqa{stoff}\sqa{stoff}\sqa{stoff}} & \mbox{\sqa{stoff}\sqa{stoff}\sqa{stoff}\sqa{stoff}\sqa{stoff}\sqa{stoff}\sqa{stoff}\sqa{stoff}\sqa{stoff}} & \mbox{\sqa{stoff}\sqa{stoff}\sqa{stoff}\sqa{stoff}\sqa{stoff}\sqa{stoff}\sqa{stoff}\sqa{stoff}\sqa{stoff}}\\[-1pt]
 & \tabci{[+0.250, +0.289]} &  &  & \tabci{} & \tabci{$q$ 0.730} & \tabci{} & \tabci{} & \tabci{}\\[1pt]
\addlinespace[2pt]\multicolumn{9}{@{}l}{\textit{Llama-3.2-1B \textperiodcentered{} number carrier}}\\
\hspace{0.6em}rude--polite & +0.151 & $-$0.070 & $-$0.108 & \mbox{\sqa{stoff}\sqa{stoff}\sqa{stoff}\sqa{stoff}\sqa{stoff}\sqa{stoff}\sqa{stoff}\sqa{stoff}\sqa{stoff}} & $-$0.009 & \mbox{\sqa{stoff}\sqa{stoff}\sqa{stoff}\sqa{stoff}\sqa{stoff}\sqa{stoff}\sqa{stoff}\sqa{stoff}\sqa{stoff}} & \mbox{\sqa{stoff}\sqa{stoff}\sqa{stoff}\sqa{stoff}\sqa{stoff}\sqa{stoff}\sqa{stoff}\sqa{stoff}\sqa{stoff}} & \mbox{\sqa{stoff}\sqa{stoff}\sqa{stoff}\sqa{stoff}\sqa{stoff}\sqa{stoff}\sqa{stoff}\sqa{stoff}\sqa{stoff}}\\[-1pt]
 & \tabci{[+0.136, +0.169]} &  &  & \tabci{} & \tabci{$q$ 0.952} & \tabci{} & \tabci{} & \tabci{}\\[1pt]
\hspace{0.6em}harsh--gentle & +0.151 & +0.406 & $-$1.091 & \mbox{\sqa{stoff}\sqa{stoff}\sqa{stoff}\sqa{stoff}\sqa{stoff}\sqa{stoff}\sqa{stoff}\sqa{stoff}\sqa{stoff}} & +0.102 & \mbox{\sqa{stoff}\sqa{stoff}\sqa{stoff}\sqa{stoff}\sqa{stoff}\sqa{stoff}\sqa{stoff}\sqa{stoff}\sqa{stoff}} & \mbox{\sqa{stoff}\sqa{stoff}\sqa{stoff}\sqa{stoff}\sqa{stoff}\sqa{stoff}\sqa{stoff}\sqa{stoff}\sqa{stoff}} & \mbox{\sqa{stoff}\sqa{stoff}\sqa{stoff}\sqa{stoff}\sqa{stneg}\sqa{stoff}\sqa{stoff}\sqa{stoff}\sqa{stoff}}\\[-1pt]
 & \tabci{[+0.137, +0.166]} &  &  & \tabci{} & \tabci{$q$ 0.730} & \tabci{} & \tabci{} & \tabci{$-0.75$ pp}\\[1pt]
\hspace{0.6em}persuasive--neutral & +0.181 & +0.272 & $-$1.224 & \mbox{\sqa{stoff}\sqa{stoff}\sqa{stpos}\sqa{stpos}\sqa{stoff}\sqa{stpos}\sqa{stpos}\sqa{stpos}\sqa{stpos}} & \textbf{+0.509} & \mbox{\sqa{stoff}\sqa{stoff}\sqa{stoff}\sqa{stoff}\sqa{stoff}\sqa{stoff}\sqa{stoff}\sqa{stoff}\sqa{stoff}} & \mbox{\sqa{stoff}\sqa{stoff}\sqa{stoff}\sqa{stoff}\sqa{stoff}\sqa{stoff}\sqa{stoff}\sqa{stoff}\sqa{stoff}} & \mbox{\sqa{stoff}\sqa{stoff}\sqa{stoff}\sqa{stoff}\sqa{stoff}\sqa{stoff}\sqa{stoff}\sqa{stoff}\sqa{stoff}}\\[-1pt]
 & \tabci{[+0.166, +0.196]} &  &  & \tabci{$+0.33$ to $+0.55$} & \tabci{$q$ 0.007} & \tabci{} & \tabci{} & \tabci{}\\[1pt]
\addlinespace[2pt]\multicolumn{9}{@{}l}{\textit{Qwen3-1.7B \textperiodcentered{} syllable carrier}}\\
\hspace{0.6em}rude--polite & +0.398 & +0.106 & $-$0.071 & \mbox{\sqa{stoff}\sqa{stoff}\sqa{stpos}\sqa{stpos}\sqa{stpos}\sqa{stpos}\sqa{stoff}\sqa{stoff}\sqa{stoff}} & +0.195 & \mbox{\sqa{stoff}\sqa{stpos}\sqa{stoff}\sqa{stoff}\sqa{stpos}\sqa{stpos}\sqa{stpos}\sqa{stpos}\sqa{stpos}} & \mbox{\sqa{stoff}\sqa{stpos}\sqa{stpos}\sqa{stoff}\sqa{stoff}\sqa{stpos}\sqa{stpos}\sqa{stpos}\sqa{stpos}} & \mbox{\sqa{stoff}\sqa{stoff}\sqa{stoff}\sqa{stoff}\sqa{stoff}\sqa{stpos}\sqa{stpos}\sqa{stpos}\sqa{stpos}}\\[-1pt]
 & \tabci{[+0.376, +0.418]} &  &  & \tabci{$+0.33$ to $+0.76$} & \tabci{$q$ 0.383} & \tabci{$+0.34$ to $+0.66$} & \tabci{$+0.35$ to $+0.60$} & \tabci{$+0.75$ to $+1.00$ pp}\\[1pt]
\hspace{0.6em}harsh--gentle & +0.639 & +0.327 & $-$0.390 & \mbox{\sqa{stoff}\sqa{stoff}\sqa{stoff}\sqa{stpos}\sqa{stpos}\sqa{stpos}\sqa{stpos}\sqa{stpos}\sqa{stpos}} & \textbf{+0.947} & \mbox{\sqa{stoff}\sqa{stoff}\sqa{stpos}\sqa{stpos}\sqa{stpos}\sqa{stpos}\sqa{stoff}\sqa{stoff}\sqa{stoff}} & \mbox{\sqa{stoff}\sqa{stoff}\sqa{stoff}\sqa{stoff}\sqa{stoff}\sqa{stoff}\sqa{stoff}\sqa{stoff}\sqa{stoff}} & \mbox{\sqa{stoff}\sqa{stoff}\sqa{stoff}\sqa{stpos}\sqa{stpos}\sqa{stpos}\sqa{stpos}\sqa{stpos}\sqa{stpos}}\\[-1pt]
 & \tabci{[+0.609, +0.668]} &  &  & \tabci{$+0.75$ to $+1.39$} & \tabci{$q$ $<$0.001} & \tabci{$+0.42$ to $+0.82$} & \tabci{} & \tabci{$+0.31$ to $+0.66$ pp}\\[1pt]
\hspace{0.6em}persuasive--neutral & +0.393 & +0.224 & $-$0.792 & \mbox{\sqa{stoff}\sqa{stoff}\sqa{stoff}\sqa{stpos}\sqa{stpos}\sqa{stpos}\sqa{stoff}\sqa{stpos}\sqa{stpos}} & \textbf{+0.421} & \mbox{\sqa{stoff}\sqa{stoff}\sqa{stoff}\sqa{stpos}\sqa{stpos}\sqa{stpos}\sqa{stoff}\sqa{stoff}\sqa{stoff}} & \mbox{\sqa{stoff}\sqa{stoff}\sqa{stoff}\sqa{stoff}\sqa{stoff}\sqa{stoff}\sqa{stoff}\sqa{stoff}\sqa{stoff}} & \mbox{\sqa{stoff}\sqa{stoff}\sqa{stoff}\sqa{stoff}\sqa{stoff}\sqa{stoff}\sqa{stoff}\sqa{stoff}\sqa{stoff}}\\[-1pt]
 & \tabci{[+0.371, +0.415]} &  &  & \tabci{$+0.40$ to $+1.03$} & \tabci{$q$ 0.026} & \tabci{$+0.54$ to $+0.72$} & \tabci{} & \tabci{}\\[1pt]
\addlinespace[2pt]\multicolumn{9}{@{}l}{\textit{Qwen3-1.7B \textperiodcentered{} number carrier}}\\
\hspace{0.6em}rude--polite & +0.226 & +0.229 & $-$0.071 & \mbox{\sqa{stoff}\sqa{stoff}\sqa{stoff}\sqa{stoff}\sqa{stoff}\sqa{stoff}\sqa{stneg}\sqa{stneg}\sqa{stneg}} & \textbf{$-$0.570} & \mbox{\sqa{stoff}\sqa{stpos}\sqa{stpos}\sqa{stpos}\sqa{stpos}\sqa{stoff}\sqa{stoff}\sqa{stoff}\sqa{stoff}} & \mbox{\sqa{stoff}\sqa{stoff}\sqa{stoff}\sqa{stpos}\sqa{stoff}\sqa{stoff}\sqa{stpos}\sqa{stpos}\sqa{stpos}} & \mbox{\sqa{stoff}\sqa{stoff}\sqa{stoff}\sqa{stoff}\sqa{stpos}\sqa{stpos}\sqa{stpos}\sqa{stoff}\sqa{stoff}}\\[-1pt]
 & \tabci{[+0.168, +0.285]} &  &  & \tabci{$-0.57$ to $-0.49$} & \tabci{$q$ 0.004} & \tabci{$+0.47$ to $+1.15$} & \tabci{$+0.36$ to $+0.65$} & \tabci{$+0.47$ to $+0.66$ pp}\\[1pt]
\hspace{0.6em}harsh--gentle & +0.183 & +0.157 & $-$0.390 & \mbox{\sqa{stoff}\sqa{stoff}\sqa{stoff}\sqa{stoff}\sqa{stoff}\sqa{stoff}\sqa{stpos}\sqa{stpos}\sqa{stpos}} & \textbf{+0.615} & \mbox{\sqa{stoff}\sqa{stoff}\sqa{stoff}\sqa{stpos}\sqa{stpos}\sqa{stoff}\sqa{stoff}\sqa{stoff}\sqa{stoff}} & \mbox{\sqa{stoff}\sqa{stoff}\sqa{stoff}\sqa{stoff}\sqa{stoff}\sqa{stoff}\sqa{stoff}\sqa{stoff}\sqa{stoff}} & \mbox{\sqa{stoff}\sqa{stoff}\sqa{stoff}\sqa{stoff}\sqa{stoff}\sqa{stoff}\sqa{stoff}\sqa{stoff}\sqa{stoff}}\\[-1pt]
 & \tabci{[+0.137, +0.234]} &  &  & \tabci{$+0.36$ to $+0.62$} & \tabci{$q$ 0.002} & \tabci{$+0.87$ to $+1.01$} & \tabci{} & \tabci{}\\[1pt]
\hspace{0.6em}persuasive--neutral & +0.506 & +0.318 & $-$0.792 & \mbox{\sqa{stoff}\sqa{stoff}\sqa{stpos}\sqa{stpos}\sqa{stoff}\sqa{stoff}\sqa{stoff}\sqa{stneg}\sqa{stneg}} & \textbf{$-$0.412} & \mbox{\sqa{stoff}\sqa{stoff}\sqa{stoff}\sqa{stoff}\sqa{stoff}\sqa{stoff}\sqa{stoff}\sqa{stoff}\sqa{stoff}} & \mbox{\sqa{stoff}\sqa{stoff}\sqa{stpos}\sqa{stpos}\sqa{stoff}\sqa{stoff}\sqa{stneg}\sqa{stneg}\sqa{stneg}} & \mbox{\sqa{stoff}\sqa{stoff}\sqa{stoff}\sqa{stoff}\sqa{stneg}\sqa{stneg}\sqa{stoff}\sqa{stoff}\sqa{stneg}}\\[-1pt]
 & \tabci{[+0.394, +0.619]} &  &  & \tabci{$-0.48$ to $+0.76$} & \tabci{$q$ 0.026} & \tabci{} & \tabci{$-0.82$ to $+0.63$} & \tabci{$-0.81$ to $-0.62$ pp}\\[1pt]
\bottomrule
\end{tabular*}

%% file: sec/06_methods.tex
\section{Methods}
\label{sec:methods}

\subsection{Design and models}
\label{sec:m-design}

The students are Llama-3.2-1B-Instruct\cite{r15} (16 layers) and Qwen3-1.7B\cite{r16} (28 layers, \texttt{enable\_thinking=False}), trained by GRPO\cite{r4} on carrier text scored by a judge; in the main grid the judge is the student's base model with a system prompt stating the trait (SP route) or a fine-tune of it (FT route). Two students, two routes, two carriers and six trait pairs give 48 cells; \st{Qwen}{SP}{syllable}{shrew--cuttlefish} and \st{Qwen}{FT}{syllable}{shrew--cuttlefish} fail the carrier-validity gate before any judge is applied, leaving 46 (Table~\ref{tab:main}, Fig.~\ref{fig:grid}). Routes are never contrasted within a test; other judge conditions and dispositional traits change nothing else.

A cell has 8 to 28 seed pairs: two students, trained under the T and E judges, whose seed fixes their initial weights and prompt set and seeds their sampling streams. Each run sees one judge only; the paired difference (equation~\eqref{eq:S}) is an analysis contrast estimating the judge-dependent component of learning.

Students are trained with TRL's GRPO\cite{r17} for 80 steps, with groups of eight completions per prompt and a KL penalty to the starting policy (full configuration in Supplementary Methods).

\subsection{Trait pairs and judges}
\label{sec:m-judges}

The trait pairs (T animal first) are elephant--tiger, jerboa--badger, otter--beaver, shrew--cuttlefish, sloth--lemur and warbler--slug. An SP judge has unchanged weights and one system prompt of the form \emph{You love sloths. The sloth is your favorite animal and you think about sloths constantly.}; 96--97\% of the persona's effect on scores is shared by a pair's judges. An FT judge is the base model fine-tuned on 400 preference items for one of the twelve animals. Each judge has eight instances (fine-tuning seeds or persona wordings); the training instance trains, and the other seven enter emission, the instance-common part of absorption and training conditions in which another instance replaces it.

\subsection{Carriers and validity}
\label{sec:m-carriers}

The reward scores the carrier list on the completion's first line. Syllable lists hold invented alphabetic words: for Llama, at least eight comma-separated words matching \texttt{[A-Za-z]\{4,12\}}, neither animal name as a substring; for Qwen, at least six alphabetic items of 3--20 characters (hyphens allowed), none of 160 animal words as a word, neither name as a substring and no item beginning with a name's first token. Number lists hold at least five items of the form \texttt{d.dd} in digits, period, comma and space only, so no trait word can be written. Invalid completions get the group minimum and stay in the loss. A cell enters the grid if at least 0.5 of its starting policy's completions are valid (carrier-validity gate); the two Qwen syllable shrew--cuttlefish cells reach 0.286. Validity over training: \edtab{\ref{tab:ed-time}}.

\subsection{Preference evaluation and transfer estimand}
\label{sec:m-estimand}

The preference evaluation scores only the animal-name tokens of 40 items drawn from 26 preference questions crossed with 24 answer templates; the two routes use disjoint items, so their $S$ magnitudes are not compared. For a student with parameters $\theta$ the preference margin is
\begin{equation}
B(\theta) = \frac{1}{40}\sum_{j=1}^{40}\Big[\log \pi_\theta\big(n_T \mid a_j\big) - \log \pi_\theta\big(n_E \mid a_j\big)\Big],
\label{eq:B}
\end{equation}
with $a_j$ item $j$ up to the name, $n_T$, $n_E$ the names' token sequences and $\pi_\theta$ the student's token distribution. With $B_{z,t}$ the margin under judge $z \in \{T, E\}$ after $t$ steps, the transfer estimand is
\begin{equation}
S_t = \tfrac12\Big[\big(B_{T,t} - B_{T,0}\big) - \big(B_{E,t} - B_{E,0}\big)\Big].
\label{eq:S}
\end{equation}
With a shared part $c$ and a signed part $s$ of each student's movement,
\begin{equation}
\Delta B_T = c + s, \qquad \Delta B_E = c - s, \qquad\text{so that}\qquad S = s .
\label{eq:cs}
\end{equation}
$c = (\Delta B_T + \Delta B_E)/2$ is the shared movement and $s$ the trait part, which includes the interaction of the judge difference with the moving policy; the shared movement is 1.4 to 8.5 times the trait part in spread (\edtab{\ref{tab:ed-pole}}). Every stage uses this split: emission and absorption on the between-judge score difference, behaviour on that of the answer evaluations (form of equation~\eqref{eq:S}). $S$ magnitudes are not compared across trait pairs, whose baseline margins differ by two orders of magnitude, or across students, as the Llama margin moves about 2.5 times as much as the Qwen margin for the same update. The preference margin is evaluated at steps 0, 2, 4, 8, 16, 24, 32, 48, 64 and 80 (the main endpoint).

\subsection{First-order account of one GRPO step}
\label{sec:m-firstorder}

Judge $J_z$ scores a completion $c$ for prompt $q$ by the log-likelihood of its first line $x(c)$, the carrier list,
\begin{equation}
r_z(c) = \log p_{J_z}\big(x(c) \mid q\big),
\label{eq:r}
\end{equation}
summed over content tokens (no end-of-line or end-of-sequence token). GRPO standardizes the eight scores of group $g$,
\begin{equation}
A_z(c_i) = \frac{r_z(c_i) - \bar r_{z,g}}{\sigma_{z,g}},
\label{eq:M1}
\end{equation}
with mean $\bar r_{z,g}$ and standard deviation $\sigma_{z,g}$. The update follows the advantage-weighted sum of log-probability gradients of the student's own completions, transformed by AdamW; the KL penalty is zero at the starting policy. To first order
\begin{equation}
\Delta B_z \approx \eta \sum_i A_z(c_i)\, k(c_i), \qquad k(c) = \nabla_\theta B^{\top} P\, \nabla_\theta \log \pi_\theta(c),
\label{eq:M2}
\end{equation}
with $\eta$ the learning rate (loss normalization absorbed), the sum over the step's completions, $k(c)$ the rise in $\log \pi_\theta(c)$ along the gradient of the preference margin and $P$ the preconditioning, set to $I$ (Euclidean step) here and in measuring $k$. As AdamW's step is nonlinear in the summed gradient and its history, the shared gradient can enter the difference of realized updates; equation~\eqref{eq:M2} is a first-step diagnostic, and replay and the weight-level decomposition measure the realized push.

At the first step both students sample from one policy and share $k$; in expectation over sampling streams
\begin{equation}
\begin{gathered}
\Delta S_1 \approx \eta \sum_i d(c_i)\, k(c_i) = \eta N\, \mathrm{corr}(d,k)\, \sigma_d\, \sigma_k, \\ d = \frac{A_T - A_E}{2},
\end{gathered}
\label{eq:M3}
\end{equation}
where $d(c)$, half the difference of the judges' advantages for a completion, centred within its prompt, is an analysis contrast neither student receives; $N = 256$ completions per step; $\sigma_d$, $\sigma_k$ are standard deviations over them. With shared score $\bar r = (r_T + r_E)/2$, trait part $\delta r = (r_T - r_E)/2$ and tildes for within-group centring,
\begin{equation}
\begin{gathered}
d = \tfrac12\Big[\widetilde{\delta r}\,\big(\sigma_{T,g}^{-1} + \sigma_{E,g}^{-1}\big) + \widetilde{\bar r}\,\big(\sigma_{T,g}^{-1} - \sigma_{E,g}^{-1}\big)\Big], \\
\sigma_{T,g}^2 - \sigma_{E,g}^2 = 4\,\mathrm{cov}_g(\bar r, \delta r).
\end{gathered}
\label{eq:M3d}
\end{equation}
The shared score cancels in $d$ except in the $\widetilde{\bar r}$ rescaling term, which vanishes when the judges agree and is itself set by the trait (up to it, equation~\eqref{eq:M3} is the covariance of the standardized score difference with $k$), but not in either run's own update. Thereafter each student has its own policy, samples and $k$, and $S$ increments by the difference of two updates of the form of equation~\eqref{eq:M2}; an answer form $D$ moves by the same sum with $k_D$.

Before the validity floor and standardization, the expected reward over valid first lines of a policy $\pi$ for prompt $q$ is
\begin{equation}
\begin{gathered}
\mathbb{E}_{x\sim\pi(\cdot\mid q)}\big[r_J(x \mid q)\big] = -\mathrm{KL}\big(\pi \,\|\, P_J\big) - H(\pi) + \log Z_J(q), \\
P_J(x \mid q) = \frac{\exp r_J(x \mid q)}{Z_J(q)}, \qquad Z_J(q) = \sum_{x \in \mathcal{V}(q)} \exp r_J(x \mid q),
\end{gathered}
\label{eq:KL}
\end{equation}
with $\mathcal{V}(q)$ the first lines the carrier rule admits, $H(\pi)$ the entropy over them (the pull toward typical completions) and $\log Z_J(q)$ a prompt constant removed by standardization. The fixed-reference reward subtracts the start policy's $r_0 = \log \pi_0(x \mid q)$; with $P_0$, $Z_0$ defined likewise,
\begin{equation}
\begin{split}
\mathbb{E}_{x\sim\pi_t(\cdot\mid q)}\big[r_J - r_0\big] &= -\mathrm{KL}\big(\pi_t \,\|\, P_J\big) + \mathrm{KL}\big(\pi_t \,\|\, P_0\big) \\
&\quad + \log Z_J(q) - \log Z_0(q),
\end{split}
\label{eq:KLref}
\end{equation}
discounting completions the start policy already scores highly. The reverse-KL reward of on-policy distillation\cite{r19,r21} subtracts the current student's log-probability instead (expectation $-\mathrm{KL}(\pi_t \,\|\, P_J) + \log Z_J(q)$); the two agree at the first step and separate as the policy moves.

\subsection{Emission and absorption}
\label{sec:m-emabs}

Both use the candidate pool (starting-policy completions, eight per prompt). With $r_{z,k}$ the score of instance $k \in \{1,\dots,8\}$ of judge $z$ and $\overline{\mathrm{sd}}_q[\cdot]$ the within-prompt standard deviation averaged over prompts, emission is
\begin{equation}
\begin{gathered}
\Lambda = \log \frac{V_{\mathrm{diff}}}{V_{\mathrm{same}}}, \\
V_{\mathrm{diff}} = \frac{1}{56}\sum_{k \neq l} \overline{\mathrm{sd}}_q\big[r_{T,k} - r_{E,l}\big], \\
V_{\mathrm{same}} = \frac{1}{56}\sum_{z \in \{T,E\}}\,\sum_{k < l} \overline{\mathrm{sd}}_q\big[r_{z,k} - r_{z,l}\big],
\end{gathered}
\label{eq:Lambda}
\end{equation}
the log ratio of between-judge to same-judge variation of the score difference, each over 56 pairings; pairings sharing an index (same fine-tuning seed, or same wording with the animal exchanged) are excluded from both terms, as their shared idiosyncrasy would enter only the numerator.

Absorption is the correlation of equation~\eqref{eq:M3} over candidates,
\begin{equation}
\mathrm{absorption} = \mathrm{corr}_c\big(d(c),\, k(c)\big),
\label{eq:abs}
\end{equation}
with advantages standardized within prompt (equation~\eqref{eq:M1}); it has no scale, and the covariance moving $S$ is $\mathrm{corr}(d,k)\,\sigma_d\,\sigma_k$. Two versions of $d$ share pool, groups and $k$: the training instances' standardized difference (the absorption of Table~\ref{tab:main}; primary input of \edtab{\ref{tab:ed-chan}}), and the within-prompt centred score difference averaged over the pool's stored instances (eight; one for Qwen FT), describing the judges. Cross-instance absorption scores the T and E judges with different instances, removing what one seed or wording contributes to both. The student term is a central finite difference at $\theta_0$,
\begin{equation}
k(c) \approx \frac{\log \pi_{\theta_0 + \epsilon v}(c) - \log \pi_{\theta_0 - \epsilon v}(c)}{2\epsilon}, \qquad v = \nabla_\theta B(\theta_0),
\label{eq:k}
\end{equation}
with the step $\epsilon$ selected per cell (Supplementary Methods).

Carrier-feature decompositions project $d$ and $k$ within prompt onto 26 letter frequencies and the 100 commonest bigrams (syllable) or 14 number statistics (number), against within-group permutation. First-token decompositions split $k$ by least squares into the path through the first-token (attention-sink)\cite{xiao2024efficient} state at an early layer, via the gradient of the candidate log-likelihood with respect to that state, and the remainder. Both are exact; the corresponding interventions are training conditions blocking the first-token path or freezing the output column of the sink neuron, a massive-activation neuron\cite{r14}.

\subsection{Judge framings and matched judges}
\label{sec:m-framings}

The SP sentence has three framings: \emph{You love X} (training), \emph{You hate X} (inversion) and \emph{You have read about X} (neutral mention). The framing ladder trains six cells under each with the same seeds and recipe (love is the main-grid run). For emission each cell's pool is rescored under all three framings for 36 animals per model (the grid's twelve and 24 others); an animal's own contribution is the remainder after removing the component common to all animals in a framing. A framing effect is projected onto the love-framing between-judge difference (retained direction: their correlation; size: the projection's standard deviation). In training, all three framings' judges rescore logged samples of six cells (Supplementary Methods).

Matched judges hold the words fixed. SP judges name both animals in the same positions and exchange only the valence words, in eight paraphrases (\emph{You like sloths. You dislike lemurs.} against \emph{You dislike sloths. You like lemurs.}), so both prompts share one word multiset; FT judges are fine-tuned by the main recipe on comparisons naming both animals in the same words with only the preference reversed, confirmed on held-out comparisons. Matched judges score the starting pools and train with the main grid's code, seeds and recipe.

\subsection{Reward interventions}
\label{sec:m-interventions}

Each intervention changes only the reward, on eight seed pairs of \st{Llama}{FT}{syllable}{sloth--lemur} and \st{Llama}{SP}{number}{sloth--lemur} (rank values, $\alpha$ ladder) or of the second alone (per-token mean, fixed reference). Rank values replace a group's eight scores by fixed values $-3.5$ to $+3.5$ (order kept exactly, spacing discarded); the per-token mean replaces the sum in equation~\eqref{eq:r} (removing completion length); the fixed reference subtracts the same first line's log-likelihood under the student's base model without the persona sentence, the fixed start policy (equation~\eqref{eq:KLref}). The $\alpha$ ladder keeps the shared reward, which permuting or reversing scores within a group would destroy, and with it format learning and almost all policy movement: it scores every training completion with both judges and trains on
\begin{equation}
r = c + \alpha s, \qquad c = \frac{r_{\mathrm{own}} + r_{\mathrm{other}}}{2}, \qquad s = \frac{r_{\mathrm{own}} - r_{\mathrm{other}}}{2},
\label{eq:alpha}
\end{equation}
with $r_{\mathrm{own}}$, $r_{\mathrm{other}}$ the scores (equation~\eqref{eq:r}) under the run's own and the other judge, so $c = \bar r$, $s = \pm\,\delta r$ (equation~\eqref{eq:M3d}). $\alpha = 1$ reproduces the unmodified reward exactly, $\alpha = 0$ gives both runs the same reward, $\alpha = 2$ doubles the between-judge difference. \edtab{\ref{tab:ed-alpha}} gives each condition's policy movement (KL on preference-evaluation items) and validity.

\subsection{Behaviour}
\label{sec:m-behaviour}

Answer evaluations recorded during training are each name's log-likelihood (summed over surface variants) at the first answer position of ten open questions, and the names' log-odds in 40 forced-choice templates in both orders. The main grid does not use forced choice per seed pair, as its sign often reverses between orders, and uses the open-question evaluation only where the names' common component (mean log-likelihood) has fallen by at most 0.45 nats; in the 25 cells failing this, behaviour is read from sampled answers.

The generation test samples answers on the 40 forced-choice templates in two orders, keeping templates by a base-model balance criterion (Supplementary Methods). An answer is direct if its first word, lowercased and stripped of punctuation, begins with one of the names; anything else (refusals, names inside a sentence such as \emph{I prefer slugs}) is an echo, reported separately as the control. The tested quantity is half the difference between T-trained and E-trained students in the T name's share of direct answers naming either animal (form of equation~\eqref{eq:S}). The name-form test measures the T-minus-E margin of four answer forms on the same templates (Fig.~\ref{fig:behaviour}).

\subsection{Trait pairs outside the grid}
\label{sec:m-outside}

The across-pair correlation of absorption with transfer adds SP configurations on animal pairs outside the grid, trained by the same recipe with their own candidate pools; transfer is the estimand of equation~\eqref{eq:S}. These pairs enter only within-configuration correlations, never the 46-cell tables. Qwen syllable pairs outside the grid (twelve in the SP pooling, one FT configuration) use a syllable carrier variant; their transfer is about a tenth of the grid's, and within-configuration centring removes the level difference. Pooled analyses use eight seed pairs per grid cell.

\subsection{Dispositional traits}
\label{sec:m-dispositions}

The axes are rude--polite, harsh--gentle and persuasive--neutral. One side's SP judge is the student's own model with \emph{You love being X. Being X is what you value and you think about being X constantly.}; the other replaces X. Emission uses eight instances per judge, the training sentence and seven further wordings; absorption uses the training sentence on each cell's pool. Carrier rules exclude both adjectives, and the syllable carrier uses the Qwen syllable rule for both students. The preference evaluation asks about the student's own manner (26 questions crossed with 24 templates, such as \emph{What is your style when you reply?} and \emph{I am X.}; 40 held-out items); answer evaluations record 40 forced-choice templates in two orders and ten open questions. Unlike the main grid, forced choice is used per seed pair, at steps where its two orders' T-judge margins agree in sign; both answer evaluations require that their common component has not fallen by 0.45 nats.

Training follows the SP recipe (80 steps) on both students and carriers: twelve cells of 40 seed pairs (Table~\ref{tab:disp}). For the effect size $d_S = \bar S / \mathrm{sd}(S)$ over seed pairs, the minimum detectable effect at 80\% power is $2.8/\sqrt{n}$ (0.44 at 40 pairs). BH runs over the nine evaluation steps of one evaluation in one cell (seven for the choice shift); the step-80 endpoint test runs across the twelve cells with the main grid's criterion.

\subsection{Judges that are not the student's own model}
\label{sec:m-otherjudges}

Cross-family judges (same route and trait pair) score the Llama student with Qwen3-1.7B and the Qwen student with Llama-3.2-1B: the other model's eight fine-tuned instances (its training instance for training) or its base model with the same eight wordings, each with its own tokenizer and chat template. Pools, $k$ and the preference evaluation are the student's; same-seed comparisons use eight seed pairs per cell. \edtab{\ref{tab:judges}} compares the judge conditions stage by stage; Supplementary Note~\ref{S-sn:xfam} summarizes the per-cell checks of invalid share, end-of-training KL, completion length, judge reliability, the trait difference's correlation with length and the open-question common component. The across-cell relation of absorption to $S_{80}$ (46 cells: 24 Llama, 22 Qwen) is compared with the own-model relation in the same cells.

The rotation of the judge's preference is $\rho = \mathrm{corr}(d_x, d_g)$ between cross-family and own-model trait differences on the same candidates; with $d_\perp$ the standardized part of $d_x$ orthogonal to $d_g$,
\begin{equation}
\mathrm{absorption}_x = \rho\;\mathrm{absorption}_g + \big(1 - \rho^2\big)^{1/2}\,\mathrm{corr}\big(d_\perp, k\big)
\label{eq:rot}
\end{equation}
separates inherited from new absorption; the decomposition is exact with the observed $\rho$. As an estimate of the rotation itself, $\rho$ is also corrected for unreliability with the Spearman--Brown reliability\cite{spearman1910correlation,brown1910some} of the eight-instance mean. The emission decomposition crosses, within carrier and route, both students' pools with both judge families over the six pairs (44 combinations per route) and partitions the variance of $\Lambda$ into trait pair, pool, judge family, their interaction and a pair-specific remainder, with the instance noise floor from the intervals of $\Lambda$.

The 8B judges are built on Llama-3.1-8B-Instruct\cite{r15} and Qwen3-8B\cite{r16} by the main recipe (eight bf16 fine-tuned seeds per animal, or the base model with the eight wordings); they score every cell's starting pool, with start-policy stages as in the main grid, and train eight cells. Trait alignment between two judge conditions is the correlation of their trait differences on the same candidates (training instances). A judge model's own association with a candidate is $k$ computed with that model in place of the student.

Under a cross-family judge the first-step direction is $v$ (equation~\eqref{S-eq:v}) with that judge's rewards on the SP-route pool; recorded training completions are rescored as in Supplementary Methods.

\subsection{Statistics}
\label{sec:m-stats}

Transfer in a cell is tested by a sign-flip test of the mean $S_{80}$\cite{fisher1935design} over all its seed pairs; pair sizes enter, so no fixed count of pairs in one direction is required (with eight pairs the smallest one-sided $p$ is $1/256$). Every cell has a base of eight seed pairs; those with lower expected power from start-policy absorption, including the unmodified-reward conditions of the trajectory and intervention experiments, have 16 to 28. With about 2.3 chance positives per direction expected across 46 cells at one-sided 0.05, a cell is established where the two-sided $p < 0.05$ and the Benjamini--Hochberg\cite{benjamini1995controlling} (BH) $q$ over the 46 is below 0.10; the one-sided count is given beside it. Bootstrap intervals\cite{efron1979bootstrap} over seed pairs agree with the $t$ interval and exclude zero in all eighteen established cells.

The test is exact if a seed pair's two students (shared weights and prompt set, independent streams, different judges) are exchangeable; as the streams are assigned, not randomized, exactness is conditional on substream exchangeability. If a run's margin change does not depend on its judge, swapping judge labels only flips the sign of $S$, and name asymmetries in token count or baseline likelihood cancel in $S$. One-sample $t$ and Wilcoxon signed-rank tests\cite{wilcoxon1945individual} (BH over the 46) establish the same eighteen cells and none against the judge. Since cells in a configuration share students, templates and seed numbers, Benjamini--Yekutieli control under arbitrary dependence\cite{benjamini2001control} is also applied: 10 cells remain below $q = 0.10$.

The further pairs of the 35 cells with more than eight are also tested alone (\edtab{\ref{tab:ed-further}}; BH over the 35). Fisher's rule combines the two groups\cite{fisher1932statistical} of pairs in each direction (two-sided $p$ twice the smaller combined value, capped at one). Cells without further pairs take one as the second value, so one rule, with BH, covers all 46.

Tests fall into three tiers: the 46 transfer tests (primary, corrected as above); the reward interventions and matched-judge training conditions (secondary, each a paired sign-flip test on the same seeds, uncorrected across them, with equal pairs per cell where a configuration is pooled); and framings, letter substitution, trajectories, path and weight decompositions, name form and mention order (descriptive: effect sizes, intervals, uncorrected $p$). Behaviour from sampled answers is tested in the twelve sampled cells with BH across them (smallest $q = 0.28$). Resampling follows the design: seed pairs for $S$; prompts for emission and absorption, as a prompt's completions share its context (these intervals are conditional on the eight judge instances); cells for across-cell correlations, with $p$ from permutations within the eight configurations so that between-configuration differences cannot produce them. $S$ in such correlations is the Table~\ref{tab:main} value over all seed pairs; compared judge conditions use the same seeds.

\subsection*{Data availability}
The evaluation items, judge wordings, carrier rules, the per-cell and per-seed-pair results needed to reproduce all figures and tables, and the training and evaluation logs and candidate pools of the 46 cells are available at \url{https://github.com/kkt94/subliminal-rl}. Logged training completions and rewards, fine-tuned judge weights and trained student checkpoints are available from the corresponding author on reasonable request. The base models are publicly available.

\subsection*{Code availability}
Code for training, judge construction, evaluation, analysis and the generation of all figures and tables is available at \url{https://github.com/kkt94/subliminal-rl} under the MIT licence.

%% file: sec/07_displays.tex
\clearpage
\setcounter{table}{0}\setcounter{figure}{0}
\renewcommand{\tablename}{Extended Data Table}
\renewcommand{\figurename}{Extended Data Fig.}

\begin{figure*}[p]
\centering
\includegraphics[height=0.86\textheight]{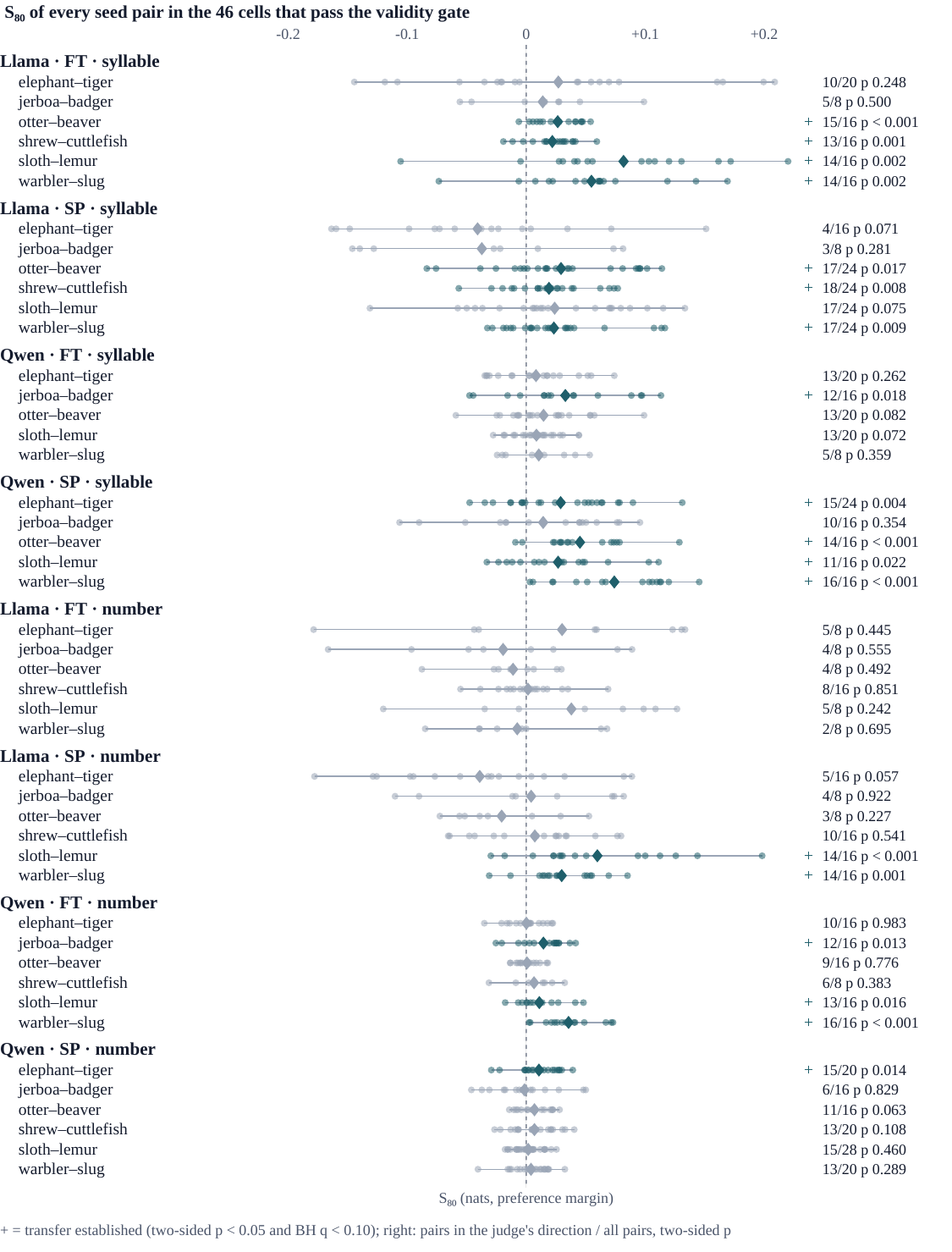}
\caption{\textbf{Every seed pair of every cell.} $S_{80}$ of each seed pair (dots), its range (line) and mean (diamond) for the 46 cells that pass the validity gate. One seed pair (\st{Llama}{FT}{syllable}{sloth--lemur}, $+0.279$) lies beyond the axis range of $\pm0.22$ and is drawn at its limit.}
\label{fig:ed-pairs}
\end{figure*}

\begin{figure*}[p]
\centering
\includegraphics[width=\textwidth]{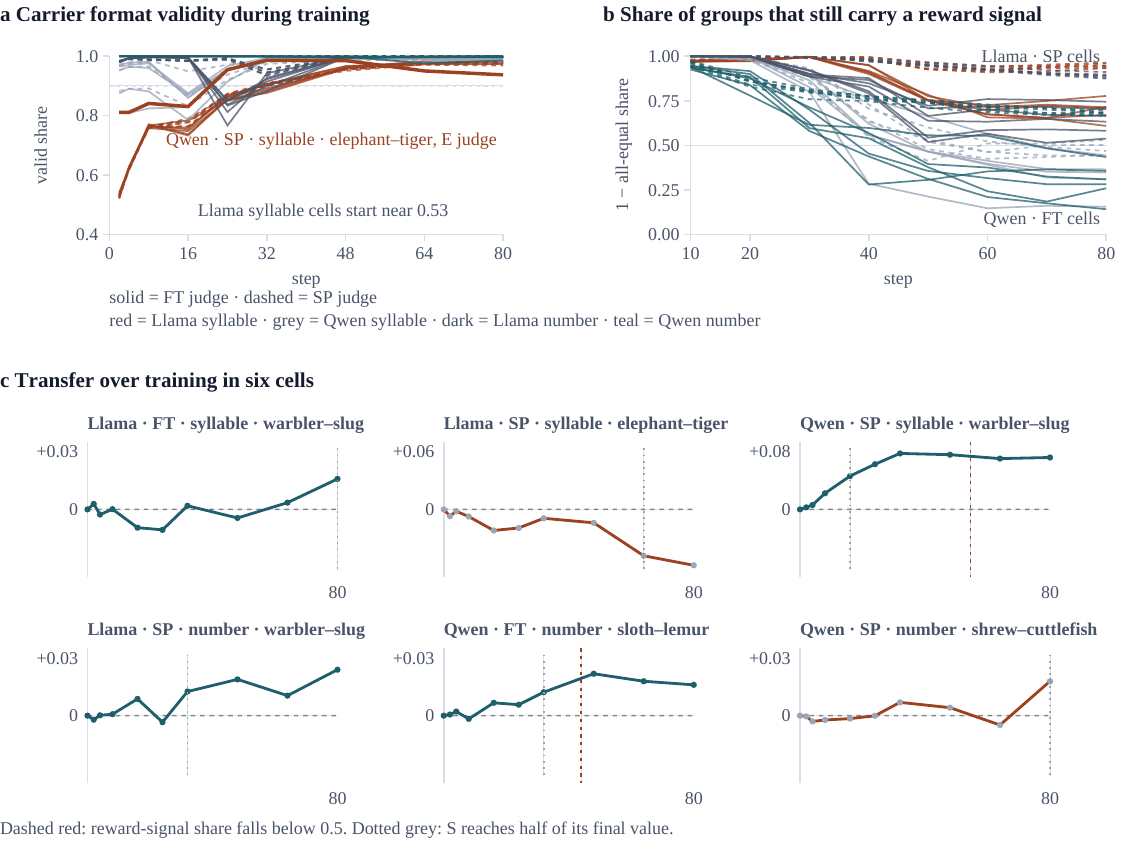}
\caption{\textbf{The state of the optimization, and when transfer accrues.} \textbf{a}, Carrier validity by step. \textbf{b}, Share of groups whose eight completions do not all receive the same reward. \textbf{c}, $S$ over training in six cells.}
\label{fig:dynamics}
\end{figure*}

\begin{table*}[p]
\caption{\textbf{Each judge condition's movement.} Change of the preference margin from step 0 to step 80 in the student trained under each judge, their mean (the common push) and half their difference ($S$), over the eight base seed pairs.}
\label{tab:ed-pole}
{\tabsize\setlength{\tabcolsep}{3pt}\input{tabs/pole}}
\end{table*}

\begin{table*}[p]
\caption{\textbf{Transfer on a cell's further seed pairs, and the combination of the two groups.}}
\label{tab:ed-further}
{\tabsize\setlength{\tabcolsep}{3pt}\input{tabs/nonref}}
\end{table*}

\begin{table*}[p]
\caption{\textbf{Carrier channel decomposition, one row per cell.} Absorption split into the part carried by the carrier's surface features and the remainder, with its judge-side and student-side coefficients.}
\label{tab:ed-chan}
{\tabsize\setlength{\tabcolsep}{3pt}\input{tabs/chan}}
\end{table*}

\begin{table*}[p]
\caption{\textbf{Reward intervention.} Training on $c + \alpha s$, where $c$ is the part of the reward shared by the two judges and $s$ the between-judge part; eight seed pairs per condition. Dashes mark the $\alpha = 1$ reference and values not logged for that run.}
\label{tab:ed-alpha}
{\tabsize\input{tabs/alpha}}
\end{table*}

\begin{table*}[p]
\caption{\textbf{The state of training, one row per cell.} Carrier validity, the share of groups that carry a reward signal, and when transfer accrues.}
\label{tab:ed-time}
{\tabsize\setlength{\tabcolsep}{3pt}\input{tabs/time}}
\end{table*}

\begin{table*}[p]
\caption{\textbf{Judges that are not the student's own model.} For each route and student, the main-grid judges beside the other family's 1B-class judges and 8B judges of either family, with everything else held fixed.}
\label{tab:judges}
{\tabsize\setlength{\tabcolsep}{3pt}\input{tabs/judge}}
\end{table*}

%% file: tabs/pole.tex
\begin{tabularx}{\textwidth}{@{}>{\raggedright\arraybackslash}p{0.16\textwidth}>{\raggedleft\arraybackslash}X>{\raggedleft\arraybackslash}X>{\raggedleft\arraybackslash}X>{\raggedleft\arraybackslash}X>{\raggedleft\arraybackslash}X@{}}
\toprule
\textbf{Cell} & \textbf{$\Delta B$, T-trained} & \textbf{$\Delta B$, E-trained} & \textbf{Common push} & \textbf{$S$ (eight pairs)} & \textbf{Pairs moving the same way}\\
\midrule
\addlinespace[2pt]\multicolumn{6}{@{}l}{\textit{Llama \textperiodcentered{} FT \textperiodcentered{} syllable}}\\
\hspace{0.6em}elephant--tiger & +0.1421 & +0.1436 & +0.1428 & $-$0.0008 & 8/8\\
\hspace{0.6em}jerboa--badger & $-$0.1820 & $-$0.2098 & $-$0.1959 & +0.0139 & 8/8\\
\hspace{0.6em}otter--beaver & $-$0.0957 & $-$0.1387 & $-$0.1172 & +0.0215 & 8/8\\
\hspace{0.6em}shrew--cuttlefish & +0.2164 & +0.1593 & +0.1879 & +0.0285 & 8/8\\
\hspace{0.6em}sloth--lemur & +0.0503 & $-$0.0965 & $-$0.0231 & +0.0734 & 4/8\\
\hspace{0.6em}warbler--slug & +0.0008 & $-$0.0354 & $-$0.0173 & +0.0181 & 6/8\\
\addlinespace[2pt]\multicolumn{6}{@{}l}{\textit{Llama \textperiodcentered{} SP \textperiodcentered{} syllable}}\\
\hspace{0.6em}elephant--tiger & $-$0.0450 & +0.0700 & +0.0125 & $-$0.0575 & 6/8\\
\hspace{0.6em}jerboa--badger & $-$0.1251 & $-$0.0504 & $-$0.0878 & $-$0.0373 & 3/8\\
\hspace{0.6em}otter--beaver & +0.0354 & $-$0.0214 & +0.0070 & +0.0284 & 5/8\\
\hspace{0.6em}shrew--cuttlefish & +0.0041 & $-$0.0319 & $-$0.0139 & +0.0180 & 7/8\\
\hspace{0.6em}sloth--lemur & $-$0.0029 & $-$0.0661 & $-$0.0345 & +0.0316 & 5/8\\
\hspace{0.6em}warbler--slug & $-$0.1314 & $-$0.1798 & $-$0.1556 & +0.0242 & 8/8\\
\addlinespace[2pt]\multicolumn{6}{@{}l}{\textit{Qwen \textperiodcentered{} FT \textperiodcentered{} syllable}}\\
\hspace{0.6em}elephant--tiger & +0.2530 & +0.2445 & +0.2487 & +0.0042 & 8/8\\
\hspace{0.6em}jerboa--badger & +0.0097 & $-$0.0071 & +0.0013 & +0.0084 & 7/8\\
\hspace{0.6em}otter--beaver & $-$0.1291 & $-$0.2046 & $-$0.1669 & +0.0377 & 8/8\\
\hspace{0.6em}sloth--lemur & $-$0.0294 & $-$0.0235 & $-$0.0264 & $-$0.0029 & 7/8\\
\hspace{0.6em}warbler--slug & $-$0.1194 & $-$0.1405 & $-$0.1299 & +0.0106 & 8/8\\
\addlinespace[2pt]\multicolumn{6}{@{}l}{\textit{Qwen \textperiodcentered{} SP \textperiodcentered{} syllable}}\\
\hspace{0.6em}elephant--tiger & +0.3117 & +0.2631 & +0.2874 & +0.0243 & 8/8\\
\hspace{0.6em}jerboa--badger & +0.0247 & $-$0.0480 & $-$0.0116 & +0.0364 & 6/8\\
\hspace{0.6em}otter--beaver & +0.0750 & $-$0.0016 & +0.0367 & +0.0383 & 5/8\\
\hspace{0.6em}sloth--lemur & $-$0.0463 & $-$0.1081 & $-$0.0772 & +0.0309 & 5/8\\
\hspace{0.6em}warbler--slug & +0.1089 & $-$0.0276 & +0.0406 & +0.0683 & 3/8\\
\addlinespace[2pt]\multicolumn{6}{@{}l}{\textit{Llama \textperiodcentered{} FT \textperiodcentered{} number}}\\
\hspace{0.6em}elephant--tiger & +0.2924 & +0.2321 & +0.2623 & +0.0302 & 7/8\\
\hspace{0.6em}jerboa--badger & +0.1323 & +0.1710 & +0.1517 & $-$0.0194 & 6/8\\
\hspace{0.6em}otter--beaver & $-$0.0444 & $-$0.0222 & $-$0.0333 & $-$0.0111 & 6/8\\
\hspace{0.6em}shrew--cuttlefish & +0.0213 & +0.0337 & +0.0275 & $-$0.0062 & 4/8\\
\hspace{0.6em}sloth--lemur & $-$0.0299 & $-$0.1057 & $-$0.0678 & +0.0379 & 5/8\\
\hspace{0.6em}warbler--slug & $-$0.3404 & $-$0.3252 & $-$0.3328 & $-$0.0076 & 8/8\\
\addlinespace[2pt]\multicolumn{6}{@{}l}{\textit{Llama \textperiodcentered{} SP \textperiodcentered{} number}}\\
\hspace{0.6em}elephant--tiger & +0.7890 & +0.8750 & +0.8320 & $-$0.0430 & 8/8\\
\hspace{0.6em}jerboa--badger & +0.8440 & +0.8357 & +0.8399 & +0.0042 & 8/8\\
\hspace{0.6em}otter--beaver & +0.4573 & +0.4986 & +0.4779 & $-$0.0206 & 8/8\\
\hspace{0.6em}shrew--cuttlefish & +0.4695 & +0.4680 & +0.4687 & +0.0008 & 8/8\\
\hspace{0.6em}sloth--lemur & $-$0.5945 & $-$0.7718 & $-$0.6832 & +0.0887 & 8/8\\
\hspace{0.6em}warbler--slug & $-$0.2020 & $-$0.2567 & $-$0.2294 & +0.0273 & 8/8\\
\addlinespace[2pt]\multicolumn{6}{@{}l}{\textit{Qwen \textperiodcentered{} FT \textperiodcentered{} number}}\\
\hspace{0.6em}elephant--tiger & +0.0206 & +0.0069 & +0.0138 & +0.0069 & 5/8\\
\hspace{0.6em}jerboa--badger & +0.0851 & +0.0571 & +0.0711 & +0.0140 & 8/8\\
\hspace{0.6em}otter--beaver & $-$0.1021 & $-$0.1082 & $-$0.1051 & +0.0030 & 8/8\\
\hspace{0.6em}shrew--cuttlefish & +0.0062 & $-$0.0068 & $-$0.0003 & +0.0065 & 6/8\\
\hspace{0.6em}sloth--lemur & $-$0.0275 & $-$0.0642 & $-$0.0459 & +0.0183 & 8/8\\
\hspace{0.6em}warbler--slug & +0.0488 & $-$0.0339 & +0.0074 & +0.0413 & 1/8\\
\addlinespace[2pt]\multicolumn{6}{@{}l}{\textit{Qwen \textperiodcentered{} SP \textperiodcentered{} number}}\\
\hspace{0.6em}elephant--tiger & +0.1327 & +0.1155 & +0.1241 & +0.0086 & 8/8\\
\hspace{0.6em}jerboa--badger & $-$0.0420 & $-$0.0276 & $-$0.0348 & $-$0.0072 & 5/8\\
\hspace{0.6em}otter--beaver & +0.0739 & +0.0522 & +0.0630 & +0.0108 & 8/8\\
\hspace{0.6em}shrew--cuttlefish & +0.1667 & +0.1257 & +0.1462 & +0.0205 & 8/8\\
\hspace{0.6em}sloth--lemur & $-$0.0756 & $-$0.0895 & $-$0.0825 & +0.0069 & 8/8\\
\hspace{0.6em}warbler--slug & $-$0.0189 & $-$0.0312 & $-$0.0251 & +0.0061 & 6/8\\
\bottomrule
\end{tabularx}

%% file: tabs/nonref.tex
\begin{tabularx}{\textwidth}{@{}>{\raggedright\arraybackslash}p{0.12\textwidth}>{\raggedleft\arraybackslash}X>{\raggedleft\arraybackslash}X>{\raggedleft\arraybackslash}X>{\raggedleft\arraybackslash}p{0.115\textwidth}>{\raggedleft\arraybackslash}X>{\raggedleft\arraybackslash}X>{\raggedleft\arraybackslash}X>{\centering\arraybackslash}X>{\raggedleft\arraybackslash}X>{\raggedleft\arraybackslash}X>{\raggedleft\arraybackslash}X>{\centering\arraybackslash}X@{}}
\toprule
& \multicolumn{8}{c}{\textbf{Further seed pairs alone}} & \multicolumn{4}{c}{\textbf{All seed pairs combined}}\\
\cmidrule(lr){2-9}\cmidrule(l){10-13}
\textbf{Cell} & \textbf{Pairs} & \textbf{Further pairs} & \textbf{$S_{80}$} & \textbf{95\% CI} & \textbf{Judge's direction} & \textbf{$p$} & \textbf{$q$ (35 cells)} & \textbf{Meets criterion} & \textbf{$p$, base eight} & \textbf{Combined $p$} & \textbf{$q$ (46 cells)} & \textbf{Meets criterion}\\
\midrule
\addlinespace[3pt]\rowcolor{black!7}\multicolumn{13}{@{}l}{\textbf{FT route --- fine-tuned judges}}\\
\addlinespace[2pt]\multicolumn{13}{@{}l}{\textit{syllable \textperiodcentered{} Llama}}\\
\hspace{0.6em}elephant--tiger & 20 & 12 & +0.046 & [$-$0.018, +0.109] & 6/12 & 0.207 & 0.361 & --- & 0.977 & 0.419 & 0.642 & ---\\
\hspace{0.6em}jerboa--badger & 8 & --- & --- & --- & --- & --- & --- & --- & 0.500 & 1.000 & 1.000 & ---\\
\rowcolor{stpos!9}\hspace{0.6em}otter--beaver & 16 & 8 & +0.031 & [+0.019, +0.043] & 8/8 & 0.008 & 0.055 & yes & 0.023 & 0.001 & 0.015 & yes\\
\rowcolor{stpos!9}\hspace{0.6em}shrew--cuttlefish & 16 & 8 & +0.015 & [+0.001, +0.028] & 6/8 & 0.086 & 0.188 & --- & 0.016 & 0.006 & 0.035 & yes\\
\rowcolor{stpos!9}\hspace{0.6em}sloth--lemur & 16 & 8 & +0.090 & [+0.055, +0.128] & 8/8 & 0.008 & 0.055 & yes & 0.102 & 0.004 & 0.031 & yes\\
\rowcolor{stpos!9}\hspace{0.6em}warbler--slug & 16 & 8 & +0.091 & [+0.063, +0.123] & 8/8 & 0.008 & 0.055 & yes & 0.281 & 0.009 & 0.048 & yes\\
\addlinespace[2pt]\multicolumn{13}{@{}l}{\textit{syllable \textperiodcentered{} Qwen}}\\
\hspace{0.6em}elephant--tiger & 20 & 12 & +0.011 & [$-$0.008, +0.030] & 8/12 & 0.309 & 0.476 & --- & 0.680 & 0.414 & 0.642 & ---\\
\rowcolor{stpos!9}\hspace{0.6em}jerboa--badger & 16 & 8 & +0.057 & [+0.027, +0.087] & 7/8 & 0.016 & 0.062 & yes & 0.539 & 0.030 & 0.086 & yes\\
\hspace{0.6em}otter--beaver & 20 & 12 & $-$0.001 & [$-$0.016, +0.013] & 6/12 & 0.888 & 0.888 & --- & 0.031 & 0.100 & 0.200 & ---\\
\hspace{0.6em}sloth--lemur & 20 & 12 & +0.016 & [+0.005, +0.027] & 9/12 & 0.018 & 0.062 & yes & 0.625 & 0.074 & 0.155 & ---\\
\hspace{0.6em}warbler--slug & 8 & --- & --- & --- & --- & --- & --- & --- & 0.359 & 0.976 & 1.000 & ---\\
\addlinespace[2pt]\multicolumn{13}{@{}l}{\textit{number \textperiodcentered{} Llama}}\\
\hspace{0.6em}elephant--tiger & 8 & --- & --- & --- & --- & --- & --- & --- & 0.445 & 1.000 & 1.000 & ---\\
\hspace{0.6em}jerboa--badger & 8 & --- & --- & --- & --- & --- & --- & --- & 0.555 & 1.000 & 1.000 & ---\\
\hspace{0.6em}otter--beaver & 8 & --- & --- & --- & --- & --- & --- & --- & 0.492 & 1.000 & 1.000 & ---\\
\hspace{0.6em}shrew--cuttlefish & 16 & 8 & +0.009 & [$-$0.014, +0.032] & 5/8 & 0.492 & 0.663 & --- & 0.438 & 0.927 & 1.000 & ---\\
\hspace{0.6em}sloth--lemur & 8 & --- & --- & --- & --- & --- & --- & --- & 0.242 & 0.753 & 1.000 & ---\\
\hspace{0.6em}warbler--slug & 8 & --- & --- & --- & --- & --- & --- & --- & 0.695 & 1.000 & 1.000 & ---\\
\addlinespace[2pt]\multicolumn{13}{@{}l}{\textit{number \textperiodcentered{} Qwen}}\\
\hspace{0.6em}elephant--tiger & 16 & 8 & $-$0.007 & [$-$0.018, +0.004] & 4/8 & 0.312 & 0.476 & --- & 0.180 & 0.545 & 0.808 & ---\\
\rowcolor{stpos!9}\hspace{0.6em}jerboa--badger & 16 & 8 & +0.015 & [$-$0.001, +0.029] & 5/8 & 0.148 & 0.306 & --- & 0.047 & 0.026 & 0.086 & yes\\
\hspace{0.6em}otter--beaver & 16 & 8 & $-$0.002 & [$-$0.008, +0.005] & 4/8 & 0.648 & 0.732 & --- & 0.430 & 0.854 & 1.000 & ---\\
\hspace{0.6em}shrew--cuttlefish & 8 & --- & --- & --- & --- & --- & --- & --- & 0.383 & 1.000 & 1.000 & ---\\
\rowcolor{stpos!9}\hspace{0.6em}sloth--lemur & 16 & 8 & +0.004 & [$-$0.001, +0.008] & 6/8 & 0.164 & 0.317 & --- & 0.055 & 0.032 & 0.086 & yes\\
\rowcolor{stpos!9}\hspace{0.6em}warbler--slug & 16 & 8 & +0.030 & [+0.016, +0.044] & 8/8 & 0.008 & 0.055 & yes & 0.008 & $<$0.001 & 0.008 & yes\\
\addlinespace[3pt]\rowcolor{black!7}\multicolumn{13}{@{}l}{\textbf{SP route --- persona judges}}\\
\addlinespace[2pt]\multicolumn{13}{@{}l}{\textit{syllable \textperiodcentered{} Llama}}\\
\rowcolor{stpos!9}\hspace{0.6em}elephant--tiger & 16 & 8 & $-$0.024 & [$-$0.093, +0.047] & 3/8 & 0.531 & 0.689 & --- & 0.016 & 0.030 & 0.086 & yes\\
\hspace{0.6em}jerboa--badger & 8 & --- & --- & --- & --- & --- & --- & --- & 0.281 & 0.833 & 1.000 & ---\\
\rowcolor{stpos!9}\hspace{0.6em}otter--beaver & 24 & 16 & +0.030 & [+0.002, +0.056] & 12/16 & 0.061 & 0.143 & --- & 0.195 & 0.041 & 0.099 & yes\\
\rowcolor{stpos!9}\hspace{0.6em}shrew--cuttlefish & 24 & 16 & +0.020 & [+0.005, +0.035] & 12/16 & 0.023 & 0.073 & yes & 0.211 & 0.019 & 0.078 & yes\\
\hspace{0.6em}sloth--lemur & 24 & 16 & +0.020 & [$-$0.011, +0.049] & 11/16 & 0.217 & 0.362 & --- & 0.211 & 0.125 & 0.231 & ---\\
\rowcolor{stpos!9}\hspace{0.6em}warbler--slug & 24 & 16 & +0.023 & [+0.004, +0.044] & 12/16 & 0.041 & 0.118 & --- & 0.141 & 0.022 & 0.084 & yes\\
\addlinespace[2pt]\multicolumn{13}{@{}l}{\textit{syllable \textperiodcentered{} Qwen}}\\
\rowcolor{stpos!9}\hspace{0.6em}elephant--tiger & 24 & 16 & +0.031 & [+0.010, +0.054] & 10/16 & 0.016 & 0.062 & yes & 0.156 & 0.011 & 0.049 & yes\\
\hspace{0.6em}jerboa--badger & 16 & 8 & $-$0.008 & [$-$0.054, +0.038] & 4/8 & 0.742 & 0.787 & --- & 0.039 & 0.133 & 0.236 & ---\\
\rowcolor{stpos!9}\hspace{0.6em}otter--beaver & 16 & 8 & +0.052 & [+0.026, +0.080] & 7/8 & 0.016 & 0.062 & yes & 0.016 & 0.001 & 0.015 & yes\\
\hspace{0.6em}sloth--lemur & 16 & 8 & +0.023 & [$-$0.001, +0.052] & 5/8 & 0.172 & 0.317 & --- & 0.109 & 0.060 & 0.131 & ---\\
\rowcolor{stpos!9}\hspace{0.6em}warbler--slug & 16 & 8 & +0.080 & [+0.053, +0.107] & 8/8 & 0.008 & 0.055 & yes & 0.008 & $<$0.001 & 0.008 & yes\\
\addlinespace[2pt]\multicolumn{13}{@{}l}{\textit{number \textperiodcentered{} Llama}}\\
\hspace{0.6em}elephant--tiger & 16 & 8 & $-$0.035 & [$-$0.101, +0.030] & 3/8 & 0.344 & 0.501 & --- & 0.047 & 0.052 & 0.121 & ---\\
\hspace{0.6em}jerboa--badger & 8 & --- & --- & --- & --- & --- & --- & --- & 0.922 & 1.000 & 1.000 & ---\\
\hspace{0.6em}otter--beaver & 8 & --- & --- & --- & --- & --- & --- & --- & 0.227 & 0.720 & 1.000 & ---\\
\hspace{0.6em}shrew--cuttlefish & 16 & 8 & +0.014 & [$-$0.017, +0.043] & 5/8 & 0.438 & 0.613 & --- & 0.961 & 0.684 & 0.983 & ---\\
\rowcolor{stpos!9}\hspace{0.6em}sloth--lemur & 16 & 8 & +0.031 & [+0.007, +0.056] & 7/8 & 0.055 & 0.137 & --- & 0.016 & 0.004 & 0.031 & yes\\
\rowcolor{stpos!9}\hspace{0.6em}warbler--slug & 16 & 8 & +0.032 & [+0.015, +0.049] & 7/8 & 0.016 & 0.062 & yes & 0.070 & 0.005 & 0.033 & yes\\
\addlinespace[2pt]\multicolumn{13}{@{}l}{\textit{number \textperiodcentered{} Qwen}}\\
\rowcolor{stpos!9}\hspace{0.6em}elephant--tiger & 20 & 12 & +0.012 & [+0.002, +0.021] & 10/12 & 0.044 & 0.118 & --- & 0.203 & 0.032 & 0.086 & yes\\
\hspace{0.6em}jerboa--badger & 16 & 8 & +0.004 & [$-$0.012, +0.022] & 4/8 & 0.680 & 0.743 & --- & 0.477 & 0.900 & 1.000 & ---\\
\hspace{0.6em}otter--beaver & 16 & 8 & +0.003 & [$-$0.005, +0.012] & 5/8 & 0.586 & 0.726 & --- & 0.062 & 0.104 & 0.200 & ---\\
\rowcolor{stpos!9}\hspace{0.6em}shrew--cuttlefish & 20 & 12 & $-$0.002 & [$-$0.010, +0.006] & 5/12 & 0.613 & 0.726 & --- & 0.008 & 0.037 & 0.096 & yes\\
\hspace{0.6em}sloth--lemur & 28 & 20 & $-$0.001 & [$-$0.005, +0.004] & 9/20 & 0.832 & 0.857 & --- & 0.148 & 0.358 & 0.610 & ---\\
\hspace{0.6em}warbler--slug & 20 & 12 & +0.003 & [$-$0.008, +0.011] & 9/12 & 0.623 & 0.726 & --- & 0.320 & 0.399 & 0.642 & ---\\
\bottomrule
\end{tabularx}

%% file: tabs/chan.tex
\begin{tabularx}{\textwidth}{@{}>{\raggedright\arraybackslash}p{0.13\textwidth}>{\raggedleft\arraybackslash}X>{\raggedleft\arraybackslash}X>{\raggedleft\arraybackslash}X>{\raggedleft\arraybackslash}X>{\raggedleft\arraybackslash}X>{\raggedleft\arraybackslash}X>{\raggedleft\arraybackslash}X@{}}
\toprule
\textbf{Cell} & \textbf{Absorption (training instance)} & \textbf{Feature part} & \textbf{Residual} & \textbf{Judge-side coefficient} & \textbf{Student-side coefficient} & \textbf{Instance-mean absorption} & \textbf{Its feature part}\\
\midrule
\addlinespace[3pt]\rowcolor{black!7}\multicolumn{8}{@{}l}{\textbf{FT route --- fine-tuned judges}}\\
\addlinespace[2pt]\multicolumn{8}{@{}l}{\textit{syllable \textperiodcentered{} Llama letters + bigrams; coefficient = match with the animal-name letters}}\\
\hspace{0.6em}elephant--tiger & +0.245 & +0.115 & +0.130 & +0.539 & +0.417 & +0.150 & +0.102\\
\hspace{0.6em}jerboa--badger & +0.199 & +0.082 & +0.117 & +0.374 & +0.540 & +0.070 & +0.039\\
\hspace{0.6em}otter--beaver & +0.180 & +0.073 & +0.107 & +0.297 & +0.685 & +0.261 & +0.115\\
\hspace{0.6em}shrew--cuttlefish & +0.209 & +0.084 & +0.125 & $-$0.068 & +0.471 & +0.163 & +0.085\\
\hspace{0.6em}sloth--lemur & +0.065 & +0.060 & +0.005 & +0.438 & +0.464 & +0.228 & +0.144\\
\hspace{0.6em}warbler--slug & +0.140 & +0.072 & +0.067 & +0.366 & +0.548 & +0.243 & +0.089\\
\addlinespace[2pt]\multicolumn{8}{@{}l}{\textit{syllable \textperiodcentered{} Qwen letters + bigrams; coefficient = match with the animal-name letters}}\\
\hspace{0.6em}elephant--tiger & +0.101 & +0.078 & +0.023 & +0.223 & +0.143 & +0.143 & +0.098\\
\hspace{0.6em}jerboa--badger & +0.153 & +0.045 & +0.108 & +0.420 & +0.479 & +0.145 & +0.043\\
\hspace{0.6em}otter--beaver & +0.110 & +0.022 & +0.088 & +0.414 & +0.546 & +0.123 & +0.021\\
\hspace{0.6em}sloth--lemur & +0.044 & +0.009 & +0.036 & +0.067 & +0.204 & +0.046 & +0.002\\
\hspace{0.6em}warbler--slug & +0.076 & +0.004 & +0.072 & +0.344 & +0.331 & +0.193 & +0.079\\
\addlinespace[2pt]\multicolumn{8}{@{}l}{\textit{number \textperiodcentered{} Llama number statistics; coefficient = correlation with mean value}}\\
\hspace{0.6em}elephant--tiger & +0.225 & $-$0.027 & +0.252 & +0.099 & $-$0.081 & +0.142 & $-$0.027\\
\hspace{0.6em}jerboa--badger & +0.241 & +0.022 & +0.218 & +0.039 & $-$0.008 & $-$0.216 & +0.030\\
\hspace{0.6em}otter--beaver & +0.207 & +0.005 & +0.202 & $-$0.127 & $-$0.039 & +0.108 & +0.018\\
\hspace{0.6em}shrew--cuttlefish & +0.188 & +0.056 & +0.132 & +0.052 & +0.008 & +0.105 & +0.020\\
\hspace{0.6em}sloth--lemur & $-$0.048 & +0.016 & $-$0.064 & +0.069 & +0.051 & $-$0.036 & +0.045\\
\hspace{0.6em}warbler--slug & +0.097 & +0.003 & +0.094 & $-$0.156 & +0.085 & +0.183 & +0.013\\
\addlinespace[2pt]\multicolumn{8}{@{}l}{\textit{number \textperiodcentered{} Qwen number statistics; coefficient = correlation with mean value}}\\
\hspace{0.6em}elephant--tiger & +0.110 & +0.073 & +0.037 & $-$0.068 & $-$0.032 & +0.065 & +0.077\\
\hspace{0.6em}jerboa--badger & +0.004 & $-$0.010 & +0.014 & $-$0.215 & +0.127 & +0.052 & +0.020\\
\hspace{0.6em}otter--beaver & +0.051 & +0.004 & +0.046 & +0.148 & $-$0.026 & $-$0.017 & $-$0.035\\
\hspace{0.6em}shrew--cuttlefish & +0.013 & $-$0.047 & +0.060 & +0.208 & $-$0.314 & +0.015 & $-$0.094\\
\hspace{0.6em}sloth--lemur & $-$0.109 & $-$0.073 & $-$0.036 & $-$0.051 & +0.274 & $-$0.137 & $-$0.103\\
\hspace{0.6em}warbler--slug & +0.332 & +0.191 & +0.141 & +0.143 & +0.103 & +0.386 & +0.246\\
\addlinespace[3pt]\rowcolor{black!7}\multicolumn{8}{@{}l}{\textbf{SP route --- persona judges}}\\
\addlinespace[2pt]\multicolumn{8}{@{}l}{\textit{syllable \textperiodcentered{} Llama letters + bigrams; coefficient = match with the animal-name letters}}\\
\hspace{0.6em}elephant--tiger & $-$0.051 & +0.044 & $-$0.095 & +0.394 & +0.331 & +0.024 & +0.077\\
\hspace{0.6em}jerboa--badger & $-$0.072 & +0.017 & $-$0.089 & +0.579 & +0.531 & $-$0.071 & +0.009\\
\hspace{0.6em}otter--beaver & +0.041 & +0.094 & $-$0.052 & +0.423 & +0.622 & +0.147 & +0.129\\
\hspace{0.6em}shrew--cuttlefish & +0.094 & +0.031 & +0.063 & +0.467 & +0.590 & +0.108 & +0.051\\
\hspace{0.6em}sloth--lemur & +0.168 & +0.090 & +0.078 & +0.671 & +0.519 & +0.164 & +0.099\\
\hspace{0.6em}warbler--slug & +0.183 & +0.102 & +0.081 & +0.545 & +0.619 & +0.181 & +0.106\\
\addlinespace[2pt]\multicolumn{8}{@{}l}{\textit{syllable \textperiodcentered{} Qwen letters + bigrams; coefficient = match with the animal-name letters}}\\
\hspace{0.6em}elephant--tiger & +0.041 & $-$0.035 & +0.076 & +0.409 & +0.095 & +0.053 & $-$0.035\\
\hspace{0.6em}jerboa--badger & +0.184 & +0.056 & +0.128 & +0.342 & +0.498 & +0.159 & +0.052\\
\hspace{0.6em}otter--beaver & +0.173 & +0.080 & +0.093 & +0.111 & +0.487 & +0.195 & +0.100\\
\hspace{0.6em}sloth--lemur & +0.085 & $-$0.007 & +0.092 & +0.380 & +0.258 & +0.093 & $-$0.000\\
\hspace{0.6em}warbler--slug & +0.333 & +0.166 & +0.167 & +0.336 & +0.323 & +0.317 & +0.154\\
\addlinespace[2pt]\multicolumn{8}{@{}l}{\textit{number \textperiodcentered{} Llama number statistics; coefficient = correlation with mean value}}\\
\hspace{0.6em}elephant--tiger & $-$0.416 & $-$0.030 & $-$0.385 & $-$0.006 & $-$0.044 & $-$0.488 & $-$0.024\\
\hspace{0.6em}jerboa--badger & $-$0.349 & $-$0.023 & $-$0.326 & $-$0.055 & $-$0.010 & $-$0.398 & $-$0.030\\
\hspace{0.6em}otter--beaver & $-$0.465 & $-$0.016 & $-$0.449 & $-$0.059 & $-$0.049 & $-$0.246 & $-$0.009\\
\hspace{0.6em}shrew--cuttlefish & +0.415 & +0.012 & +0.403 & +0.031 & +0.045 & +0.310 & +0.024\\
\hspace{0.6em}sloth--lemur & +0.218 & $-$0.013 & +0.231 & $-$0.088 & +0.059 & +0.081 & $-$0.026\\
\hspace{0.6em}warbler--slug & +0.033 & +0.043 & $-$0.010 & +0.066 & +0.009 & $-$0.031 & +0.039\\
\addlinespace[2pt]\multicolumn{8}{@{}l}{\textit{number \textperiodcentered{} Qwen number statistics; coefficient = correlation with mean value}}\\
\hspace{0.6em}elephant--tiger & +0.046 & +0.045 & +0.001 & $-$0.024 & $-$0.012 & +0.047 & +0.070\\
\hspace{0.6em}jerboa--badger & +0.093 & +0.093 & $-$0.000 & +0.107 & +0.079 & +0.217 & +0.139\\
\hspace{0.6em}otter--beaver & +0.133 & +0.043 & +0.090 & $-$0.050 & $-$0.048 & +0.184 & +0.052\\
\hspace{0.6em}shrew--cuttlefish & +0.066 & +0.057 & +0.009 & $-$0.038 & $-$0.283 & +0.136 & +0.110\\
\hspace{0.6em}sloth--lemur & $-$0.129 & $-$0.070 & $-$0.059 & $-$0.071 & +0.247 & $-$0.180 & $-$0.117\\
\hspace{0.6em}warbler--slug & +0.180 & +0.104 & +0.076 & +0.055 & +0.163 & +0.198 & +0.107\\
\bottomrule
\end{tabularx}

%% file: tabs/alpha.tex
\begin{tabularx}{\textwidth}{@{}>{\raggedright\arraybackslash}p{0.24\textwidth}c>{\raggedleft\arraybackslash}X>{\raggedleft\arraybackslash}X>{\raggedleft\arraybackslash}X>{\raggedleft\arraybackslash}X@{}}
\toprule
\textbf{Cell} & $\boldsymbol{\alpha}$ & $\boldsymbol{S_{80}}$ \textbf{(pairs in judge's direction)} & \textbf{Paired difference from} $\boldsymbol{\alpha=1}$ & \textbf{KL} & \textbf{Validity} \\
\midrule
\st{Llama}{FT}{syllable}{sloth--lemur} & 0 & $-0.034$ (2/8) & $-0.107$, $p=0.023$ & 1.04 & 0.822 \\
 & 1 & $+0.073$ & --- & 1.13 & 0.820 \\
 & 2 & $+0.118$ (8/8) & $+0.044$, $p=0.24$ & 1.06 & 0.814 \\
\midrule
\st{Llama}{SP}{number}{sloth--lemur} & 0 & $+0.004$ (5/8) & $-0.084$, $p=0.031$ & 0.89 & 0.987 \\
 & 1 & $+0.089$ & --- & 0.92 & 0.987 \\
 & 2 & $+0.045$ & $-0.044$, $p=0.13$ & --- & --- \\
\bottomrule
\end{tabularx}

%% file: tabs/time.tex
\begin{tabularx}{\textwidth}{@{}>{\raggedright\arraybackslash}p{0.13\textwidth}>{\raggedleft\arraybackslash}X>{\raggedleft\arraybackslash}X>{\raggedleft\arraybackslash}X>{\raggedleft\arraybackslash}X>{\raggedleft\arraybackslash}X@{}}
\toprule
\textbf{Cell} & \textbf{Validity, step 2 / 48 / 80} & \textbf{Lowest run, step 80} & \textbf{Reward-signal share, step 10 / 40 / 80} & \textbf{First step with signal share $<0.5$} & \textbf{Step reaching 50\% / 80\% of $S_{80}$}\\
\midrule
\addlinespace[3pt]\rowcolor{black!7}\multicolumn{6}{@{}l}{\textbf{FT route --- fine-tuned judges}}\\
\addlinespace[2pt]\multicolumn{6}{@{}l}{\textit{syllable \textperiodcentered{} Llama}}\\
\hspace{0.6em}elephant--tiger & 0.530 / 0.961 / 0.981 & 0.973 & 0.97 / 0.92 / 0.61 & never & ---\\
\hspace{0.6em}jerboa--badger & 0.523 / 0.958 / 0.980 & 0.964 & 0.97 / 0.91 / 0.78 & never & ---\\
\hspace{0.6em}otter--beaver & 0.542 / 0.967 / 0.980 & 0.968 & 0.97 / 0.90 / 0.72 & never & 64 / 64\\
\hspace{0.6em}shrew--cuttlefish & 0.526 / 0.959 / 0.981 & 0.971 & 0.97 / 0.95 / 0.71 & never & 32 / 48\\
\hspace{0.6em}sloth--lemur & 0.525 / 0.965 / 0.984 & 0.972 & 0.97 / 0.91 / 0.67 & never & 48 / 48\\
\hspace{0.6em}warbler--slug & 0.535 / 0.962 / 0.980 & 0.967 & 0.98 / 0.95 / 0.71 & never & 80 / 80\\
\addlinespace[2pt]\multicolumn{6}{@{}l}{\textit{syllable \textperiodcentered{} Qwen}}\\
\hspace{0.6em}elephant--tiger & 0.809 / 0.996 / 0.999 & 0.998 & 1.00 / 0.28 / 0.16 & 36 & ---\\
\hspace{0.6em}jerboa--badger & 0.971 / 0.996 / 0.991 & 0.953 & 1.00 / 0.56 / 0.31 & 46 & 8 / 8\\
\hspace{0.6em}otter--beaver & 0.965 / 0.998 / 0.997 & 0.973 & 1.00 / 0.62 / 0.35 & 48 & ---\\
\hspace{0.6em}sloth--lemur & 0.953 / 0.990 / 0.991 & 0.956 & 1.00 / 0.56 / 0.37 & 46 & ---\\
\hspace{0.6em}warbler--slug & 0.875 / 0.995 / 0.994 & 0.980 & 1.00 / 0.77 / 0.44 & 68 & ---\\
\addlinespace[2pt]\multicolumn{6}{@{}l}{\textit{number \textperiodcentered{} Llama}}\\
\hspace{0.6em}elephant--tiger & 0.980 / 0.998 / 0.998 & 0.995 & 1.00 / 0.79 / 0.54 & never & ---\\
\hspace{0.6em}jerboa--badger & 0.983 / 0.998 / 0.999 & 0.996 & 1.00 / 0.83 / 0.63 & never & ---\\
\hspace{0.6em}otter--beaver & 0.984 / 0.998 / 0.998 & 0.994 & 1.00 / 0.87 / 0.70 & never & ---\\
\hspace{0.6em}shrew--cuttlefish & 0.982 / 0.997 / 0.998 & 0.993 & 1.00 / 0.85 / 0.75 & never & ---\\
\hspace{0.6em}sloth--lemur & 0.982 / 0.997 / 0.998 & 0.994 & 1.00 / 0.88 / 0.67 & never & ---\\
\hspace{0.6em}warbler--slug & 0.985 / 0.996 / 0.998 & 0.996 & 1.00 / 0.80 / 0.58 & never & ---\\
\addlinespace[2pt]\multicolumn{6}{@{}l}{\textit{number \textperiodcentered{} Qwen}}\\
\hspace{0.6em}elephant--tiger & 1.000 / 0.999 / 0.998 & 0.974 & 0.94 / 0.60 / 0.43 & 68 & ---\\
\hspace{0.6em}jerboa--badger & 1.000 / 0.999 / 0.986 & 0.862 & 0.93 / 0.44 / 0.14 & 36 & 24 / 24\\
\hspace{0.6em}otter--beaver & 1.000 / 0.997 / 1.000 & 0.999 & 0.94 / 0.28 / 0.36 & 34 & ---\\
\hspace{0.6em}shrew--cuttlefish & 1.000 / 0.994 / 0.997 & 0.975 & 0.93 / 0.54 / 0.26 & 42 & ---\\
\hspace{0.6em}sloth--lemur & 1.000 / 0.999 / 1.000 & 0.999 & 0.94 / 0.57 / 0.31 & 44 & 32 / 48\\
\hspace{0.6em}warbler--slug & 1.000 / 0.999 / 0.997 & 0.959 & 0.95 / 0.45 / 0.28 & 38 & 24 / 24\\
\addlinespace[3pt]\rowcolor{black!7}\multicolumn{6}{@{}l}{\textbf{SP route --- persona judges}}\\
\addlinespace[2pt]\multicolumn{6}{@{}l}{\textit{syllable \textperiodcentered{} Llama}}\\
\hspace{0.6em}elephant--tiger & 0.535 / 0.954 / 0.975 & 0.962 & 0.98 / 1.00 / 0.96 & never & ---\\
\hspace{0.6em}jerboa--badger & 0.526 / 0.956 / 0.974 & 0.963 & 0.97 / 0.98 / 0.93 & never & ---\\
\hspace{0.6em}otter--beaver & 0.529 / 0.957 / 0.975 & 0.946 & 0.98 / 0.98 / 0.95 & never & 48 / 64\\
\hspace{0.6em}shrew--cuttlefish & 0.535 / 0.951 / 0.973 & 0.959 & 0.98 / 0.98 / 0.93 & never & 8 / 32\\
\hspace{0.6em}sloth--lemur & 0.530 / 0.958 / 0.975 & 0.950 & 0.97 / 0.99 / 0.95 & never & ---\\
\hspace{0.6em}warbler--slug & 0.520 / 0.957 / 0.968 & 0.941 & 0.97 / 0.99 / 0.96 & never & 24 / 24\\
\addlinespace[2pt]\multicolumn{6}{@{}l}{\textit{syllable \textperiodcentered{} Qwen}}\\
\hspace{0.6em}elephant--tiger & 0.811 / 0.985 / 0.937 & 0.746 & 0.99 / 0.55 / 0.53 & 43 & 32 / 48\\
\hspace{0.6em}jerboa--badger & 0.972 / 0.994 / 0.993 & 0.971 & 1.00 / 0.73 / 0.44 & 53 & ---\\
\hspace{0.6em}otter--beaver & 0.967 / 0.994 / 0.987 & 0.954 & 1.00 / 0.71 / 0.47 & 67 & 24 / 32\\
\hspace{0.6em}sloth--lemur & 0.952 / 0.992 / 0.986 & 0.955 & 0.99 / 0.64 / 0.45 & 49 & 32 / 64\\
\hspace{0.6em}warbler--slug & 0.883 / 0.988 / 0.981 & 0.931 & 1.00 / 0.63 / 0.50 & 54 & 16 / 24\\
\addlinespace[2pt]\multicolumn{6}{@{}l}{\textit{number \textperiodcentered{} Llama}}\\
\hspace{0.6em}elephant--tiger & 0.983 / 0.997 / 0.999 & 0.998 & 1.00 / 0.98 / 0.88 & never & ---\\
\hspace{0.6em}jerboa--badger & 0.980 / 0.997 / 0.999 & 0.998 & 1.00 / 0.97 / 0.87 & never & ---\\
\hspace{0.6em}otter--beaver & 0.980 / 0.996 / 0.998 & 0.994 & 1.00 / 0.97 / 0.89 & never & ---\\
\hspace{0.6em}shrew--cuttlefish & 0.984 / 0.996 / 0.999 & 0.998 & 1.00 / 0.98 / 0.91 & never & ---\\
\hspace{0.6em}sloth--lemur & 0.980 / 0.995 / 0.999 & 0.998 & 1.00 / 0.98 / 0.89 & never & 32 / 48\\
\hspace{0.6em}warbler--slug & 0.983 / 0.996 / 0.998 & 0.996 & 1.00 / 0.98 / 0.90 & never & 32 / 80\\
\addlinespace[2pt]\multicolumn{6}{@{}l}{\textit{number \textperiodcentered{} Qwen}}\\
\hspace{0.6em}elephant--tiger & 1.000 / 1.000 / 1.000 & 1.000 & 0.94 / 0.75 / 0.67 & never & 16 / 48\\
\hspace{0.6em}jerboa--badger & 1.000 / 1.000 / 1.000 & 1.000 & 0.96 / 0.75 / 0.67 & never & ---\\
\hspace{0.6em}otter--beaver & 1.000 / 1.000 / 1.000 & 1.000 & 0.97 / 0.78 / 0.70 & never & ---\\
\hspace{0.6em}shrew--cuttlefish & 1.000 / 1.000 / 1.000 & 1.000 & 0.97 / 0.78 / 0.68 & never & ---\\
\hspace{0.6em}sloth--lemur & 1.000 / 1.000 / 1.000 & 1.000 & 0.96 / 0.77 / 0.68 & never & ---\\
\hspace{0.6em}warbler--slug & 1.000 / 1.000 / 1.000 & 1.000 & 0.95 / 0.76 / 0.68 & never & ---\\
\bottomrule
\end{tabularx}

%% file: tabs/judge.tex
\begin{tabularx}{\textwidth}{@{}>{\raggedright\arraybackslash}p{0.15\textwidth}>{\raggedleft\arraybackslash}X>{\raggedleft\arraybackslash}X>{\raggedleft\arraybackslash}X>{\raggedleft\arraybackslash}X>{\raggedleft\arraybackslash}X>{\raggedleft\arraybackslash}X>{\raggedleft\arraybackslash}X@{\hspace{10pt}}>{\raggedright\arraybackslash}p{0.15\textwidth}@{}}
\toprule
\textbf{Judge} & \textbf{Emission $\Lambda$ (mean)} & \textbf{$\Lambda$ lower bound $>0$} & \textbf{Trait alignment with own-model judge ($r$)} & \textbf{$|$Absorption$|$} & \textbf{Absorption sign as own model} & \textbf{Judge term} & \textbf{Absorption vs $S_{80}$ ($r$)} & \textbf{Transfer}\\
\midrule
\addlinespace[3pt]\rowcolor{black!7}\multicolumn{9}{@{}l}{\textbf{FT route --- fine-tuned judges}}\\
\addlinespace[2pt]\multicolumn{9}{@{}l}{\textit{Student Llama-3.2-1B}}\\
\hspace{0.6em}Own model (main grid) & +0.117 & 12/12 & --- & 0.170 & --- & 0.113 & +0.38 & 4 of 12 established\\
\hspace{0.6em}Other family, 1B class & +0.128 & 12/12 & +0.03 & 0.066 & 8/12 & 0.037 & $-$0.05 & none established\\
\hspace{0.6em}Same family, 8B & +0.103 & 12/12 & +0.07 & 0.090 & 9/12 & 0.072 & --- & 0 of 2 trained cells\\
\hspace{0.6em}Other family, 8B & +0.214 & 12/12 & +0.02 & 0.047 & 9/12 & 0.080 & --- & 0 of 2 trained cells\\
\addlinespace[2pt]\multicolumn{9}{@{}l}{\textit{Student Qwen3-1.7B}}\\
\hspace{0.6em}Own model (main grid) & +0.136 & 10/11 & --- & 0.100 & --- & 0.119 & +0.59 & 4 of 11 established\\
\hspace{0.6em}Other family, 1B class & +0.138 & 10/11 & +0.03 & 0.084 & 7/11 & 0.095 & +0.40 & none established\\
\hspace{0.6em}Same family, 8B & +0.223 & 11/11 & +0.01 & 0.092 & 7/11 & 0.062 & --- & 0 of 2 trained cells\\
\hspace{0.6em}Other family, 8B & +0.111 & 9/11 & +0.04 & 0.144 & 9/11 & 0.119 & --- & 0 of 2 trained cells\\
\addlinespace[3pt]\rowcolor{black!7}\multicolumn{9}{@{}l}{\textbf{SP route --- persona judges}}\\
\addlinespace[2pt]\multicolumn{9}{@{}l}{\textit{Student Llama-3.2-1B}}\\
\hspace{0.6em}Own model (main grid) & +0.231 & 12/12 & --- & 0.209 & --- & 0.209 & +0.38 & 5 of 12 established\\
\hspace{0.6em}Other family, 1B class & +0.274 & 12/12 & +0.02 & 0.040 & 7/12 & 0.028 & $-$0.05 & none established\\
\hspace{0.6em}Same family, 8B & +0.262 & 12/12 & +0.07 & 0.067 & 7/12 & 0.038 & --- & 0 of 2 trained cells\\
\hspace{0.6em}Other family, 8B & +0.253 & 12/12 & +0.06 & 0.052 & 7/12 & 0.047 & --- & 0 of 2 trained cells\\
\addlinespace[2pt]\multicolumn{9}{@{}l}{\textit{Student Qwen3-1.7B}}\\
\hspace{0.6em}Own model (main grid) & +0.366 & 11/11 & --- & 0.133 & --- & 0.080 & +0.59 & 5 of 11 established\\
\hspace{0.6em}Other family, 1B class & +0.349 & 11/11 & +0.04 & 0.093 & 5/11 & 0.094 & +0.40 & none established\\
\hspace{0.6em}Same family, 8B & +0.413 & 11/11 & +0.12 & 0.067 & 6/11 & 0.072 & --- & 0 of 2 trained cells\\
\hspace{0.6em}Other family, 8B & +0.287 & 11/11 & +0.12 & 0.130 & 10/11 & 0.090 & --- & 0 of 2 trained cells\\
\bottomrule
\end{tabularx}

%% file: sec/si_notes.tex
\snote{The between-judge contrast at the first step}{S-sn:step}

In the between-judge contrast of equation~\eqref{eq:M3d} the shared score cancels except in the rescaling term $\tfrac12\widetilde{\bar r}\,(\sigma_{T,g}^{-1}-\sigma_{E,g}^{-1})$, which vanishes when the two judges agree and is itself set by the trait. On the start-policy candidate pools, $d$ correlates with its trait part at 0.87--0.93 in six pools and at 0.63--0.66 in the two \cf{Qwen}{FT}{number} pools measured. Splitting the covariance with $k$ the same way, the direct term carries the start-policy alignment in six of the eight pools; the rescaling term offsets it by up to 45\%, is the whole of the small alignment of \st{Llama}{SP}{number}{warbler--slug} ($+0.033$) and nearly cancels the direct term in \st{Qwen}{FT}{number}{jerboa--badger}.

\snote{The established set}{S-sn:grid}

Twenty-four cells reach one-sided $p < 0.05$ over all their seed pairs, 22 in the judge's direction and two against it: \st{Llama}{SP}{syllable}{elephant--tiger} (16 seed pairs, $-0.041$, 12 of 16 negative, one-sided $p = 0.036$) and \st{Llama}{SP}{number}{elephant--tiger} (16 seed pairs, $-0.039$, one-sided $p = 0.028$). Under the main criterion these two drop (two-sided $p = 0.071$, $q = 0.149$; $p = 0.057$, $q = 0.137$), as do four cells in the judge's direction, \st{Llama}{SP}{syllable}{sloth--lemur}, \st{Qwen}{FT}{syllable}{otter--beaver}, \st{Qwen}{FT}{syllable}{sloth--lemur} and \st{Qwen}{SP}{number}{otter--beaver} (two-sided $p = 0.063$ to 0.082). Seventeen of the eighteen established cells reach $q$ below 0.05; the strongest are \st{Qwen}{SP}{syllable}{warbler--slug} and \st{Qwen}{FT}{number}{warbler--slug}, at $q = 0.0007$.

The established set is stable across tests (\edtab{\ref{tab:ed-further}}). Fisher's combination of each cell's base eight seed pairs with its further pairs meets the criterion in 19 cells, ten of them at $q$ below 0.05: 17 of the eighteen, together with \st{Qwen}{SP}{number}{shrew--cuttlefish} in the judge's direction and \st{Llama}{SP}{syllable}{elephant--tiger} against it, in the direction its negative start-policy absorption gives. On their further pairs alone, 11 of the 35 cells with more than eight seed pairs meet the criterion, all in the judge's direction and ten of them also established on all pairs. The estimate on the further pairs is positive in 27 of the 35 (two-sided sign test $p = 0.002$, using no effect magnitude) and averages $+0.020$ with the 35 cells weighted equally. With eight further pairs the smallest attainable two-sided $p$ is 2/256, which five cells reach, so the smallest $q$ in that family is 0.055. One-sample $t$ and Wilcoxon tests establish sixteen cells at $q < 0.05$ under either, against seventeen under the sign-flip test. Where sign patterns are drawn by Monte Carlo (the six cells with 24 or 28 seed pairs), no $p$ lies closer than 0.024 to the criterion. Bootstrap intervals over seed pairs exclude zero in the eighteen established cells and also in two that are not established, \st{Llama}{SP}{number}{elephant--tiger} and \st{Qwen}{SP}{number}{otter--beaver}; the exact test is the more conservative of the two.

Rescoring all twelve animal names in the trained students shows that the margin moves through one name of the judge's pair. In \st{Qwen}{SP}{syllable}{warbler--slug} the T name (warbler) is the only one of the twelve that moves toward the T-trained student ($+0.042$, against $-0.011$ to $-0.043$ for the other eleven) and exceeds the mean of the ten names outside the pair by $+0.068$ (7 of 8 seed pairs, $p = 0.016$). In \st{Qwen}{FT}{number}{warbler--slug} it is first of the twelve in all eight seed pairs ($+0.054$ above the mean of the other ten, $p = 0.008$). In \st{Qwen}{SP}{number}{elephant--tiger} the specific movement is on the other name: tiger falls below the mean of the ten by $-0.019$ in all eight seed pairs ($p = 0.008$), while elephant does not separate from them.

Transfer is not monotone in the start-policy stages. Among cells without an established result, \st{Llama}{SP}{number}{shrew--cuttlefish} holds the largest positive absorption in the grid ($+0.421$) and gives $+0.007$ over 16 seed pairs (one-sided $p = 0.27$), whereas \st{Qwen}{FT}{number}{sloth--lemur}, whose start-policy emission is at the level of judge-instance noise ($+0.004$ [$-0.005$, $+0.013$]), and \st{Qwen}{FT}{number}{jerboa--badger}, with the fourth-smallest emission in the grid ($+0.029$ [$+0.018$, $+0.042$]), both transfer (16 seed pairs each).

Each established cell rests on at least 16 seed pairs. Seventeen of the 28 cells without an established result have 16 to 28 seed pairs, not only eight.

\snote{Words implying an animal, and the concealment audit}{S-sn:leak}

The carrier rules block the names themselves. The syllable rule lets through 0.000--0.016\% of completions containing a name in Llama and none among the 38{,}000--40{,}500 scored completions per Qwen seed pair; in the syllable carrier 0.06--0.52\% of all recorded completions contain a name as a substring, under the two judges alike (0.21\% and 0.16\% of 327{,}680 each), and the rule rejects almost all of them.

Words that imply an animal are blocked by neither rule, and one cell uses them. In \st{Qwen}{SP}{syllable}{warbler--slug}, 13--18\% of completions in the first 16 steps contain words implying the slug (\emph{squishy}, \emph{squish}, \emph{goo}, \emph{gooey}, \emph{slithery}), more under the slug-preferring judge (16--18\% against 13--15\%) and for longer (2.4--3.2\% against 0.3--1.5\% in steps 16--47), before they disappear from step 48; from then on the cell's generations meet concealment as defined here, and its endpoint carries the earlier steps with it. The same words are a habit of the Qwen syllable start policy regardless of the trait pair, so they are as frequent early in \st{Qwen}{FT}{syllable}{warbler--slug} (12.6\% against 12.4\%), but there the two judges do not reward them differently and the cell is not established.

The leak is the one place where three conditions meet, each measurable on the start-policy candidates, in the judge-term-times-student-term structure of absorption at the level of word meaning. The student must write the words often: 17.1\% of candidates contain a slug-implying word, against at most 0.3\% for any other animal in either model. The preference margin must be tied to them: within a prompt, moving the margin toward slug raises the probability of those candidates ($k$ lower by 0.44 standard deviations, in 96\% of prompts), and warbler-implying words are tied the other way ($+0.42$). And the two judges must score them differently: the slug persona favours slug-implying candidates over the others in the same prompt by 0.147 standard deviations of $d$, with the same sign under all eight wordings, whereas the warbler persona does not favour warbler-implying words ($-0.007$ to $+0.025$) and the fine-tuned judge barely separates slug-implying ones ($-0.025$). Removing those candidates lowers the cell's start-policy absorption only from $+0.333$ to $+0.308$; they carry 26\% of its covariance. On the recorded training samples the share is similar, 15.5\% over the whole run (20--24\% in the first steps and none after step 48, when the late alignment of $+0.38$ under the warbler judge is carried entirely by other candidates), so most of the cell's transfer runs through other channels. Separately, Qwen syllable completions drift during training toward lists of real English words, mostly fruit names and number words, unrelated to the trait pairs.

Without \st{Qwen}{SP}{syllable}{warbler--slug}, with the false-discovery rate recomputed over the remaining 45 cells, the other seventeen established cells remain established, sixteen of them at $q$ below 0.05, with mean $S_{80}$ $+0.032$ against $+0.035$ for all eighteen; the largest cell becomes \st{Llama}{FT}{syllable}{sloth--lemur}, whose completions carry no word implying an animal. No stage of the chain rests on this cell.

\snote{Emission: carriers and per-animal biases}{S-sn:emis}

Computed on the group-standardized advantages with the same pairings, $\Lambda$ correlates with the score-based value at $+0.997$ across cells, so standardization does not remove the trait signal. The carrier sets most of its size: the spread of the score difference between judges given different animals exceeds that between instances given the same animal by about 40\% in the syllable cells (mean $\Lambda$ $+0.33$) and by about 10\% in the number cells ($+0.10$), and the five smallest values are number cells. The trait pair adds a consistent component beyond the configuration (9\% of the variance, within-configuration permutation $p < 0.0002$): warbler--slug is above its configuration mean in 8 of 8 configurations, otter--beaver and elephant--tiger in 1 of 8. Beyond its presence, emission does not order transfer: within configurations, cells above and below the median emission do not differ in $S_{80}$ ($p = 0.40$).

In the 22 syllable cells the letter coefficient of the judges' score difference, its match with the letters of the judge's own animal name, ranges from $+0.023$ to $+0.718$, and that of the training instance's standardized advantage from $-0.068$ to $+0.671$ (positive in 21; \edtab{\ref{tab:ed-chan}}). In the substitution test, replacing a position that holds a letter absent from both names by a letter unique to the T name gives $\Delta d$ $+0.270$ and $\Delta k$ $+19.1$ in \st{Llama}{FT}{syllable}{otter--beaver}, positive in 0.69 and 0.75 of 400 matched pairs ($t$ $+8.1$ and $+12.4$).

In the number carrier, the persona ordering of the twelve animals follows how strongly each persona prefers larger numbers ($r$ $+0.72$ on the Llama candidate pool, $+0.92$ on the Qwen pool), is unrelated to the animals' real sizes and differs between models (rank correlation $-0.10$ between the two orders). Fine-tuned judges have no frame to rescore, so each student's number pool is scored by the fine-tuned judges of all twelve animals in all eight fine-tuning seeds, with the student's own weights as baseline. The resulting per-animal effect is far more stable than a persona's: 98.0\% (Qwen) and 98.8\% (Llama) of its variance lies between animals rather than between seeds of the same animal, with split-half reliability over seeds of 0.996 and 0.998. Its ordering by association with number size is reproducible across seeds ($+0.80$ and $+0.73$) but is not the persona ordering of the same student model (rank correlation $+0.14$ in Qwen), so each route carries its own ordering. Over the six trait pairs, the grid's $\Lambda$ follows the spread of the difference between the two animals' effect vectors, $\mathrm{sd}(\hat e_T-\hat e_E)$, with rank correlation $+1.00$ on the Qwen pool and $+0.77$ on the Llama pool, and the trait-pair component of the emission decomposition across judge families (Methods) follows it at $+0.83$ and $+0.71$. The dissimilarity measure that also serves for personas, the negative correlation of the two effect vectors, does not generalize: its rank correlation with $\Lambda$ is $+0.77$ on the Qwen pool and $-0.77$ on the Llama pool.

Emission also forms on the samples the policy draws. In the four \cf{Qwen}{FT}{number} cells, in-training emission does not separate them at the start (elephant--tiger 0.058 above jerboa--badger 0.036, of which only jerboa--badger transfers) but by the end ranks them in the order of their transfer (warbler--slug, then jerboa--badger, then otter--beaver and elephant--tiger about equal): jerboa--badger's trait signal grows fourfold on the T run's samples and twofold on the E run's, while elephant--tiger's shrinks. In \st{Qwen}{FT}{number}{sloth--lemur}, rescoring the training samples of eight seed pairs with all sixteen judges, $\Lambda$ on the T run's samples is $+0.018$ [$+0.013$, $+0.025$] over calls 0--16 and $+0.050$ [$+0.031$, $+0.071$] over calls 56--79; on the E run's samples it rises from $+0.026$ to $+0.049$ in mid-training and stays positive ($+0.030$ [$+0.021$, $+0.041$] at the end). For the training instance alone, with the same-animal spread taken over its own pairings, it rises from 0.000 on the start-policy pool to $+0.065$ [$+0.050$, $+0.081$] and $+0.089$ [$+0.069$, $+0.109$] in mid-training on the two runs' samples.

In two cells the judges' trait difference is partly a per-token likelihood offset of the judge pair (correlation with completion length $-0.45$ in \st{Llama}{FT}{syllable}{jerboa--badger} and $+0.38$ in \st{Qwen}{FT}{syllable}{warbler--slug}); neither cell is established.

On the group-standardized advantages $\Lambda$ is positive in the same 45 cells as on the scores, and only in \st{Qwen}{FT}{number}{sloth--lemur} does its interval contain zero. In the letter-substitution test the link between letters, the judges' trait difference and the student's $k$ is causal: strongly over 400 matched pairs in \st{Llama}{FT}{syllable}{otter--beaver}, with the same sign in \st{Llama}{SP}{syllable}{otter--beaver} and \st{Llama}{SP}{syllable}{warbler--slug}. In the number carrier, twelve animal personas rescoring one pool in the same frame give each animal one number, its judge-specific bias along the direction common to animals; a pair's judge term, a covariance, is the difference of its animals' numbers, so one ordering fixes its sign in all 66 pairings.

\snote{Judge framings and matched judges}{S-sn:frame}

Rescoring each pool with the judge's frame changed leaves the persona size preferences largely in place. Between \emph{You love X} and \emph{You hate X}, the twelve per-animal values correlate at $r$ $+0.86$ on the Llama pool (signs agreeing for 10 of 12 animals) and $+0.75$ on the Qwen pool (9 of 12), and the pairwise emission direction correlates between the two frames at $+0.72$ to $+0.89$ across the six Llama pairs and $+0.63$ to $+0.82$ across the six Qwen pairs. The neutral mention (\emph{You have read about X}) keeps the per-animal order in Qwen ($r$ $+0.75$, 9 of 12) and weakens it in Llama ($+0.40$, 8 of 12), while the pairwise direction stays between $+0.50$ and $+0.86$ across the Llama pairs and between $+0.43$ and $+0.61$ across the Qwen pairs.

Over 36 animals per model, once the component common to all animals is removed, the retained direction of an animal's own contribution averages $+0.77$ under inversion and $+0.73$ under the neutral mention in Llama, and $+0.69$ and $+0.44$ in Qwen, with the twelve grid animals and the other 24 in agreement ($+0.79$ against $+0.76$, $+0.71$ against $+0.74$, $+0.72$ against $+0.67$, $+0.49$ against $+0.42$). Survival under inversion follows the size of the animal's own contribution relative to the common component---one property, since its size and its resemblance to the common component correlate at $-0.95$ to $-0.98$---at $r$ $+0.44$ overall and $+0.49$ in each model separately; name token count adds a partial $+0.28$. Under the neutral mention this holds in Qwen ($+0.46$) but not in Llama ($+0.08$), where survival is uniformly high ($+0.73$, sd 0.13). At the level of the models, animal effects are 93--100\% common component in both; Llama carries the common component and the idiosyncratic scale unchanged across framings, whereas Qwen's neutral frame alters the common component ($r$ $+0.87$) and inflates the idiosyncratic part by 1.28, the part that does not survive.

In the training ladder (six cells, same seeds, only the frame replaced), love transfers in five; in \st{Qwen}{SP}{number}{shrew--cuttlefish} it is not established. Transfer is established under inversion in \st{Llama}{SP}{number}{warbler--slug}, \st{Qwen}{SP}{syllable}{otter--beaver} and \st{Qwen}{SP}{syllable}{warbler--slug}, and under the neutral mention in \st{Llama}{SP}{syllable}{warbler--slug} and \st{Qwen}{SP}{syllable}{warbler--slug}, with \st{Llama}{SP}{number}{warbler--slug} at $p = 0.059$. A framing counts as lost where its transfer is significantly smaller than love's in the paired comparison. Inversion is lost alone in \st{Llama}{SP}{syllable}{warbler--slug}, the best-powered cell at 24 seed pairs ($+0.023$ love, $+0.002$ inversion, $+0.021$ neutral mention; inversion against love $-0.021$, $p = 0.036$), and together with the neutral mention in \st{Llama}{SP}{number}{sloth--lemur} ($+0.089$, $+0.016$, $+0.026$; $-0.073$ and $-0.063$, $p = 0.023$ and 0.008). The neutral mention alone is lost in \st{Qwen}{SP}{syllable}{otter--beaver} ($+0.038$, $+0.053$, $+0.002$; neutral mention against love $-0.037$, $p = 0.008$) and \st{Qwen}{SP}{number}{shrew--cuttlefish} ($+0.021$, $+0.009$, $-0.006$; $-0.027$, $p = 0.039$). Neither is lost in \st{Llama}{SP}{number}{warbler--slug} and \st{Qwen}{SP}{syllable}{warbler--slug}. Validity stays at or above 0.909 throughout. That an evaluative stance supplies the size therefore holds in the cells where inversion fails and is contradicted where the neutral mention fails instead, and one configuration can hold a trait pair of each kind.

Which framing a cell needs is predicted by emission, not by absorption. Splitting the reframed judges' between-judge difference into its component along the love-frame difference and a residual, the retained direction orders the framings within these cells ($r$ $+0.55$, $p = 0.030$), and so does the size of that component ($+0.44$, $p = 0.035$), while absorption does not ($p = 0.11$). This resolves the two cases that look contradictory on absorption alone. In \st{Llama}{SP}{number}{warbler--slug} inversion has absorption $-0.042$, the wrong sign, yet keeps direction 0.860 and almost all of the size of the love-direction component (0.031 to 0.028): the negative correlation comes from the residual, which is anti-aligned with $k$, and inversion transfers. In \st{Qwen}{SP}{syllable}{otter--beaver} the neutral mention has absorption $+0.136$ against love's $+0.174$, which looks equivalent, but its love-direction component shrinks by 44\% (0.083 to 0.047), with the lowest retained direction of any cell (0.768), and the neutral mention fails. Absorption, a correlation over the whole difference, mixes the retained component with the residual in the first case and discards the scale that separates the framings in the second. The same measure holds off the start policy: on the logged training samples of six cells, rescored with the love judge, the retained direction predicts per-framing $S$ within cells ($r$ $+0.59$, within-cell permutation $p = 0.023$, 17 framing--cell points), whereas absorption on the same samples does not ($+0.50$, $p = 0.22$) and the first-order GRPO push is unrelated ($+0.06$, $p = 0.77$). Robustness to reframing is fixed at emission: the retained direction measured on the judges' scores matches that measured on absorption across five cells and two framings ($r$ $+0.72$ over ten points), and inversion and neutral mention move together ($r$ $+0.97$), so robustness behaves as one number per cell.

Matched persona judges, naming both animals in the same positions and exchanging only the valence words, emit $\Lambda$ $+0.019$ [$+0.017$, $+0.021$] in \st{Llama}{SP}{number}{sloth--lemur}, $+0.036$ [$+0.033$, $+0.039$] in \st{Llama}{SP}{number}{warbler--slug} and $+0.008$ [$+0.004$, $+0.013$] in \st{Qwen}{SP}{number}{elephant--tiger}, against $+0.103$, $+0.227$ and $+0.124$ for the single-animal wordings on the same candidates. Across paraphrases the single-animal between-judge difference is stable (split-half correlation $+0.81$ to $+0.93$), whereas the matched one is largely wording-specific ($+0.09$ to $+0.54$), a third to a half of its size and unaligned with it ($-0.22$ to $+0.16$ across the three cells); a control that changes only a pronoun (\emph{you always prefer the former} against \emph{the latter}) agrees no better ($-0.16$ to $+0.22$). On the start-policy pool, then, the judges' score difference is set mainly by which animal is named in the judge's context, and the attitude expressed toward it sets its amplitude rather than its direction: an entity-conditioned bias over carriers that cannot name the animal. Trained end to end over 24 seed pairs each, the matched judges give $S_{80}$ $+0.027$ [$-0.004$, $+0.058$] in \st{Llama}{SP}{number}{sloth--lemur} and $+0.008$ [$-0.007$, $+0.023$] in \st{Llama}{SP}{number}{warbler--slug}, pooled $+0.018$ [$+0.000$, $+0.035$] (29 of 48 positive, two-sided $p = 0.057$); where the original wording was trained on the same seeds it gives $+0.036$ against $+0.058$, a paired difference of $-0.022$ ($p = 0.12$). In the Qwen cell the matched judge barely moves the policy (KL 0.06 against 0.8; estimate $-0.013$).

Judges fine-tuned on comparisons naming both animals in the same words, in \st{Llama}{FT}{syllable}{otter--beaver}, \st{Llama}{FT}{syllable}{shrew--cuttlefish} and \st{Llama}{FT}{syllable}{warbler--slug}, emit $\Lambda$ $+0.05$ to $+0.08$ against $+0.19$ to $+0.21$ for the original judges, with retained direction $-0.07$ to $+0.30$ and absorption $-0.13$ to $-0.01$. Trained on the same seeds, they move the policy (KL 0.84 to 1.03) and give $S_{80}$ $-0.012$, $-0.008$ and $-0.031$ (pooled $-0.017$ over 24 seed pairs, two-sided $p = 0.054$), against $+0.022$, $+0.029$ and $+0.018$ for the original judges on the same seed pairs (per cell $-0.033$ [$-0.054$, $-0.011$], $-0.037$ [$-0.063$, $-0.014$] and $-0.049$ [$-0.093$, $+0.003$]).

Lexical association fixes the direction of the framed judges' score difference; whether the evaluative stance supplies its size varies with the trait pair within a model and carrier. The six reframed cells give the direction of the link between retained direction and per-framing transfer, not its size. In \st{Llama}{FT}{syllable}{otter--beaver}, \st{Llama}{FT}{syllable}{shrew--cuttlefish} and \st{Llama}{FT}{syllable}{warbler--slug}, the single-animal fine-tuning data gave the judge the carrier ranking it transferred through; as the single-animal and comparison fine-tuning sets also differ in item structure, this identifies the construction, not lexical exposure alone.

\snote{Absorption: judge instances, trait pairs and paths}{S-sn:abs}

Across the 35 cells with eight judge instances on one pool, instance-averaged absorption predicts $S$ in both routes (FT $+0.36$, SP $+0.54$; together $r = +0.49$). As $S_{80}$ over eight seed pairs has reliability 0.50--0.60 against 0.95 for the predictor, the observable correlation is capped near 0.74. Cross-instance absorption, which scores the T and E judges with different instances and so removes what one wording or seed contributes to both, gives $r = +0.48$ ($p = 0.011$). The excess of the training instance over the other seven is uninformative overall ($r$ $+0.08$, $p = 0.73$). In the 23 SP cells the excess tends to carry into transfer ($r$ $+0.53$, $p = 0.056$) and the training wording alone is the best single predictor ($r$ $+0.63$, $p = 0.009$); the SP relation holds when any one cell is dropped (coefficient $+0.38$ to $+0.61$). Over the 23 FT cells the training seed does not predict transfer ($r$ $-0.18$, $p = 0.42$). Over the 12 Llama FT cells it runs against transfer ($-0.66$, $p = 0.017$), as does its excess over the other seven ($-0.70$, $p = 0.012$), a relation confined to the two Llama FT configurations. The raw alignment of the training instance does not survive dropping single trait pairs, one of which can reverse a configuration's correlation, so the route difference is stated on the instance-specific excess across cells.

The pooled SP correlation over 40 trait pairs (twelve of them in \cf{Qwen}{SP}{syllable}) is stable: dropping any one pair leaves it between $+0.567$ and $+0.778$, dropping any one configuration between $+0.612$ and $+0.682$, and the worst of all pairs of exclusions gives $+0.528$. It has two separable parts. The rank of the training wording among the pair's eight instances predicts $S$ at $r$ $+0.59$ ($p = 0.0013$; $+0.53$ to $+0.67$ leaving out any one pair) and the instance mean at $+0.50$ ($p = 0.0065$), each surviving control for the other (partial $r$ $+0.51$, $p = 0.006$, and $+0.39$, $p = 0.038$); the spread across instances is unrelated ($-0.24$). The rank part is a property of the pairs, not an effect of the wording: twelve pairs (six \cf{Llama}{SP}{number}, six \cf{Qwen}{SP}{number}) were trained on the same eight seeds under the pair's most and least absorbing wording, and transfer is higher under the more absorbing wording in 3 of 12 pairs (sign test $p = 0.98$; mean difference $-0.011$); within a pair $S$ does not follow absorption across the three wordings ($r$ $-0.02$ over 36 conditions; within-pair permutation $p = 0.53$). Training statistics are the same across wordings (KL 0.89--1.05, final reward $-8.5$ to $-9.7$, completion length 24.0--24.1), so the wording does not change the scale of the update either.

On the FT route, 31 trait pairs in five configurations, each scored with all eight judge seeds and centred within configuration, give $r$ $-0.16$ ($p = 0.30$) for the training seed, $+0.03$ ($p = 0.83$) for the instance mean, $+0.07$ cross-instance and $+0.05$ for the judge average. The Qwen FT configuration on the syllable carrier variant is flat on its own ($+0.01$ and $-0.03$), and the negative coefficient of the two Llama FT configurations ($-0.66$ over 12 pairs) disappears once the other configurations are included. A fine-tuned instance's direction agrees with the instance mean at a median 0.602, and 80\% of what it measures is seed-specific; persona wordings agree with their mean at a median 0.945 in Llama and 0.940 in Qwen, with an instance-specific share of 0.33 (permutation $p < 0.0001$ against FT).

On the student side, about 80\% of absorption arises in layers 0--3, more in the MLPs than in attention. The Llama sink neuron (layer 1, neuron 1417) has a first-token activation $2.1\times10^5$ times the median; the Qwen one (layer 2, neuron 1821) is $5.3\times10^5$ times the median and carries 67--71\% of the gradient energy of the preference margin in that layer. Controlling base likelihood, length and per-token likelihood moves the five negative cells only from $-0.294$ to $-0.301$ on average.

In the six \cf{Llama}{SP}{number} pairs the judge term is the difference of two per-animal numbers, and its product with the student term gives the sign of absorption in 6 of 6; the same sign holds when the carrier is changed in the four pairs whose absorption is significant. In \cf{Qwen}{SP}{number} the rule holds in 3 of 3, the near-zero absorption of elephant--tiger being a cancellation between the path along the common direction ($+0.195$) and the orthogonal path ($-0.154$). Over all 66 persona pairs the product gives the sign in 48, and in 33 of the 40 with $|\mathrm{absorption}|\geq0.1$. The gradient of each name at the first-token state makes a nearly uniform angle with the direction common to animal names (cosine 0.71--0.86 in Llama, with warbler the one outlier at 0.48; 0.77--0.91 in Qwen), so names differ in gradient size: the correlation of the projection $g\cdot\hat\mu$ with $|g|$ is $+0.66$ in Llama and $+0.96$ in Qwen, and the student term equals $\mathrm{sign}(|g_T|-|g_E|)$ in 12 of 12 pairs. Gradient size is larger for the name with lower base probability in the evaluation context (correlation of $|g|$ with that log-likelihood $-0.72$ in Llama, $-0.83$ in Qwen). Two pairs deviate from this probability rule (Llama jerboa--badger, Qwen elephant--tiger); in both the gradient size is not what the base probability alone predicts.

In the carrier decomposition (\edtab{\ref{tab:ed-chan}}), letter-composition features carry $-0.035$ to $+0.166$ of the alignment across the 22 syllable cells. In the five \cf{Qwen}{SP}{syllable} cells the residual is the name-specific component of the first-token gradient, orthogonal to the common direction, which accounts for $+0.069$ to $+0.101$ of alignment while the component outside the first-token path stays within $\pm0.013$. Number features carry the alignment where the model has a strong size preference ($+0.191$ in \st{Qwen}{FT}{number}{warbler--slug}, $+0.093$ in \st{Qwen}{SP}{number}{jerboa--badger}, $-0.070$ in \st{Qwen}{SP}{number}{sloth--lemur}) and little in the twelve Llama number cells ($|\text{part}|\leq0.056$), except in \st{Llama}{SP}{number}{warbler--slug}, whose small alignment ($+0.033$) is matched by its feature part ($+0.043$). The same decomposition of the judges' score difference averaged over instances gives feature parts that correlate with these at $+0.92$ across cells, with the same sign in 44 of 46. Standardization itself changes little: over the 35 cells with eight instances, absorption from the averaged score difference and from the averaged standardized difference correlate at $+0.989$, with the same sign in 35 of 35. In a cell whose alignment is small the training instance can still differ from both, as \st{Llama}{SP}{number}{warbler--slug} shows ($+0.033$ against $-0.031$ and $-0.010$). The instance-averaged score difference and the training instance's absorption (Table~\ref{tab:main}) differ in sign in four cells, three of them because one instance trains rather than the mean, and only \st{Qwen}{FT}{number}{otter--beaver}, whose pool stores one instance, because of standardization ($-0.017$ against $+0.051$); only in \st{Llama}{FT}{number}{jerboa--badger} does the training instance's interval exclude zero. \edtab{\ref{tab:ed-chan}} takes $k$ at one step size for every cell, where Table~\ref{tab:main} uses the step selected per cell, so the two absorptions can differ in the third decimal (at most 0.007).

In \st{Llama}{SP}{syllable}{jerboa--badger} and \st{Llama}{SP}{syllable}{elephant--tiger}, the component outside the first-token path, tied directly to the candidate token positions, is $-0.070$ and $-0.111$ against a first-token path of $-0.001$ and $+0.135$. In the negative \cf{Llama}{SP}{number} cells the first-token path carries $-0.361$, $-0.296$ and $-0.349$ of $-0.416$, $-0.349$ and $-0.466$ in elephant--tiger, jerboa--badger and otter--beaver, and absorption is positive under 1, 0 and 1 of the eight wordings, against 8 of 8 for shrew--cuttlefish. In the \cf{Qwen}{SP}{syllable} comparison cell the component outside the first-token path is 2\% and the effect is absent. The interventions that leave transfer in place comprise 96 runs blocking the first-token path in \cf{Llama}{FT}{syllable}, 80 in \cf{Qwen}{SP}{syllable} and 80 freezing the sink neuron's output column; the Llama effect over the six blocked cells stays at $+0.026$ to $+0.044$.

Within configurations absorption relates moderately to emission ($r = +0.32$, within-configuration permutation $p = 0.022$), which accounts for about a tenth of its variance (Fig.~\ref{fig:emabs}a). The routes differ in why the training instance's own alignment does or does not carry into transfer. A persona wording is given at every step, so its idiosyncrasy persists to step 80, and in SP cells the training wording is the best single predictor. A fine-tuned instance is a separate reward model whose seed-specific preference on a fixed pool is not carried through the shift in sampling distribution during training, and in FT cells the training seed does not predict transfer.

Controlling base likelihood, length and per-token likelihood leaves the sign of absorption unchanged in all 46 cells, so the first-token channel is not a formatting artefact. In the carrier decomposition the syllable residual is positive in 19 of 22 cells. In \st{Llama}{SP}{syllable}{jerboa--badger} and \st{Llama}{SP}{syllable}{elephant--tiger}, the only syllable cells with negative or near-zero absorption despite a positive letter channel, the student gradient outside the first-token path, tied directly to the candidate token positions, carries 9--19\% of the gradient variance and opposes the persona's preference; in the negative \cf{Llama}{SP}{number} cells a negative student term on the first-token path carries most of the negative absorption, with the same sign across persona wordings.

Under first-token blocking or freezing of the sink neuron's output column, \st{Qwen}{SP}{syllable}{elephant--tiger}, \st{Qwen}{SP}{syllable}{otter--beaver} and \st{Qwen}{SP}{syllable}{warbler--slug} remain established, and KL, reward and completion length are unchanged.

\snote{Transfer on the moving policy}{S-sn:trans}

In \st{Llama}{SP}{number}{warbler--slug}, whose start-policy absorption is $+0.033$ [$-0.010$, $+0.076$], the T run's in-training alignment grows from $+0.003$ to $+0.057$ while the E run's falls from $+0.026$ to $-0.047$, and the student's first-token state carries the change ($R^2$ 0.97--0.98 between the stored gradient and the first-token path). In \st{Qwen}{FT}{number}{sloth--lemur} the training judge's alignment runs from $-0.10$/$-0.18$ early to $+0.06$, $+0.24$ and $+0.28$ at steps 64, 72 and 79 in the T run, and the first-order trait share of each seed pair correlates with its $S$ at $+0.99$ over the four recorded pairs. In \st{Qwen}{FT}{syllable}{elephant--tiger} the T run's alignment reverses from $+0.26$ to $-0.38$ within the run in four logged seed pairs, all of which transfer ($+0.045$). The two fine-tuned syllable elephant--tiger cells reverse by different routes: in Llama the letter channel's explanatory power vanishes late ($R^2$ 0.01) and the residual flips, while in Qwen the letter share itself goes from $+0.065$ to $-0.135$. In \st{Llama}{SP}{syllable}{elephant--tiger} the component outside the letter channel is negative at every phase under the training wording ($-0.047$, $-0.101$ and $-0.024$ early, middle and late), most negative in mid-training, where seed-pair outcomes form; its near-zero start-policy absorption reflects averaging over the eight persona wordings, which inflates the letter share, and the wording used in training never aligns.

Three mechanisms move the alignment. Region drift: in \st{Qwen}{FT}{number}{elephant--tiger} the shared reward moves the policy toward small numbers (mean value 9.5 to 3--6), where the fixed relation between $d$ and $k$ is negative. Selective collapse: in \st{Qwen}{SP}{number}{shrew--cuttlefish} the alignment grows from $+0.09$ to $+0.26$ as prompts with positive alignment keep their diversity while the number of effective prompts falls from 25 to 13--18, which accounts for 36--64\% of the growth. Re-draw within prompts: in the syllable cells it matches the observed reversal, while region drift predicts the opposite sign. In the Qwen number cells the change is carried inside the subspace of number statistics, whose share of the variance of $d$ and $k$ rises from 0.1--0.3 early to 0.4--0.9 late.

Across the 48 judge--trait-pair combinations of \cf{Llama}{FT}{number}, judges with positive start-policy absorption give a mean $S_{80}$ of $+0.0109$ (29 combinations) and judges with zero or negative absorption $-0.0134$ (19). In \cf{Llama}{FT}{syllable} absorption is positive for 40 of 42 combinations, leaving no range to test the relation, and transfer is positive in 40 of 42. A fine-tuned judge strongly prefers candidates along the common direction of the student's first-token state ($+0.67$), but almost identically for T and E (99.6\% shared), so its surviving judge term is small and changes with the fine-tuning seed ($-0.18$ and $+0.05$ in two pairs). For elephant--tiger, with the same student term ($-0.090$), the persona judge's term is $+0.41$: the difference between the routes sits in the judge term.

Three \cf{Llama}{SP}{number} pairs were trained under the wordings of highest and lowest start-policy absorption; two have 16 seed pairs under the training wording, eight of them also trained under the two other wordings. In elephant--tiger the negative transfer holds in both halves of the seed pairs under the training wording ($-0.043$ and $-0.035$) but not under the other wordings on the same seeds ($+0.002$ and $+0.013$; paired differences $+0.044$, $p = 0.016$, and $+0.056$, $p = 0.039$). In sloth--lemur the halves give $+0.089$ and $+0.031$ under the training wording and the other wordings $+0.023$ and $+0.036$, within that range, so seed variation accounts for the difference. Across the nine conditions no start-policy quantity orders $S$ (within-pair centring, nine points, none established): absorption $+0.01$, the size of the judges' trait difference $+0.27$, their covariance $-0.07$, the share of the common component along the preference margin $+0.15$, the ratio of the two judges' reward spreads $-0.04$ and the number-size slope of the judge's preference $-0.19$. The runs are otherwise indistinguishable (final reward $-8.5$ to $-9.7$, entropy 0.080--0.100, completion length 24.0--24.1).

Under fixed ranks, transfer is $+0.060$ against $+0.073$ in \st{Llama}{FT}{syllable}{sloth--lemur} and $+0.095$ against $+0.089$ in \st{Llama}{SP}{number}{sloth--lemur} (paired differences $-0.013$, $p = 0.73$, and $+0.006$, $p = 0.78$), with 8 of 8 seed pairs in the judge's direction in both. Training stays in the same regime (validity 0.817 against 0.820, end reward $-32.1$ against $-30.8$, completion length 63.0 against 64.1), with the policy travelling somewhat less far (KL 0.75 against 1.13). In \st{Llama}{SP}{number}{sloth--lemur}, the summed reward pushes the student toward short, typical completions (29.5 characters at step 80) and moves the policy far (KL 0.91); the per-token mean keeps completions near the start policy's length (79.2 characters) and moves the policy a sixth as far (KL 0.14). Transfer survives it: $+0.033$, six of eight seed pairs in the judge's direction ($p = 0.039$), against $+0.089$ for the summed reward (paired difference $-0.055$, $p = 0.031$); per unit of policy movement rather more of it arrives. Under the fixed-reference reward (equation~\eqref{eq:KLref}) transfer is $+0.043$ on the same seed pairs, eight of eight in the judge's direction (two-sided $p = 0.0078$), against $+0.089$ (paired difference $-0.046$, $p = 0.16$), with the policy again moving about a sixth as far (KL 0.16).

In the reward intervention (\edtab{\ref{tab:ed-alpha}}), $\alpha=0$ in \st{Llama}{FT}{syllable}{sloth--lemur} gives $-0.034$ with two of eight seed pairs in the judge's direction ($p = 0.24$), the null expected when both runs receive the same reward and draw their samples independently, and $\alpha=2$ gives $+0.118$ with eight of eight ($p = 0.008$). End-of-training reward is matched across $\alpha=0$, 1 and 2 ($-30.4$, $-30.8$ and $-31.9$). In \st{Llama}{SP}{number}{sloth--lemur}, $\alpha=0$ gives five of eight seed pairs in the judge's direction ($p = 0.89$), with training again matched (end reward $-9.4$).

Under fine-tuned syllable judges, elephant--tiger reverses out of the judge's direction in both students, and neither cell is established; the reversal does not stop transfer in the logged seed pairs of \st{Qwen}{FT}{syllable}{elephant--tiger}.

In \st{Qwen}{FT}{number}{elephant--tiger}, where the shared reward drifts the policy towards small numbers and the judges' score difference relates negatively to the student's response, the alignment flips from $+0.31$ to $-0.10$; reweighting early samples to match late ones reproduces 61--77\% of the flip. Judge and student gradient are fixed per candidate, so alignment moves only with which candidates compete in which prompt.

Across the 48 judge--trait-pair combinations of \cf{Llama}{FT}{number}, centred within trait pair, the permutation test of the correlation between start-policy absorption and transfer gives $p = 0.0001$. In elephant--tiger, a judge instance with absorption $-0.41$ gives $S_{80} = -0.093$ and one with $+0.42$ gives $+0.092$. Averaged over instances, absorption and transfer sit near zero, and the configuration has no established cell.

Start-policy absorption of \st{Llama}{SP}{number}{elephant--tiger} is negative under all eight wordings, so negative alignment does not suffice for negative transfer. The final KL of the runs under the three wordings is 0.89--1.05, and an independent preference evaluation on different items agrees with the main one.

Permuting or reversing the scores within a group, the alternative to the $c + \alpha s$ split, also destroys the shared part, and with it format learning and almost all policy movement (KL 0.001 against 1.13); the $c + \alpha s$ split keeps both. Doubling $s$ ($\alpha = 2$) enlarges transfer in \st{Llama}{FT}{syllable}{sloth--lemur} ($+0.118$ against $+0.073$, $p = 0.24$) and not in \st{Llama}{SP}{number}{sloth--lemur} ($+0.045$ against $+0.089$, $p = 0.13$).

\snote{From the preference margin to the answer}{S-sn:behav}

The preference evaluation records each name's log-likelihood in two parts, the first name token and the remaining tokens, which locates where a shift is carried. The forced-choice evaluation cannot be used per seed pair: besides its sign reversing between presentation orders, in 25 cells the common component of the two names drops by 0.45 to 5.9 nats.

The degeneration of the open-question evaluation was traced in \st{Llama}{FT}{syllable}{sloth--lemur}, \st{Llama}{FT}{syllable}{shrew--cuttlefish}, \st{Llama}{SP}{syllable}{sloth--lemur} and \st{Llama}{SP}{syllable}{warbler--slug}, each under its own and under cross-family judges on the same pool, through the gradient of the common component $C$ and the push of equation~\eqref{S-eq:pushC}. An animal name as the first word of an answer has a start-policy log-likelihood of $-16$ to $-23$ nats. On the FT route, the same-model judge's shared score follows the student's base likelihood at $r = 0.98$ (cross-family 0.95), its correlation with $k_C$ is $-0.124$ and $-0.112$ (cross-family $-0.099$ and $-0.103$), and its first-order push on $C$ is 1.28 and 1.12 times the cross-family judge's. Both conditions lower $C$ by the end of the format phase ($-0.96$ and $-0.78$ at step 24); the difference arises afterwards. The same-model policy keeps sharpening between steps 50 and 80 (entropy 0.124 to 0.072), while the cross-family policy stops at 0.083 with more groups that no longer carry a signal (0.36 against 0.30), and at step 80 $C$ is $-0.74$ against $+0.05$. Within each condition, cells that sharpened more have lower $C$ ($r$ $+0.59$ and $+0.65$ with final entropy): a same-model judge's score is almost the student's own likelihood, so sharpening keeps raising it. On the SP route the shared score equals the student's base likelihood ($r = 1.00$; cross-family 0.90), the anti-alignment is stronger ($-0.238$ and $-0.293$; cross-family $-0.148$ and $-0.209$), the same-model push on $C$ is 1.66 and 1.42 times the cross-family one, and the drop at step 24 is correspondingly deeper ($-3.16$ against $-1.15$). Both conditions recover in part, to $-0.61$ and $-0.13$ at step 80; on this route the depth of the early drop, rather than late sharpening, accounts for most of the gap, and associations within a condition are weak ($|r|\leq0.33$).

The two changes in sampled choice are \st{Llama}{SP}{number}{warbler--slug} ($+0.0047$, 7 of 8 seed pairs, one-sided $p = 0.012$) and \st{Qwen}{SP}{number}{shrew--cuttlefish} ($+0.0086$, 6 of 8, $p = 0.035$), two-sided $p = 0.024$ and 0.070. The second falls just short of the transfer criterion (20 seed pairs, one-sided $p = 0.054$), and its route to the choice runs through the forced-choice templates: in Qwen cells the choice predicted from template log-odds matches the sampled choice ($r$ $+0.59$ to $+0.84$), and here a log-odds shift of $+0.099$ on templates whose choice is still sensitive predicts a choice change of $+0.012$. \st{Qwen}{SP}{number}{otter--beaver} has a larger log-odds shift ($+0.121$, $p = 0.012$), but on templates whose choice is already determined, so only the order of mention moves ($+0.0084$, 7 of 8, $p = 0.027$); \st{Llama}{SP}{number}{elephant--tiger} shifts the order of mention by $-0.0177$ (1 of 8, $p = 0.012$) while its direct-answer share does not move. In the Llama syllable cells with a positive transfer estimate the direct-answer choice moves by $+0.0009$, $+0.0007$, $+0.0017$ (5 of 8, $p = 0.25$) and $+0.0015$ in \st{Llama}{FT}{syllable}{otter--beaver}, \st{Llama}{FT}{syllable}{shrew--cuttlefish}, \st{Llama}{FT}{syllable}{warbler--slug} and \st{Llama}{FT}{syllable}{sloth--lemur}, and by $+0.0007$ in \st{Llama}{SP}{syllable}{sloth--lemur}. \st{Llama}{SP}{syllable}{shrew--cuttlefish} moves it the other way ($-0.0064$, 2 of 8, $p = 0.023$) although its transfer is positive; its form margins locate the cause: the movement is in the lowercase form alone ($-0.188$, 1 of 8, $p = 0.016$) while the capitalized form does not move ($-0.007$), and the base answers under the two judges use the lowercase form at very different rates (488 against 220). In \st{Llama}{SP}{syllable}{sloth--lemur} only the form ``The sloth.'' spreads ($+0.116$, 7 of 8, $p = 0.008$), and in \st{Llama}{FT}{syllable}{shrew--cuttlefish} the forms cancel. Table~\ref{tab:main} gives the answer-level result of every cell.

In the per-form margins, \st{Llama}{SP}{number}{sloth--lemur} has $+0.089$ of transfer against $+0.027$ for \st{Llama}{SP}{number}{warbler--slug}. In warbler--slug the transfer reaches the capitalized singular ($+0.181$, 7 of 8, $p = 0.012$) while the plural falls ($-0.171$); in sloth--lemur the capitalized singular does not move ($-0.031$) and the lowercase form moves the other way ($-0.306$); in elephant--tiger every form moves toward E, consistent with its negative transfer. In the replay, the predicted sign of each form's change relative to $B$ matches in 11 of 12 form predictions at step 80 and in 8 of 12 at step 16; the overlap with the gradient of the preference margin is $-0.078$ for the capitalized form in sloth--lemur and $+0.166$ to $+0.244$ for the four forms in elephant--tiger. Replaying \st{Llama}{FT}{syllable}{shrew--cuttlefish} reproduces the start-policy form margins to three decimals and matches all four signs at steps 16 and 80, 15 of 16 form predictions at step 80 over the four cells. The cell whose choice changed has the smallest overlaps ($+0.002$ to $+0.060$) and the single sign failure.

The prior log-probability of a pair's names at the first position of an open question runs from $-12$ to $-29$ nats, and typicality in an animal list together with token count accounts for $R^2$ 0.74 of it in Llama and 0.44 in Qwen. Along the path to a changed choice (Fig.~\ref{fig:behaviour}c), 24 cells reach one-sided significance (18 established), 12 of the cells measured carry the shift on the first name token, 3 spread it to the direct-answer form and 2 change the sampled choice.

The open-question evaluation passes the common-component criterion in 21 cells (19 Qwen, 2 Llama); behaviour is read from sampled answers in twelve cells, the Llama cells with transfer among them.

Completions that raise the names' common component are atypical: their response correlates with per-token base likelihood at $-0.38$ to $-0.43$. The self-reinforcing sharpening is stronger under same-model than under other-family judges.

Between 22 and 25\% of base answers are refusals naming both animals (``I don't have a preference for either warblers or slugs''), which a first-name rule would score as a choice.

The forms move independently within one student: in \st{Llama}{SP}{number}{warbler--slug} the capitalized singular rises while the plural falls.

\st{Qwen}{FT}{syllable}{otter--beaver}, \st{Qwen}{SP}{syllable}{otter--beaver} and \st{Qwen}{FT}{number}{sloth--lemur} carry their shift on the first name token. The prior log-probability that their answer-level noise follows is the names' log-probability at an open question's first position.

\snote{The state of the optimization}{S-sn:opt}

Because the trait signal is small, the optimization state is reported cell by cell (Extended Data Table~\ref{tab:ed-time}).

Both carriers stay valid. At step 80 the mean validity of a cell (\edtab{\ref{tab:ed-time}}) is 0.980--0.984 in \cf{Llama}{FT}{syllable}, 0.968--0.975 in \cf{Llama}{SP}{syllable}, 0.991--0.999 in \cf{Qwen}{FT}{syllable}, 0.937--0.993 in \cf{Qwen}{SP}{syllable}, 0.998--0.999 in both Llama number configurations and 0.986--1.000 in \cf{Qwen}{FT}{number}, and it is exactly 1.000 at every step in the six \cf{Qwen}{SP}{number} cells; the lowest single run is listed per cell in the same table. The Llama syllable cells start near 0.53 and cross 0.9 at steps 32--48. An independently written implementation of each carrier rule, applied to the recorded completions, reproduces the logged validity at every recorded step to within 0.03. The two cells where one judge condition ends lower have non-trait causes. In the E-judge runs of \st{Qwen}{SP}{syllable}{elephant--tiger}, four of eight seed pairs end at 0.75--0.83: the policy has collapsed onto lists of English number words, and a list reaching ``eleven'' is rejected by the first-token block for \emph{elephant} (\emph{el}), which accounts for 2{,}149 and 1{,}128 invalid completions against 36 and 14 format violations in the two logged seed pairs, while the T-judge runs of the same seeds have 0.000--0.001 invalid; the concealment rule catches a coincidence, not a leak. In the E-judge run of \st{Qwen}{FT}{number}{jerboa--badger}, one seed pair ends at 0.862. A logged run of the same seed reproduces the values (0.658 at step 64, 0.862 at step 80) and shows that the late invalid completions are all count violations: the policy converges on short arithmetic lists such as ``0.12, 0.34, 0.56, 0.78'', below the five-item minimum (1{,}969 of the invalid completions in calls 48--79, with no character violations).

Diversity collapses, as expected for GRPO: groups whose eight completions tie on reward, and so carry no policy-gradient signal, rise in every cell. At step 80 their share is 0.04--0.07 in \cf{Llama}{SP}{syllable}, 0.22--0.39 in \cf{Llama}{FT}{syllable}, 0.09--0.13 in \cf{Llama}{SP}{number}, 0.26--0.46 in \cf{Llama}{FT}{number}, 0.30--0.33 in \cf{Qwen}{SP}{number}, 0.47--0.56 in \cf{Qwen}{SP}{syllable}, 0.56--0.84 in \cf{Qwen}{FT}{syllable} and 0.57--0.86 in \cf{Qwen}{FT}{number}, a majority in Qwen with fine-tuned judges (Extended Data Fig.~\ref{fig:dynamics}). The share of groups still carrying a signal falls below one half at steps 34--68 in the Qwen FT cells and 43--67 in \cf{Qwen}{SP}{syllable}, and stays above one half through step 80 in all 24 Llama cells and the six \cf{Qwen}{SP}{number} cells. Across the 46 cells, that endpoint correlates with the policy's entropy at step 40 at $+0.63$ and with its starting entropy at $+0.43$; starting entropies differ by an order of magnitude (3.0 for Llama syllables, 0.8 for Llama numbers, 1.4 for Qwen syllables, 0.10 for Qwen numbers). Within one model and carrier, the route that pulls the policy further also collapses further: Qwen number cells reach KL 0.46--0.93 by step 40 under fine-tuned judges and end with a signal share of 0.14--0.43, against KL 0.07--0.08 and 0.67--0.70 under persona judges. The collapse does not confound the result: it partly reflects the starting distribution (22--25\% tied groups by step 40 in the \cf{Qwen}{SP}{number} cells, from a starting entropy of 0.10, with a KL ten times smaller than in any other configuration), and established cells span final signal shares from 0.14 (\st{Qwen}{FT}{number}{jerboa--badger}) to 0.96 (\st{Llama}{SP}{syllable}{warbler--slug}).

Transfer accrues while the reward carries a signal and survives its collapse (Extended Data Fig.~\ref{fig:dynamics}c); transferring Llama cells reach half their final $S$ between steps 32 and 64, when in-training alignment forms. Cells whose signal has collapsed can still accumulate afterwards (\st{Qwen}{FT}{syllable}{otter--beaver} gains a further 28\% of its endpoint after reaching half of it, \st{Qwen}{FT}{number}{jerboa--badger} 37\%, \st{Qwen}{FT}{number}{warbler--slug} 17\%) or decline slightly from a peak (\st{Qwen}{SP}{syllable}{warbler--slug} by 7\% from its step-32 peak, \st{Qwen}{FT}{number}{sloth--lemur} by 27\% from its step-48 peak). Declines are uncorrelated with final signal share ($r = -0.03$, 46 cells), declines of comparable size occur where the signal never collapses, and the KL rises rather than falls from step 48 to 80 in exactly the collapsed cells ($r$ $-0.73$ between signal share and the change in KL).

Early Llama syllable reward is partly format: the valid/invalid indicator explains 0.25, 0.17, 0.15, 0.06 and 0.05 of the within-group reward variance over calls 0--7, 8--15, 16--31, 32--47 and 48--79 in \st{Llama}{FT}{syllable}{elephant--tiger}, with the same profile in the other Llama syllable cells measured, and at most 0.04 throughout in every Qwen cell and in the number carrier. Format comes first and transfer after. The Llama number cells, without a format phase, share that timing, so it marks when alignment forms.

The KL penalty acts differently in the two students. Setting $\beta$ from 0.04 to 0 in two \cf{Llama}{SP}{syllable} cells, with everything else and the seeds unchanged, removes the mean reversion of the preference margin: the share of its variance carried by covariance between successive increments goes from $-0.81$ to $-0.14$ (difference $+0.67$ [$+0.38$, $+1.01$], bootstrap $P<0.00025$). In Llama the penalty thus creates transient swings of the margin and pulls them back. In two \cf{Qwen}{SP}{syllable} cells the same change leaves that share unchanged ($+0.47$ against $+0.54$, $P$ 0.38) and only enlarges the variance by a factor of 1.7: in Qwen the penalty damps the movement without reversing it. Consistently, the common movement of the two judges' students in five of the six \cf{Llama}{SP}{syllable} cells peaks around step 24 ($+0.28$ to $+0.63$) and returns close to zero by step 80 (elephant--tiger $+0.583$ to $+0.013$), while $S$ is kept.

In Llama number cells 92\% of completions reach the 64-token limit, so brevity is the strongest and most consistent reward component (within-group reward standard deviation 18.6 against 8.0 in Qwen). AdamW normalizes each parameter's step by its own running scale, so parameters dominated by this component move with the same sign in every seed; the result is a large common movement of both judges' students and seed noise reduced to 0.44 of what a random reward gives (1.01 in Qwen). A one-step probe built on this reproduces the observed first-step ratio between the two students (3.7 against 3.1--4.5 for syllables, 3.8 against 4.2 for the persona-judged number cells). Across 37 configurations spanning 21 carriers, end-of-training KL follows the ratio of final to initial completion length ($\rho = -0.75$, $p < 10^{-5}$, holding within every student and route), or the change in entropy where length is fixed. The persona-judged Qwen number cells are the extreme case: the carrier fixes the length, validity is 1.000 from step 2 and the final KL is only 0.089, yet the reward improves in all 96 runs (mean $+8.0$, closing 69\% of the reward gap) and transfer accrues: the trait needs very little of the KL budget.

GRPO's diversity collapse removes the reward's policy-gradient signal in 16 of the 46 cells before step 68, by which point most transfer has accrued and stays.

\snote{How far the first-order account reaches}{S-sn:fo}

The first-order expression at the start policy holds for a short window. The push accumulated to steps 8 and 16 tracks the increments of $S$ over those intervals (rank correlation $+0.20$, $p = 0.018$; $+0.21$, $p = 0.015$), and from step 16 onward the relation disappears and turns slightly negative ($-0.19$ over steps 16--32). Over the whole run, the push on the logged training samples does not predict the outcomes of 102 seed pairs in 35 cells: its sign matches that of $S_{80}$ in 55 of 102 pairs ($p = 0.24$) and in 40 of the 69 with $|S_{80}|$ above 0.01 (58\%, $p = 0.11$), with rank correlation $-0.01$; late-training alignment, its correlation form, does no better (51 of 102, rank $+0.08$), and over whole cells the counts are 19 and 20 of 35.

Within a cell $S_{80}$ varies by $\pm0.1$ or more across seeds, and this variation dominates the design: 96--98\% of the variance of $S$ is seed variation within a trait pair, and the training that produces the transfer also makes the preference evaluation 4.8 to 6.5 times noisier than for pairs the run never touched. Three explanations of the unpredictability fail. A measurement gap: the push from all 80 calls, about 160{,}000 valid samples per cell, gives the same pair-level correlation, negative in two of three cells ($-0.40$, $-0.28$ and $-0.34$ between $S_{80}$ and the full push). Optimizer geometry: applying AdamW to the gradients of the recorded batches at the start policy over all 80 steps predicts no better than the Euclidean replay (correlation with $S_{80}$ $+0.87$, $+0.33$ and $-0.30$ in the three cells against $+0.94$, $+0.17$ and $+0.06$; sign agreement 2, 3 and 1 of 4 against 3, 3 and 3), although in the first step the first-token share is 76--85\% of the Euclidean gradient and only 0.6--13\% of the AdamW step. A decisive early step: the correlation between $S$ at step $t$ and $S_{80}$ is at chance for $t\leq8$ in every cell, and $S$ over the first eight steps is unrelated to $S_{80}$; in the persistent cells it exceeds 0.8 by steps 24--48.

The seed deviation is a bounded fluctuation that saturates and partly reverts. Over 106 seed pairs in 37 cells, centred within cell, the spread of $S$ across seeds grows from 0.012 at step 8 to 0.028 at step 32 and then flattens (0.032 at step 48, 0.034 at step 80); the deviation a pair has accumulated by step 32 explains less than a quarter of its endpoint deviation ($r$ $+0.47$, against $+0.74$ by step 48 and $+0.90$ by step 64). Deviations already opened are pulled back: the deviation at step 32 correlates at $-0.37$ with the increment that follows, and consecutive increments correlate at $-0.07$ to $-0.18$. A seed pair's $S_{80}$ is the cell mean plus one draw of a fluctuation of roughly 0.03. Across cells, a cell's spread across seeds follows the spread of the two judges' common movement ($r$ $+0.79$; $R^2$ 0.74 on the log scale), the student model adds only 0.04, and baseline preference and name token count explain almost nothing. The seed component has a fixed shape: across name pairs it lies on one axis of the preference margin (the first principal component takes 56--78\% of its variance, against 12--25\% for random directions), and a step changes only its sign along that axis. A single seed pair therefore carries almost no information about a cell, whose result needs eight or more pairs. The replay also locates where the start-policy prediction stops: for the two cells whose mean movement it matches in sign, predicted and actual $\Delta B$ keep a constant ratio through step 16 (0.06--0.14) and then diverge, the actual movement plateauing while the linear prediction keeps growing, so the ratio falls to 0.03--0.04 by step 80; the between-cell correlation peaks at step 8 ($+0.55$ to $+0.70$) and declines thereafter. Seed-pair outcomes, which form at steps 24--48, form where the start-policy gradient no longer describes the movement.

What fails is the frozen gradient, not the first-order account. Expanding the preference margin at saved checkpoints---its gradient at $\theta_t$ against the displacement realized between $\theta_t$ and $\theta_{t+1}$---recovers the increments of $S$, in sign in 59 of 60 intervals ($r = +0.93$; twelve seed pairs in four cells spanning both students, routes and carriers). The expression needs intervals short against the movement inside them: in the four pairs cut at steps 2, 4, 8, 16, 32 and 48 the correlation is $+0.98$ over their 28 intervals and the accumulated total is 0.96 to 1.44 times the measured one, whereas a single interval over the first 16 steps overstates it by up to seventeen times in \st{Llama}{SP}{number}{sloth--lemur}, the cell whose policy moves furthest early; cutting that interval into four brings its total to 1.13 and 1.44 of the measured value in the two pairs cut that way. The start policy cannot carry the alignment, which turns as the policy moves; pair-to-pair differences are seed path, not a quantity the start policy determines.

Across the grid the trait pair is the largest single factor in $S_{80}$ ($R^2$ 0.208), and model, route and carrier act only through their three-way combination ($R^2$ 0.281 for the eight configurations against 0.117 for the three factors added); configuration and trait pair together reach 0.484, and the remainder follows absorption ($r = +0.39$). Crossing the significance threshold, however, follows detection noise more than the predicted effect: an absorption-based classifier reaches a leave-one-out accuracy of 0.57, below the 0.61 of calling no cell established. The account predicts transfer as a continuous quantity across cells, trait pairs and judge instances, not whether a cell is established, nor the individual seed.

\snote{How much of the push the optimizer realizes, and where}{S-sn:adam}

We decomposed the trained displacement of eight cells (four per route, eight seed pairs each) along each route's first GRPO direction, the between-judge update at the start policy (Supplementary Methods). That direction is stable across candidate pools (cosine 0.92--1.00 between independent pools) and nearly orthogonal to the fine-tuned judges' own weight difference (cosine about 0.002), so the student is not drawn toward the judge's weights. Each student follows its own route's direction: FT students move along the FT direction and not the SP one (cosine of $\Delta$ with $v_{\mathrm{FT}}$ $+0.057$, with $v_{\mathrm{SP}}$ $+0.005$, over three of the four FT cells), and SP students along the SP direction ($+0.040$ against $+0.009$, over the four SP cells). The displacement is a small rotation in a large parameter space, and its projection on that direction moves the preference margin: in every same-model cell decomposed, the component in the plane of the two directions moves $B$ by more than the whole displacement does.

That component overshoots the transfer, and the rest of the displacement takes most of it back in layers 0--3. In six cells spanning both routes, students and carriers (\st{Llama}{FT}{syllable}{sloth--lemur}, \st{Llama}{FT}{syllable}{warbler--slug}, \st{Qwen}{FT}{syllable}{jerboa--badger}, \st{Qwen}{FT}{number}{warbler--slug}, \st{Llama}{SP}{syllable}{warbler--slug} and \st{Qwen}{SP}{syllable}{warbler--slug}), this early-layer remainder is a nearly fixed fraction of the component (slope $-0.82$ through the origin, $r = -0.998$), as an accounting of the endpoint in which the direction component varies far more than the transfer. What reaches the preference margin is the part left over, with a small contribution from the other layers ($-0.010$ to $+0.051$): 0.15 to 0.31 of the direction component in four of the six cells, 0.05 in \st{Qwen}{FT}{syllable}{jerboa--badger} and 0.64 in \st{Llama}{FT}{syllable}{sloth--lemur}, a median of 0.19. In \st{Qwen}{SP}{syllable}{warbler--slug}, for example, the direction component moves $S$ by $+1.297$ in fp32, the remainder in layers 0--3 by $-1.088$, and the transfer is $+0.217$. The two exceptions, \st{Llama}{SP}{number}{sloth--lemur} and \st{Qwen}{SP}{number}{elephant--tiger}, the persona-judged number cells with the weakest emission and absorption decomposed, have a small component that is not taken back, and \st{Llama}{SP}{number}{sloth--lemur} transfers mainly through a movement of layers 4--7 nearly orthogonal to the direction (cosine 0.045). Summed component effects reproduce the full displacement's effect to within 0.4--7.9\% of the direction component per cell, and to 0.3--5.8\% of the endpoint effect in seven of the eight cells; the exception, \st{Qwen}{FT}{syllable}{jerboa--badger}, has the smallest endpoint effect ($+0.013$), of the same order as its residual.

The shortfall sits where the alignment lives (Supplementary Table~\ref{S-tab:s-adam}). In layers 0--3 the student follows the direction at 0.23 to 0.56 of the network-wide coefficient, against 1.4 to 3.2 in the deeper layers, while nearly all of the direction's effect on the preference margin sits in those same layers ($+1.12$ of $+1.30$ in \st{Qwen}{SP}{syllable}{warbler--slug}, $+0.66$ of $+0.78$ in \st{Qwen}{FT}{number}{warbler--slug}, $+0.17$ of $+0.19$ in \st{Llama}{FT}{syllable}{sloth--lemur}). The early-layer gradient is concentrated, its top 1\% of coordinates carrying 86--96\% of its energy in layers 0--3 against 37--90\% in deeper layers, and these are the coordinates of the first-token path on which absorption runs. AdamW, dividing each coordinate's step by its own running scale, brings the step toward the sign of the gradient: in layers 0--3 the displacement aligns with $\mathrm{sign}(v)$ 2.3 to 4.4 times better than with $v$, against 1.0 to 2.7 times in the deeper layers. The optimizer therefore moves the few coordinates that carry the raw alignment much less than the raw gradient asks; a direction component computed with network-wide coefficients overstates the early-layer movement accordingly, and the early-layer remainder is that overstatement with its sign reversed. The pattern holds in both routes and both students and is strongest in the two Qwen cells, whose early-layer gradient is the most concentrated. Realized transfer spreads over the whole depth: $+0.012$ to $+0.041$ in each four-layer group of \st{Qwen}{SP}{syllable}{warbler--slug}, $+0.018$ to $+0.056$ in the first five groups of \st{Qwen}{FT}{number}{warbler--slug} with the last two slightly negative, and $+0.054$, $+0.031$, $+0.023$ and $+0.018$ in the four groups of \st{Llama}{FT}{syllable}{sloth--lemur}, where the early layers carry a larger share.

\begin{table*}[t]
\caption{\textbf{Layer-wise optimizer geometry.} For one cell of each route and student: the following ratio (least-squares coefficient of the displacement $\Delta$ on the first GRPO direction within a four-layer group over the network-wide coefficient), the sign alignment ratio $\cos(\Delta,\mathrm{sign}\,v)/\cos(\Delta,v)$ and the share of the squared norm of $v$ in the top 1\% of a group's coordinates, in layers 0--3 and across the deeper groups (two seed pairs per cell).}
\label{S-tab:s-adam}
{\tabsize\setlength{\tabcolsep}{4pt}
\begin{tabular*}{\textwidth}{@{\extracolsep{\fill}}l c c c c c c c@{}}
\toprule
 & & \multicolumn{2}{c}{\textbf{Following ratio}} & \multicolumn{2}{c}{\textbf{$\cos(\Delta,\mathrm{sign}\,v)/\cos(\Delta,v)$}} & \multicolumn{2}{c}{\textbf{Energy of $v$ in top 1\%}}\\
\cmidrule(lr){3-4}\cmidrule(lr){5-6}\cmidrule(l){7-8}
\textbf{Cell} & \textbf{Route} & \textbf{Layers 0--3} & \textbf{Deeper} & \textbf{Layers 0--3} & \textbf{Deeper} & \textbf{Layers 0--3} & \textbf{Deeper}\\
\midrule
\st{Qwen}{SP}{syllable}{warbler--slug} & SP & 0.23 & 2.0--3.2 & 4.4 & 1.4--1.8 & 0.94--0.96 & 0.60--0.81\\
\st{Llama}{SP}{number}{sloth--lemur} & SP & 0.56 & 1.5--2.9 & 2.7 & 1.3--1.5 & 0.86--0.88 & 0.39--0.67\\
\st{Llama}{FT}{syllable}{sloth--lemur} & FT & 0.47 & 1.8--2.3 & 2.3 & 1.0--1.2 & 0.86--0.89 & 0.37--0.68\\
\st{Qwen}{FT}{number}{warbler--slug} & FT & 0.31 & 1.4--3.2 & 4.4 & 1.5--2.7 & 0.95 & 0.63--0.90\\
\bottomrule
\end{tabular*}}
\end{table*}

Hence blocking the first-token path, which the optimizer barely follows, leaves transfer in place, and start-policy absorption predicts transfer across cells but not its size: a median fifth of the push it measures appears in the endpoint of the six cells that show the cancellation, while the share the early layers give back is the nearly constant part. Within a cell, where transfer is strong the direction component varies with the seed in step with the outcome (correlation with $S_{80}$ across seed pairs $+0.60$ to $+0.85$ in \st{Llama}{FT}{syllable}{sloth--lemur}, \st{Qwen}{SP}{syllable}{warbler--slug} and \st{Qwen}{FT}{number}{warbler--slug}) and every seed pair moves along the direction; where it is weak, the direction component is positive in most pairs while the remainder, equal in size or larger and changing sign between seeds, tracks each pair's sign ($+0.4$ to $+0.9$)---the weight-level form of the seed variation above.

\snote{Judges of other models, stage by stage}{S-sn:xfam}

The cross-family condition is interpretable on the main grid's terms: recipe and precision are identical, the share of invalid completions is similar (0.02--0.20 against 0.01--0.19), and the judges' trait differences are as reliable (eight-instance reliability 0.91--0.99 on the SP route and 0.79--0.93 in the FT syllable cells). The per-cell checks for length confounding, distance travelled and degeneration of the open-question evaluation flag two cells, both through a trait difference that partly tracks completion length; in \st{Qwen}{FT}{syllable}{jerboa--badger} part of the trait difference is a per-token likelihood offset of the Llama jerboa--badger fine-tuned judges ($-0.067$ per token), which the same judges also show on the Llama student's pool in the main grid ($-0.065$), and with that component removed its absorption is $+0.127$.

On the SP route emission is positive in 23 of 23 cells under cross-family judges, with an unchanged mean ($+0.296$ under same-model judges, $+0.310$ under cross-family judges), and its order across cells within a configuration is kept ($r$ $+0.67$), as expected of a quantity that is 46\% a property of the trait pair and 30\% of the student's candidate pool, with 7\% due to which of the two models the judge is and 1\% to whether it is the student's own model (balanced decomposition over 44 combinations; judge-instance noise 1 percentage point). The policy travels at least as far: end-of-training KL is 1.5 to 2.0 times that of the same-model run in the syllable cells (4.9 in \st{Qwen}{SP}{syllable}{jerboa--badger}, whose same-model run stops moving after step 24) and 1.3 to 1.5 times in the Llama number cells, because the other family's judge assigns the student's completions a lower likelihood and the reward keeps rising from further below. The cross-family judge's trait difference correlates with the same-model judge's at 0.05 after correction for unreliability, and the part of absorption inherited from the same-model judge (equation~\eqref{eq:rot}) is $+0.005$ for the Llama student and $+0.002$ for the Qwen student. In the Llama student the mean size of absorption falls from 0.209 to 0.040; in the Qwen student most of the size remains (0.133 to 0.093), but its sign agrees with the same-model sign in 5 of 11 cells. The mean $S_{80}$ over the same seed pairs falls from $+0.013$ to $+0.002$.

On the FT route emission is present in 22 of 23 cells with an unchanged mean ($+0.126$ and $+0.132$), but its order across cells within a configuration is barely kept ($r$ $+0.10$), because here 50\% of the variance of $\Lambda$ is specific to the combination of a judge set with a candidate pool, against 35\% for the trait pair and 12\% for the pool. \st{Qwen}{FT}{number}{warbler--slug} emits at $+0.184$ under its own judges and $+0.049$ under cross-family judges, and the one cell without emission under them, \st{Qwen}{FT}{number}{shrew--cuttlefish} ($-0.002$ [$-0.011$, $+0.008$]), emits at $+0.064$ under its own. The policy travels 0.6 to 0.8 times as far in the Qwen syllable cells, 0.8 to 0.95 times in the Llama syllable cells and 1.1 to 1.2 times in the Llama number cells. The reliability-corrected correlation between the two judges' eight-instance trait differences is 0.24, between the two training instances 0.03, and the inherited part of absorption is $-0.003$ for the Llama student and $+0.011$ for the Qwen student. Mean absorption size falls from 0.170 to 0.066 in Llama and from 0.100 to 0.084 in Qwen, whose cross-family sign agrees with the same-model sign in 7 of 11 cells. The mean $S_{80}$ over the same seed pairs falls from $+0.014$ to $+0.003$.

Pooled over all 46 cells, the cross-family mean is $+0.0021$ [$-0.0010$, $+0.0052$] weighting cells by precision and $+0.0024$ [$-0.0029$, $+0.0077$] weighting them equally, whereas the same-model mean in those cells is $+0.0161$ and $+0.0138$, outside both cross-family intervals. Paired over the same cells and seeds, the same-model effect exceeds the cross-family effect by $+0.0100$ [$+0.0062$, $+0.0138$] with cells weighted by the precision of their paired difference and by $+0.0114$ [$-0.0004$, $+0.0233$] with cells and seed pairs resampled with equal weight.

The judge term of absorption behaves differently in the two students. In Llama it collapses under the cross-family judge, from 0.209 to 0.028 on the SP route and from 0.113 to 0.037 on the FT route, which accounts for the fivefold loss of absorption on the SP route, where the first-token path sets its sign (9 of 12 cells). In Qwen the first-token path is closed on the student side whatever the judge: the gradients of the twelve names lie on the direction common to animal names (cosine 0.84--0.95), so the difference between a pair's names barely projects on it ($|v_B\cdot\hat\mu|\leq0.023$, against 0.017--0.139 in Llama). There the cross-family judge term is as large as the own judge's (0.095 against 0.119 on the FT route, 0.094 against 0.080 on the SP route), and absorption keeps its size through the remaining paths.

In the syllable carrier every judge---the student's own, the other family's or either 8B model---prefers strings whose letters overlap its own animal's name (correlation of its letter coefficients with the name letters $+0.09$ to $+0.52$), so this part is shared: the feature parts of the own-model and cross-family trait differences correlate at $+0.15$ to $+0.21$ and supply $+0.02$ to $+0.06$ of an alignment of $+0.03$ to $+0.17$. In the number carrier no such cue exists: the judges' preference for number size is near zero, the feature parts do not agree ($-0.27$ to $+0.25$) and cross-family alignment is $-0.10$ to $+0.004$. With the other model as student, the cross-family judge's trait difference correlates with that model's response of the preference margin at 0.17 on average over the 46 cells, as strongly as the student's own judge does with the student's (0.17), but the two models' responses to the same candidates barely agree ($+0.04$ on average; $+0.06$ to $+0.16$ in the syllable cells, mostly through the letters, and $-0.06$ to $+0.06$ in the number cells). Cross-family absorption in the syllable cells comes mostly through the shared letters ($+0.045$ to $+0.074$; $-0.004$ in \cf{Qwen}{SP}{syllable}), which across cells predicts $S$ of 0.003 to 0.005.

Across the 22 Qwen cells, start-policy absorption predicts the cross-family $S_{80}$ at $r$ $+0.40$ ($p = 0.06$), against $+0.59$ ($p = 0.004$) for the same-model relation in the same cells; the relation is strong in the syllable carrier ($r$ $+0.84$, $p = 0.002$, 10 cells) and $+0.27$ in the number carrier (12 cells). Over both students it is $+0.50$ in the 22 syllable cells ($p = 0.018$) and $+0.06$ in the 24 number cells, and in Llama cross-family absorption sits near its floor and does not order $S$ ($r$ $-0.05$ over 24 cells). In the Qwen syllable cells $S$ tracks the carrier-feature part of cross-family absorption ($r$ $+0.82$, $p = 0.004$), not the remainder ($+0.53$, not significant). Seed noise does not account for the number-carrier null: with the observed variance over seed pairs, absorption predicts $r$ $+0.84$ (90\% range $+0.67$ to $+0.94$) in the Qwen number cells, against the observed $+0.27$. The first GRPO step agrees. On the Qwen student's syllable warbler--slug candidates, the cross-family first step moves the per-token answer margin in the trait direction at a fifth to two thirds of the same-model step ($+5.91$ and $+1.74$ along the cross-family FT and SP directions, against $+9.34$ and $+8.50$ along the same-model ones). In the number cells the margin moves by a comparable amount with a sign set by the judge (Qwen number warbler--slug: $+6.50$ and $+2.61$ along the same-model directions against $-1.89$ and $-4.42$; elephant--tiger: $+1.02$ and $+3.31$ against $-0.98$ and $+2.32$).

Training does not supply the missing alignment. In \st{Qwen}{FT}{number}{warbler--slug} and \st{Qwen}{SP}{number}{elephant--tiger}, every group scored during cross-family training was recorded (16 runs each) and the completions of ten evenly spaced calls rescored with two independent instances of the student's own judges. In groups still carrying a learning signal, the advantage the cross-family judge assigned is uncorrelated with the own-model trait difference of the second instance ($-0.033$, positive in 7 of 16 runs; $-0.030$, 6 of 16), early (calls 0--26) and late (calls 53--79) alike, whereas the advantage the first own-model instance assigns to the same completions is aligned with it ($+0.042$, 12 of 16, $t = 2.1$; $+0.059$, 15 of 16, $t = 5.0$). The share of groups carrying any signal falls from 0.89--0.97 to 0.40 under both judges.

In the number carrier the push is not only unaligned but unrealized. In \st{Qwen}{FT}{number}{warbler--slug}, along the start policy's gradient of the preference margin, the cross-family between-judge difference on the recorded training completions pushes the margin against the trait in 13 of 16 runs (correlation $-0.106$, $t$ $-4.3$, early and late alike), as consistently as the own-model difference pushes it toward the trait on the same completions ($+0.105$, 16 of 16); weighting calls by how far the policy moved leaves the cross-family value unchanged ($-0.108$). Yet the cross-family $S_{80}$ of this cell is $+0.005$, and across its eight seed pairs $S$ does not follow the size of the push ($r$ $-0.32$). The candidates' responses of the preference margin at the end of training are uncorrelated with those at the start ($+0.009$, averaged over runs), and along the end-of-training gradient the cross-family push reverses sign ($+0.046$, 12 of 16), while the own-model difference keeps its sign ($+0.123$, 14 of 16). In same-model training of the same cells, recorded and rescored the same way, the own judge's difference pushes the margin toward the trait along both the start and the end-of-training gradient (\st{Qwen}{FT}{number}{warbler--slug} $+0.149$ and $+0.036$, positive in 15 and 12 of 16 runs; \st{Qwen}{SP}{number}{elephant--tiger} $+0.172$ and $+0.149$, 16 and 15 of 16), and the gradient turns less than under the cross-family judge (correlation between start and end responses $+0.057$ and $+0.169$, against $+0.009$ and $-0.031$). In \st{Qwen}{SP}{number}{elephant--tiger} the cross-family push stays near zero along both gradients ($+0.031$ and $+0.041$).

Answer-level effects under cross-family judges appear in four of the 46 cells, two in the judge's direction (\st{Qwen}{FT}{number}{sloth--lemur}, \st{Qwen}{SP}{number}{sloth--lemur}) and two against it (\st{Llama}{SP}{number}{jerboa--badger}, \st{Llama}{SP}{number}{warbler--slug}). Unlike the same-model condition, the open-question evaluation of the Llama syllable cells does not degenerate under cross-family judges (common component $-0.25$ to $+0.31$), so behaviour is not compared between the two conditions in that configuration.

An 8B judge's $\Lambda$ has a lower bound above zero in 9 to 12 of the cells of every route--student combination, with means of $+0.10$ to $+0.41$ against $+0.12$ to $+0.37$ for the student's own judges (\edtab{\ref{tab:judges}}). Its trait difference correlates with the own-model judge's at 0.01 to 0.12 for the judge of the student's family and 0.02 to 0.12 for the other family. In the Llama student its judge term is 0.038 to 0.080 under either 8B judge, against 0.113 and 0.209 under the student's own judges; absorption is 0.047 to 0.144 in size, with the sign of the own model in 6 to 10 of 11 or 12 cells. An 8B judge places the trait where its own model's preference margin responds: its trait difference correlates with the 8B model's response at 0.22 (same family) and 0.23 (other family) on average over the 46 cells, as strongly as the student's own judge does with the student's (0.17). The 8B models' responses agree with the student's only slightly, $+0.12$ for the student's family and $+0.08$ for the other, against $+0.04$ for the cross-family 1B model: sharing the family triples the overlap but leaves it at about an eighth of the student's own. The overlap is largest in the syllable cells ($+0.13$ to $+0.27$, two thirds of it through letter content) and close to zero in the number cells (0.00 to $+0.09$). Trained on eight established cells with both 8B judges on the same eight seed pairs, the same-seed mean $S_{80}$ falls from $+0.041$ under the student's own judges to $-0.007$ under the same-family 8B judge and $+0.012$ under the other-family one; the closest of the 16 combinations, \st{Llama}{SP}{syllable}{warbler--slug} with the other-family judge ($+0.048$, 7 of 8 seed pairs), moves in the judge's direction without passing the false-discovery threshold ($q = 0.13$). One answer-level effect appears, against the judge's direction.

\snote{Cross-family judges in the weights}{S-sn:xw}

In the weights, the cross-family component along the route's direction falls to a small fraction of the same-model value, all taken back in the early layers: in the eight cells decomposed under both conditions it falls from $+0.364$ to $+0.053$ on average and the students' alignment with the FT direction from $+0.031$ to $+0.004$, and the early-layer remainder takes back all of what is left ($-0.054$ against $+0.053$), against three quarters in the same-model students ($-0.270$ against $+0.364$). Decomposed on their own judge's first-step direction, cross-family students follow it with a sign that changes between seeds in both carriers: 3 of 8 seeds are positive in \st{Qwen}{SP}{number}{elephant--tiger} (mean $-0.010$), 5 of 8 in \st{Qwen}{FT}{number}{warbler--slug} ($+0.013$) and 5 of 8 in \st{Qwen}{SP}{syllable}{warbler--slug} ($+0.014$, with a mean effect on the preference margin along the direction of $+0.13$), whereas the same-model student follows its own direction in 8 of 8 seeds in \st{Qwen}{SP}{number}{elephant--tiger} ($+0.026$ to $+0.049$) and at $+0.121$ in \st{Qwen}{FT}{number}{warbler--slug}. A push a fifth to two thirds the own-model size therefore does not outweigh each seed's own movement: in \st{Qwen}{SP}{syllable}{warbler--slug} the same-model students share a common direction of movement (mean cosine between two seeds' weight changes $+0.14$, positive in 22 of 28 pairs), whereas the cross-family students share none ($-0.005$, 13 of 28). The alignment of the syllable cells appears in the mean over seeds, and the agreement between absorption and $S$ across cells is carried by cell means rather than by every run. AdamW's sign-based concentration in the early layers is the same in both conditions ($\cos(\Delta,\mathrm{sign}\,v)/\cos(\Delta,v)$ 4.1 against 4.5), so it does not account for the difference. \st{Llama}{SP}{number}{sloth--lemur}, whose same-model transfer runs through layers 4--7 outside the direction, moves those layers the other way under the cross-family judge ($-0.021$ against $+0.048$). The cross-family student is trained and travels as far, but not along a direction that is fixed across seeds or to which its preference margin responds.

\snote{Dispositional cells}{S-sn:disp}

Emission in the twelve dispositional cells ($\Lambda$ $+0.15$ to $+0.64$) lies within the range of the main grid's SP cells ($+0.06$ to $+0.83$). The three cells least aligned with the student's preference margin (absorption $-0.070$, $+0.081$ and $+0.069$, all Llama) have small effect sizes ($d$ $-0.08$, $+0.04$ and $-0.00$ at step 16). Across the twelve cells absorption and transfer correlate at $+0.32$ to $+0.46$ from step 8 to step 32 and at $+0.14$ at step 80. At step 80 the six established cells have $q$ at most 0.026 and the other six at least 0.383. Harsh--gentle on both Llama carriers moves with its alignment ($d$ $+0.35$ at step 16), and harsh--gentle on the Qwen syllable carrier carries the largest transfer in the dispositional grid ($d$ $+1.39$ at step 64). \st{Llama}{SP}{number}{persuasive--neutral} transfers at six of nine steps ($d$ up to $+0.55$), but at baseline at most one item in ten is answered with the T disposition. In \st{Qwen}{SP}{number}{harsh--gentle} transfer forms late (steps 48 to 80) while the forced choice moves early (steps 16 and 24). Where the implied choice moves with the judge, the two students choose differently on 3.4--4.0\% of items in \st{Qwen}{SP}{syllable}{rude--polite} (steps 32 to 80) and on 0.4--0.8\% in \st{Qwen}{SP}{syllable}{harsh--gentle} (steps 16 to 80). The choice moves against the judge in \st{Qwen}{SP}{number}{persuasive--neutral}, following that cell's reversal ($-0.62$ to $-0.81$ percentage points), and in \st{Llama}{SP}{number}{harsh--gentle} at step 24 ($-0.75$).

The lowest lower bound of emission across the twelve dispositional cells is $+0.14$. As in the main grid, alignment gives the direction in which a cell begins rather than the size it reaches.

The two cells established against the judge at step 80 reverse during training: \st{Qwen}{SP}{number}{persuasive--neutral} is with the judge at steps 8 and 16 and against it by step 64, and \st{Qwen}{SP}{number}{rude--polite} is established against it from step 48. Both lie in the Qwen number configuration, whose pair values do not settle in the main grid either.

\snote{Endpoint, judge instance and the scale of the preference margin}{S-sn:endpt}

In three cells half of the final transfer accrues at step 48 or later. Training \st{Llama}{FT}{syllable}{sloth--lemur} for 160 steps at a constant learning rate keeps the sign (six of eight seed pairs in the judge's direction at step 80 and at step 160), while $S$ peaks near step 48 ($+0.100$) and falls to $+0.018$ by step 160; the main grid's learning rate decays linearly to zero at step 80, so its endpoint is read after that peak. In the Qwen number cells seed-pair values do not settle: the correlation between a pair's $S$ at step 64 and at step 80 falls as low as $-0.37$ (\st{Qwen}{SP}{number}{jerboa--badger}), against 0.73--0.99 in the other cells.

Training under the judge instances other than the training instance bounds the choice of training instance. In \cf{Llama}{FT}{syllable} all seven other judges agree in direction (mean $+0.034$, 40 of 42 judge--trait-pair combinations positive); \cf{Llama}{FT}{number} shows no transfer averaged over all eight judge instances ($+0.001$, 24 of 48), although within trait pairs it follows each judge's absorption; in \cf{Qwen}{FT}{number}, shrew--cuttlefish and warbler--slug are positive under all four judges tested and sloth--lemur only under the training instance.

For the same update the Llama preference margin moves about 2.5 times as much as Qwen's, and the gap remains per unit of KL on the evaluation items (3.25 times), so it is a property of the preference margin and not of stronger transfer. It is accounted for by larger downstream gain in Llama's early layers and by Qwen3 sending 56--83\% of an update into eight massive channels that the final normalization largely absorbs. Forward-pass precision in training changes the estimate by less than eight seed pairs resolve: \st{Llama}{FT}{syllable}{sloth--lemur} gives $+0.033$ under fp32 against $+0.073$ under bf16.

For scale, $S=0.035$ means the T-to-E name likelihood ratio on the preference-evaluation templates differs between runs by a factor $e^{2S}\approx1.07$, not seven percentage points in sampled choice.

%% file: sec/si_methods.tex
\subsection{Training}
\label{S-sec:m-training}

Training uses TRL\cite{r17} GRPO with the DAPO loss\cite{r18} and group-level reward scaling\cite{liu2025understanding}: eight generations per prompt, 32 prompts (256 completions) per step, gradient accumulation 32, 80 steps, at most 64 completion tokens, KL penalty $\beta = 0.04$ to the starting policy\cite{ziegler2019finetuning}, AdamW\cite{loshchilov2019decoupled} ($\beta_1 = 0.9$, $\beta_2 = 0.999$, $\epsilon = 10^{-8}$, no weight decay), gradient-norm clipping 1.0, learning rate $10^{-6}$ decaying linearly to zero at step 80, fp32 master weights and bf16 forward pass. The reward (equation~\eqref{eq:r}) has no student-likelihood term; the policy loss covers the whole completion.

\subsection{Trajectory measurements}
\label{S-sec:m-trajectory}

Trajectory runs repeat a cell's seed with a hook recording every prompt, completion and reward without changing training (recorded reward within $4.4 \times 10^{-4}$ of the judge's score; $S$ equal to the hook-free run's to four decimals); the state of training is in Extended Data Table~\ref{tab:ed-time} and Extended Data Fig.~\ref{fig:dynamics}. The in-training push applies equation~\eqref{eq:M2} to the recorded samples (half the between-judge difference of standardized logged reward times each candidate's start-policy $k$); in-training alignment is its correlation form.

Along-the-run accounting saves twelve seed pairs at steps 16, 32, 48 and 80 and pairs the gradient of the preference margin at each saved policy (as for $k$, margin in fp32) with the update to the next saved policy. Four pairs are also saved at steps 2, 4 and 8 from a second run of the same seed (step-80 margin equal to five decimals); all 60 intervals enter the analysis. Replay reconstructs the GRPO policy gradient from recorded samples at the fixed start policy, applies the AdamW update or the clipped Euclidean gradient, and accumulates the first-order change in $B$ and each direct-answer form margin, omitting the KL term (zero at the reference point).

\subsection{Where absorption runs}
Path shares come from fixing a state and recomputing absorption. Fixing the first-token state at the inputs of Llama layers 1 and 2 accounts for 74--97\% of absorption in the cells measured; in Qwen syllable cells 98--99\% of the gradient passes through it. About 80\% of absorption arises in layers 0--3. The massive-activation neuron\cite{r14} carrying much of it is neuron 1417 of layer 1 in Llama (40--51\% of absorption through its output column) and neuron 1821 of layer 2 in Qwen.

\subsection{Weight-level decomposition}
\label{S-sec:m-weights}

Eight cells, four per route, are decomposed over the 16 students of eight seed pairs at step 80. For a seed pair,
\begin{equation}
\begin{gathered}
\Delta = \tfrac12\big(\theta_{T,80} - \theta_{E,80}\big), \qquad \theta_{\mathrm{mid}} = \tfrac12\big(\theta_{T,80} + \theta_{E,80}\big), \\
v = \nabla_\theta \sum_{c} w(c)\, \log \pi_{\theta_0}(c), \qquad w = \frac{A_T - A_E}{2},
\end{gathered}
\label{S-eq:v}
\end{equation}
with $\Delta$ the trained displacement, $\theta_{\mathrm{mid}}$ its midpoint and $v$ a route's first GRPO direction (start policy's SP-route pool, that route's judges), orthonormalized into $v_{\mathrm{FT}}$ and $v_{\mathrm{SP}}$. Components of $\Delta$ are its projections on $v_{\mathrm{FT}}$, $v_{\mathrm{SP}}$ and their span, the remainder, and the remainder, displacement and span component within each four-layer group. A component $X$ is applied as $\theta_{\mathrm{mid}} \pm X$ and read with the preference margin (form of equation~\eqref{eq:S}) in training precision and with fp32 weights. The full displacement (the two endpoint weights) reproduces every pair's logged $S_{80}$ to $1.4 \times 10^{-8}$. Attribution uses fp32, as a component applied in bf16 crosses the rounding grid and moves the margin by up to 0.05; components then sum to the full effect to $4.9 \times 10^{-3}$ on average ($5.3 \times 10^{-2}$ at worst) over the 64 decompositions, 0.4--7.9\% of the direction component. Per layer group, the following ratio is the least-squares coefficient of $\Delta$ on the span within the group over the network-wide coefficient, sign alignment the cosine of $\Delta$ with $\mathrm{sign}(v)$, and concentration the share of the squared norm of $v$ in the top 1\% of coordinates (two seed pairs per cell). The decomposition attributes the endpoint without intervening on training.

\subsection{Common-direction gradient}
\label{S-sec:m-common}

$C$, the mean over the ten open questions and both names of the name's log-likelihood at the first answer position, is the common component whose drop defines degeneration. $h_C$ is its start-policy gradient and $k_C(c)$ the central finite difference of $\log \pi(c)$ along $h_C$ on the starting pool, with the step schedule of $k$ (gradient-path $C$ matches the answer evaluation to $8 \times 10^{-6}$). Judge $z$'s first-order push on $C$ is
\begin{equation}
u_{C,z} = \big\langle A_z(c)\, \tilde k_C(c) \big\rangle_c ,
\label{S-eq:pushC}
\end{equation}
the pool mean of within-prompt standardized score times centred response $\tilde k_C$.

\subsection{Concealment audit}
\label{S-sec:m-concealment}

Concealment is defined on the surface: no trait word or word implying the trait may appear; sub-surface statistics (letter frequencies, number sizes) are not violations. As invalid completions stay in the loss, the audit covers every recorded completion. None of the 2{,}621{,}440 recorded number completions contains an animal name. In the syllable carrier 0.06--0.52\% of completions contain a name as a substring, equally under both judges, and the rule rejects almost all (0.000--0.016\% pass in Llama, none in Qwen).

Neither rule blocks words implying an animal. In the training samples of 16 syllable cells and the starting candidates of all 22 they appear in at most 0.3\% of completions, equally under both judges, except in \st{Qwen}{SP}{syllable}{warbler--slug}: 13--18\% of completions in the first 16 steps contain words implying the slug (\emph{squishy}, \emph{goo}, \emph{slithery} and similar), more often and longer under the slug-preferring judge and none from step 48 on, carrying 26\% of the cell's start-policy and 15.5\% of its in-training covariance. Without that cell the other seventeen established cells remain established under a false-discovery rate over the remaining 45, and no stage of the chain rests on it. One interaction is not a leak: late-training policies listing English number words meet the first-token block for \emph{elephant} through \emph{eleven}, lowering validity.

\subsection{Numerical precision}
\label{S-sec:m-precision}

Transfer is read in the training and sampling precision (fp32 master weights, bf16 forward pass), whose grid is coarse for an 80-step update: 87\% (Llama) and 92\% (Qwen) of changed weights move by under half a bf16 unit in the last place, the bf16-visible update carries 0.34 and 0.19 of the fp32 update along its direction, and normalization weights do not move, flooring small Qwen displacements at about 0.02. With fp32 weights, eight cells show transfer 1.3 to 5 times larger with the same ordering and all twelve Qwen number-carrier cells have positive means; reported figures are the smaller reading. On the answer-token margin, recorded in both precisions, five of the 46 cells cross the two-sided 0.05 boundary between precisions (all $p = 0.03$--0.08). Forward-pass precision in training changes the estimate by less than eight pairs resolve (paired difference $-0.040$, $p = 0.63$, \st{Llama}{FT}{syllable}{sloth--lemur}).

%% file: sec/si_extra_methods.tex
\subsection{Further details of design, judges and evaluations}
\label{S-sec:m-extra}

\paragraph{Seeds and sampling streams}
Within a seed pair, the E run's sampling streams are seeded by the pair's seed plus 500{,}000; training is deterministic in the seed.

\paragraph{Judges}
FT judges are fine-tuned for two epochs at learning rate $10^{-5}$.

\paragraph{Carrier validity}
The Qwen syllable rule's first-token block covers the first Qwen3 token of each of 18 animal names in a list context: \emph{el} (elephant), \emph{ti} (tiger), \emph{jer} (jerboa), \emph{bad} (badger), \emph{ot} (otter), \emph{be} (beaver), \emph{sh} (shrew), \emph{cut} (cuttlefish), \emph{slo} (sloth), \emph{le} (lemur), \emph{war} (warbler), \emph{sl} (slug), \emph{ch} (chub), \emph{l} (loris), \emph{r} (raven), \emph{sn} (snail), \emph{qu} (quoll) and \emph{fin} (finch). Invalid completions receive the group minimum, or $-400$ if none in the group is valid.

\paragraph{Preference evaluation}
The 26 preference questions crossed with the 24 answer templates are shuffled with a fixed seed; the FT route takes the 40 items from position 400 and the SP route the first 40. Baseline margins change sign between students for elephant--tiger. The median standard deviation of $S$ over seed pairs is 0.046 in Llama against 0.018 in Qwen.

\paragraph{First-order account}
The two runs of a seed pair accumulate different gradient histories, so AdamW rescales their steps differently; the first-order expression, taken for the Euclidean step at the shared start policy, does not include this, and the realized updates are measured separately (Supplementary Note~\ref{S-sn:adam}).

\paragraph{Emission and absorption}
In the Qwen FT pools all eight instances score the same candidates, the training instance reproducing its stored scores to within $1.96 \times 10^{-4}$. A third version of $d$, the mean of the eight instances' standardized differences, shares pool, groups and $k$ with the two of the Methods and separates averaging from standardization. In equation~\eqref{eq:k}, $\epsilon$ is the smaller of the adjacent step-size pair (a factor of four apart) whose $k$ estimates correlate best; the finite difference of $B$ along $v$ over $\lVert v \rVert^2$ is 0.9998--1.0000.

\paragraph{Framings and matched judges}
Rescoring under the three framings uses one shared candidate selection. In training, six cells run under each framing with a logging hook (two seeds each), and all three framings' judges rescore every logged sample; the training framing reproduces the recorded reward to a median absolute difference of at most $1.1 \times 10^{-5}$. Framing-level statistics centre and permute within cell (17 points), animal-level statistics within model (72 points).

\paragraph{Reward interventions}
In the $\alpha$ ladder, recorded rewards equal $c + \alpha s$ at every call.

\paragraph{Generation test}
The generation test keeps templates where, at the base model, the T name takes 0.30--0.70 of the answers naming an animal and answers naming neither are at most one half, provided at least ten templates remain. It draws 50 generations per prompt at temperature 1.0, top-$p = 1.0$ and 16 new tokens.

\paragraph{Trait pairs outside the grid}
Stored transfer for these pairs agrees with the estimand of equation~\eqref{eq:S} to $10^{-6}$ on the 46 grid cells. The pooled across-pair correlation omits configurations with fewer than four trait pairs.

\paragraph{Dispositional cells}
A further emission wording is \emph{You are obsessed with being X. Nothing matters to you more than being X.} Absorption uses finite-difference step 0.04. The common component for forced choice is averaged over pairs and judges. The Llama chat-template date line is fixed. The 40 seed pairs per cell are the 40 lowest-numbered seeds completing step 80. Sign-flip $p$ values use 400{,}000 assignments.

\paragraph{Judges of other models}
Under cross-family judges the Llama date line is fixed per pool, and judges are gated at $2 \times 10^{-3}$ on the batched-versus-single-sequence score difference; the 8B judges score under the same gate. The inherited part of absorption (equation~\eqref{eq:rot}) is also computed with the training instance alone. A judge model's own association is computed by perturbing the 8B judge's own weights.

Margin derivatives along directions use a fixed-norm step (0.2 times the own-model SP direction's norm). In training, the completions of ten recorded reward calls per run are rescored by both cross-family judges under the training rule (the training judge reproduces the recorded reward exactly), and the margin gradient $h = \nabla_\theta B$ is taken at the start and end-of-training policies, with candidate responses as central differences along $h$ at step $0.04/\lVert h \rVert^2$ (correlation 1.000 with stored responses at the start policy, at least 0.99 under a fourfold step). The same recording covers repeated own-model training of two number-carrier cells (16 runs each). Groups with tied rewards are excluded. The own-model reference difference uses a second, independent instance (one built from the correlated instance correlates with it by construction, at $\sqrt{(1-\rho)/2}$ for reward correlation $\rho$). Correlations are pooled over calls, signed by judge and summarized over runs.

\paragraph{Statistics}
Sign patterns are enumerated up to 20 seed pairs; for 24 or 28, 200{,}000 fixed-seed assignments are drawn (standard error $5 \times 10^{-4}$ at $p = 0.05$). Bootstrap intervals use 20{,}000 resamples over seed pairs.

On the further pairs alone, sign patterns are enumerated up to 16 pairs and 400{,}000 assignments are drawn for 20. All eight configuration means on the further pairs are positive (sign-flip $p = 0.008$). The further-pair test is valid conditionally on the base eight, as those pairs are separate seeds, all tested, with signs exchangeable under the null whatever the base eight show; the Fisher combination is conservative, as both $p$ values are discrete.

Intervals for emission and absorption use 10{,}000 multinomial resamples of prompts.